# Crowd Scene Analysis Using Deep Learning Techniques

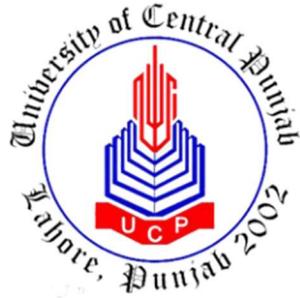

MASTER OF SCIENCE
IN
DATA SCIENCE

Submitted By
Er. Muhammad Junaid Asif
L1S22MSDS0012

DEPARTMENT OF COMPUTER SCIENCES
FACULTY OF INFORMATION TECHNOLOGY & COMPUTER SCIENCES
UNIVERSITY OF CENTRAL PUNJAB

March 2024

# Crowd Scene Analysis Using Deep Learning Techniques

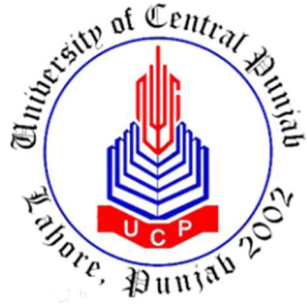

A Thesis submitted in partial fulfillment
of the requirements for the degree of

MASTER OF SCIENCE
IN
DATA SCIENCE

Submitted By
Er. Muhammad Junaid Asif
L1S22MSDS0012

Supervised By
Dr. Shazia Saqib

DEPARTMENT OF COMPUTER SCIENCES
FACULTY OF INFORMATION TECHNOLOGY & COMPUTER SCIENCES
UNIVERSITY OF CENTRAL PUNJAB

March 2024

# ABSTRACT

With the recent advancement in the field of deep learning and computer vision, crowd scene analysis has gained significant attention. UN predicts world population growth of 0.82% by 2035, driving people to cities for better lifestyles and social events like concerts, shopping, political gatherings, and educational conferences. Crowd scene analysis is crucial for ensuring a safe environment in public spaces, but manual monitoring can be laborious due to the risk of missing important information. An automatic solution is needed for efficient real-life applications.

Our research is focused on two main applications of crowd scene analysis: crowd counting, and anomaly detection. In recent years, a large number of researches have been presented in the domain of crowd counting. We addressed two main challenges in this domain *1).* Deep learning models are data-hungry paradigms and always need a large amount of annotated data for the training of algorithm. It is time-consuming and costly task to annotate such large amount of data. Self-supervised training is proposed to deal with this challenge. 2). M-CNN consists of multi-columns of CNN with different sizes of filters) by presenting a novel approach based on a combination of self-supervised training and Multi-Column CNN; This enables the model to learn features at different levels and makes it effective in dealing with challenges of occluded scenes, non-uniform density, complex backgrounds and scale in-variation. The proposed model was evaluated on publicly available data sets such as ShanghaiTech and UCF-QNRF by means of MAE and MSE.

A spatio-temporal model based on VGG19 is proposed for crowd anomaly detection, addressing challenges like lighting, environmental conditions, unexpected objects, and scalability. The model extracts spatial and temporal features, allowing it to be generalized to real-world scenes. Spatial features are learned using CNN, while temporal features are learned using LSTM blocks. The model works on binary classification and can detect normal or abnormal behavior. The model's performance is improved by replacing fully connected layers with dense residual blocks. Experiments on the Hockey Fight dataset and SCVD dataset show our models outperform other state-of-the-art approaches.

# DEDICATION

"To the three incredible Persons of my life;

My Grand Father Muhammad Safdar,

My Incredible Mother Prof. Nabila,

and

My Loving Wife Maham"

# ACKNOWLEDGEMENTS

- All Praises to **GOD Almighty** by whose Grace I was able to complete this research project. Thanks to HIM who blessed me with best at every step of my life.

- I would like to express my deepest gratitude and appreciation to **Prophet Muhammad (S.A.W)** for his profound teachings and exemplary life, which have served as a guiding light and source of inspiration throughout my thesis journey.

- We wish to offer our thanks to **Dr. Amjad Iqbal, Dean, FoIT&CS, UCP** who has been the main icon in our gearing up towards accomplishing something fruitful and helpful for the amelioration of society around us.

- I am deeply grateful to my research supervisor **Dr. Shazia Saqib, Associate Professor, FoIT&CS, UCP** for her constant support, and knowledge throughout the thesis process. Her leadership and effort have helped shape the direction and success of this research project. Her valuable comments, suggestion, and generous criticism helped me in writing this Thesis. I am very grateful to her from core of my heart for her expert guidance and sympathetic attitude. Their mentorship and guidance have been pivotal in my academic and research journey, and I am truly fortunate to have had the opportunity to work under their supervision.

- Next to her, I am really grateful to **Dr. Mujtaba Asad, Post Doctoral Research Fellow, SJTU, China (Ex. Assistant Professor, FoIT&CS, UCP)** for his valuable guidance and suggestion in selecting the topic for my thesis. Their insightful input and expertise have been instrumental in shaping the direction and significance of this research project. Dr. Mujtaba's extensive knowledge and expertise have significantly enhanced the research, providing valuable insights, guidance in identifying relevant literature,

- methodologies, and approaches. Dr. Mujtaba's unwavering support and encouragement throughout the thesis process have been a constant source of inspiration and motivation, helping me navigate challenges and uncertainties.

- I specially want to thanks to my mother **Prof. Nabila Shahzadi, Ex. Assistant Professor, Chemistry Department, PGC** for her unconditional love, unwavering support, and invaluable contributions to my whole life. I am truly fortunate to have my mother as my pillar of strength and guidance. Her wisdom, experience, and valuable insights have been instrumental in shaping the direction and scope of my life. Her constant encouragement, providing a peaceful environment, and taking care of everyday responsibilities have allowed me to focus on my graduate studies and research with utmost dedication.

- I want to express my sincere appreciation for my wife's continuous support in creating a peaceful and supportive environment for me to study and work on my thesis. Her understanding, encouragement, and belief in my capabilities have been instrumental in my success. I am grateful for my wife's understanding and patience during times when my studies demanded long hours of work and dedication.

- I would like to pay gratitude to my immediate boss **Engr. Farhan Ata Arain** for their unwavering support and encouragement throughout my studies.

- I would also like to express my sincere gratitude to my office colleague **Mr. Muhammad Imran Parvez** for their invaluable support in providing the necessary software for my research thesis. Their prompt assistance and expertise have been instrumental in facilitating the execution of my research project and ensuring access to essential tools for my research completion.

- Special thanks to **Mr. Shaheer Imran** for providing help while writing python code for my research project. I would also like to Thanks to **Mr. Saad** for teaching us Tools and Techniques in Data Science in such way that he provided us a solid foundation of different techniques of data science. I would also like to Thanks **Dr. Shazia Saqib** for teaching us the course of **Deep Learning** by adopting the practical approach.

# DECLARATION

I, *Muhammad Junaid Asif* S/O *Asif Bashir*, a student of **"Master of Science in Data Science"**, at **"Faculty of Information Technology & Computer Sciences"**, University of Central Punjab (UCP), hereby declare that this thesis titled, *"Crowd Scene Analysis using Deep Learning Techniques"* is my own research work and has not been submitted, published, or printed elsewhere in Pakistan or abroad. Additionally, I will not use this thesis for obtaining any degree other than the one stated above. I fully understand that if my statement is found to be incorrect at any stage, including after the award of the degree, the University has the right to revoke my MS/M.Phil. degree.

**Signature of Student:** *Muhammad Junaid*

**Name of Student:** Muhammad Junaid Asif

**Registration Number:** L1S22MSDS0012

**Date:**

VII

# PLAGIARISM UNDERTAKING

I solemnly declare that the research work presented in this thesis titled, ***"Crowd Scene Analysis using Deep Learning Techniques"*** is solely my research work, and that the entire thesis has been completed by me, with no significant contribution from any other person or institution. Any small contribution, wherever taken, has been duly acknowledged.

I understand the zero-tolerance policy of the HEC and University of Central Punjab towards plagiarism. Therefore, I as an author of the above titled thesis declare that no portion of my thesis has been plagiarized and that every material used from other sources has been properly acknowledged, cited, and referenced.

I undertake that if I am found guilty of any formal plagiarism in the above titled thesis, even after the award of MS/MPhil. degree, the University reserves the right to revoke my degree, and that HEC and the University have the right to publish my name on the HEC/University website for submitting a plagiarized thesis.

**Signature of Student:** *Muhammad Junaid*

**Name of Student:** Muhammad Junaid Asif

**Registration Number:** L1S22MSDS0012

**Date:**

# CERTIFICATE OF RESEARCH COMPLETION

It is certified that thesis titled, "Crowd Scene Analysis using Deep Learning Techniques", submitted by **Muhammad Junaid Asif**, Registration No. **L1S22MSDS0012**, for MS degree at **"Faculty of Information Technology and Computer Sciences"**, University of Central **Punjab (UCP)**, is an original research work and contains satisfactory material to be eligible for evaluation by the Examiner(s) for the award of the above stated degree.

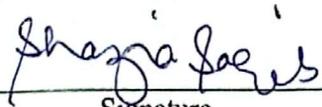
Signature

**Dr. Shazia Saqib**
Associate Professor
Faculty of IT and CS
University of Central Punjab

Date: 30-04-2024

# CERTIFICATE OF EXAMINERS

It is certified that the research work contained in this thesis titled **"Crowd Scene Analysis using Deep Learning Techniques"** is up to the mark for the award of **"Master of Science in Data Science"**.

**Internal Examiner**

Signature: 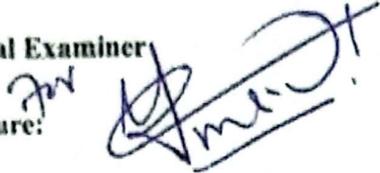

Name: Dr. Abdullah Yousafzai

Date: 30-04-24

**External Examiner:**

Signature: 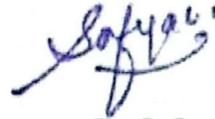

Name: Dr. Sufyan Ch.

Date: 30-04-2024

**Dean**
Faculty of IT & CS
University of Central Punjab (UCP)

Signature: 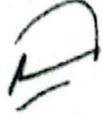

Name: Dr. Muhammad Amjad Iqbal

Date:

# TABLE OF CONTENTS

# LIST OF FIGURES

# LIST OF TABLES

`

# LIST OF ABBREVIATIONS AND ANCRONYM

| | |
|---|---|
| **CC** | Crowd Counting |
| **CAD** | Crowd Anomaly Detection |
| **UN** | United Nations |
| **CCTV** | Closed Circuit TeleVision |
| **NN** | Neural Network |
| **ANN** | Artificial Neural Network |
| **SNN** | Simulated Neural Network |
| **CNN** | Convolutional Neural Network |
| **M-CNN** | Multicolumn Convolutional Neural Network |
| **RCN** | Regions with CNN Features |
| **LSTM** | Long Short-Term Memory |
| **VAE** | Variational Auto Encoders |
| **VGG** | Virtual Geometry Group |
| **RNN** | Recurrent Neural Network |
| **GAN** | Generative Adversarial Network |
| **DM** | Density Maps |
| **GT** | Ground Truth |
| **ADAM** | Adaptive Moment Estimation |
| **Nadam** | Nesterov-accelerated Adaptive Moment Estimation |
| **RMSProp** | Root Mean Square Propagation |
| **WRB** | Wide Residual Block |
| **WDRB** | Wide Dense Residual Block |
| **ACSCP** | Adversarial Cross-Scale Consistency Pursuit |
| **RESNET** | Residual Network |
| **CSRNet** | Congested Scene Recognition Network |
| **CMTL** | Cascaded Multi Task Learning |
| **BRISK** | Binary Robust Invariant Scalable Keypoints |
| **BREIF** | Binary Robust Independent Elementary Features |
| **GMM** | Gaussian Mixture Modelling |
| **FEN** | Feature Extraction Network |
| **ReLU** | Rectified Linear Unit |
| **MRF** | Markov Random Field |
| **BoW** | Bag of Words |
| **ViF** | VIolent Flow |
| **SVM** | Support Vector Machine |
| **PCA** | Principal Component Analysis |
| **LSC-CNN** | Locate, Size, and Count Convolutional Neural Network |
| **SIFT** | Scale Invariant Feature Transform |
| **MAE** | Mean Absolute Error |
| **MSE** | Mean Squared Error |
| **HSV** | Hue, Saturation and Value |
| **AF** | Activation Function |
| **MLP** | Multi-Layer Perceptron |
| **BPNN** | Back Propagated Neural Network |

# CHAPTER ONE: INTRODUCTION

## 1.1 Crowd

*A large group of people gathered at a place to attend any public gathering, to enjoy any event, or to travel from one place to another can be defined as a crowd.* A crowd can be of different sizes, and its size may range from a few numbers of people to a large number of people in billions and millions. Based on features, purpose, and statistics, crowds can be of two main types; heterogeneous crowd and homogeneous crowd (*as shown in* **Figure 1.***1*) [1]

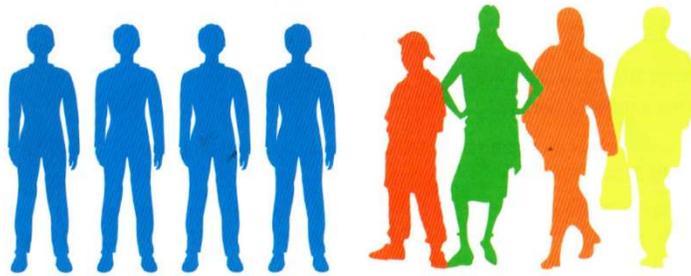

**Figure 1.1 Homogenous vs Heterogenous Crowds**

A *heterogeneous crowd* consists of people who have different characteristics, and different demographics. In this type of crowd, people may belong to different ethnicities, different age groups, different gender identities, and different physical characteristics. People gathered in this type of crowd may have different purposes of gathering, and different interests. For example; people gathered at any public transport station or airport, any public rally, political gatherings, cultural celebrations, and international conferences.

While homogeneous crowds consist of people having the same characteristics, the same purposes of gathering, same interests. Attendees from the same ethnicity, same age group, and same gender (male, female, or transgender) are part of homogeneous crowds. Common examples of this type of crowd include sports events, persons attending speeches of their favorite leader or political party, and persons gathering for a musical concert of their favorite singer or musical band.

Crowds can exhibit a wide range of behaviors, which can vary from peaceful to violent. It is influenced by a number of variables, including the situation, the people in the crowd, and the surroundings. Peaceful crowd behavior is characterized by calmness, cooperation, and orderly conduct. People might come together in these groups for shared objectives, such as taking part in a nonviolent demonstration, attending a peaceful protest, or celebrating an occasion. The crowd usually stays polite and amicable in such scenarios. However, crowds might also act more violently and aggressively. This can happen when tensions rise, emotions run high, or social order breaks down. In such scenarios, a number of individuals in the crowd might behave violently, aggressively, or vandalism towards other people or property. Factors such as perceived injustice, political unrest, or mob mentality can contribute to the emergence of violent behavior within a crowd. It is essential to keep in mind that crowd behavior is an intricate process that is affected by a wide range of variables, such as situational, psychological, and social aspects. Researchers, lawmakers, and law enforcement organizations can create measures that foster safe and peaceful crowd management while decreasing the possibility of violence by having an improved awareness of these dynamics.

Crowds can be physical or it can be virtual (*as shown in* **Figure 1.*2***) [2], [3]. For Example; people gather together for a political speech, mosques, and universities are considered as ***physical crowds***. While people connected through social media such as Microsoft Teams, Zoom, and Google Meet to attend any educational conference, debates will be considered ***virtual crowds.***

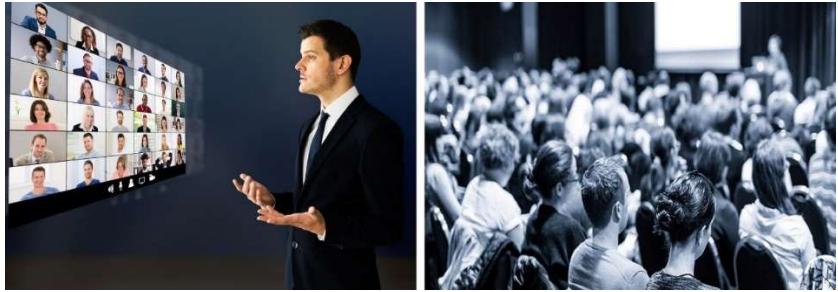

**Figure 1.2: Actual vs. Physical Crowds**

Virtual crowds are continuously monitored and recorded through cameras. While physical crowd requires installation of CCTV cameras for monitoring. It requires a big investment for installation of large set-up of CCTV cameras. Attentive manpower is also required to monitor the large number of cameras installed and analyze any kind of abnormality or unusual behaviors in the crowd. A single CCTV human operator isn't enough to analyze the sheer volume of surveillance videos collected from various points makes Therefore, a large number of operators are required to perform this task. In the case of a few operators, it may lead to the loss of any important scene or information.

It is very necessary to automate the analysis process of crowd scenes in order to study the crowd behavior, crowd density estimation, crowd flow analysis, crowd semantic analysis, and detecting of crowd anomalies. The main applications of crowd scene analysis include as following:

1. Crowd management and safety,
2. Security and surveillance,
3. Transportation and urban planning,
4. Customer behavior analysis,
5. Emergency and disaster management,
6. Event planning,
7. Political event planning and management.

In the light of above applications, crowd scene analysis can be divided into six major areas such as crowd counting, object detection and tracking, motion analysis, behavior analysis, anomaly detection, and crowd prediction [4].

1.  **Crowd counting:** Determines how many people are present in a certain location.
2.  **Objection detection:** Locates and identifies interesting/unusual objects in a crowd.
3.  **Motion analysis:** Assesses the crowd's general movement condition.
4.  **Behavior analysis:** Determines the crowd's collective characteristics.
5.  **Anomaly detection:** Recognizes odd and aberrant behaviors and occurrences.
6.  **Crowd prediction:** Foretells the proactive gathering or inflow of individuals into or out of a given area.

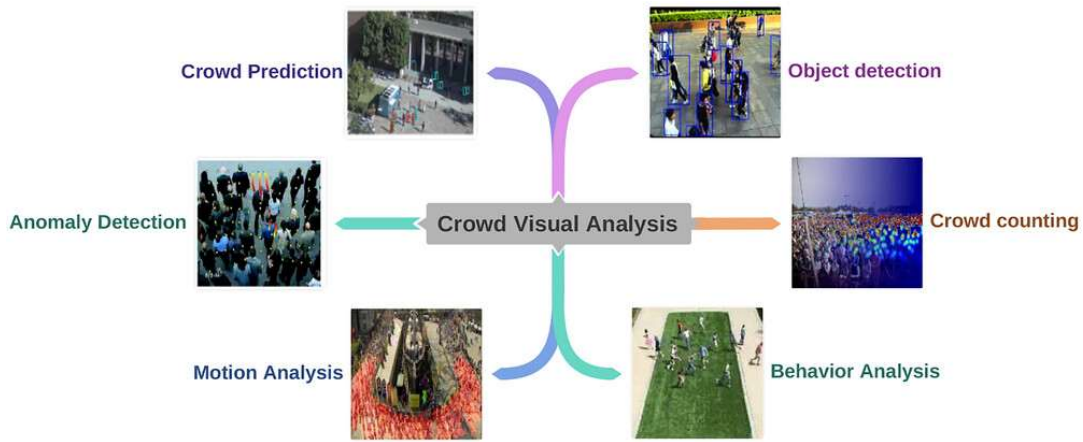

**Figure 1.3: Major applications of Crowd Scene Analysis**

This research utilizes computer vision and deep learning approaches to explore two sub fields of crowd scene analysis: *crowd anomaly detection, and crowd counting*. We propose a comprehensive solution for crowd counting by leveraging advanced deep learning and machine learning techniques. Furthermore, we explore the use of crowd anomaly detection in real-time applications.

*Crowd counting* plays a crucial role in counting the number of individuals within a specific area. This information is vital for crowd management and ensuring compliance with safety regulations. By utilizing advanced deep learning models and machine learning algorithms, we aim to develop an accurate and efficient crowd counting system that can handle various crowd scenarios.

Our work also focuses on ***crowd anomaly detection*** in addition to crowd counting. This field aims to identify unusual or abnormal behaviors or events within a crowd in real-time. By employing computer vision and deep learning techniques, we seek to develop algorithms that can detect and classify anomalous activities or events, enabling prompt attention and intervention when necessary.

### 1.1.1 Crowd Counting (CC):

***Crowd counting is a method used to accurately determine the number of individuals present in a crowded scene*** (***as shown in*** **Figure 1.4**) [5].

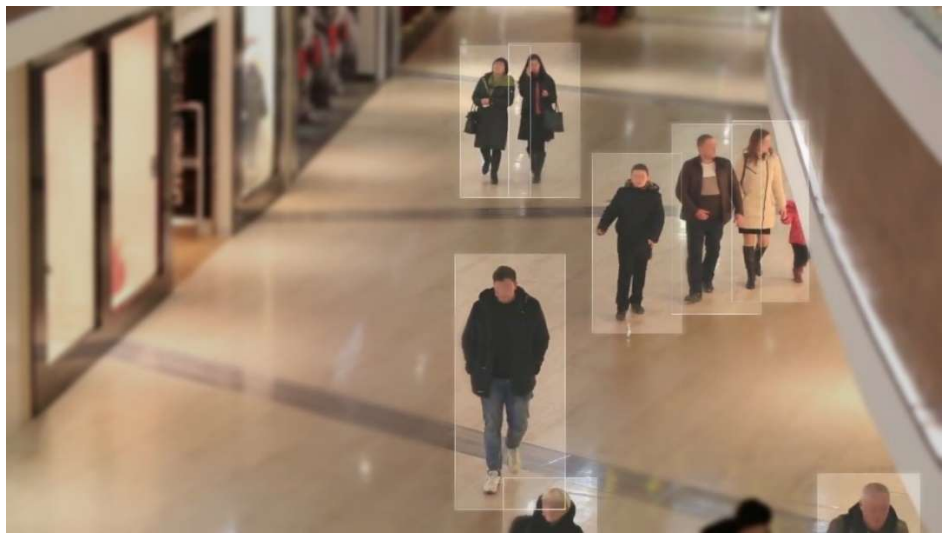

**Figure 1.4: A general picture depicting crowd counting**

It involves the utilization of computer vision and deep learning techniques to analyze videos or images and count the attendees in a crowd. It has many real-world applications such as crowd management, public safety, urban design and infrastructure, retail analytics, and event management. By counting an accurate number of people in a crowd we may be able to manage and plan our events. It is also helpful for transportation management in such a way that we can plan routes and schedules of public transport by analyzing the number of people traveling on a specific route. In case of any disaster, we can use the videos to analyze the areas where the number of people were high and required immediate evacuation. Political gatherings can also be analyzed for security purposes, estimating the fan followers of political figures, and identifying any signs of violence.

### 1.1.2 Crowd Anomaly Detection (CAD):

*Crowd anomaly detection is the process of identifying and flagging unusual or abnormal behaviors within a crowded scene.*

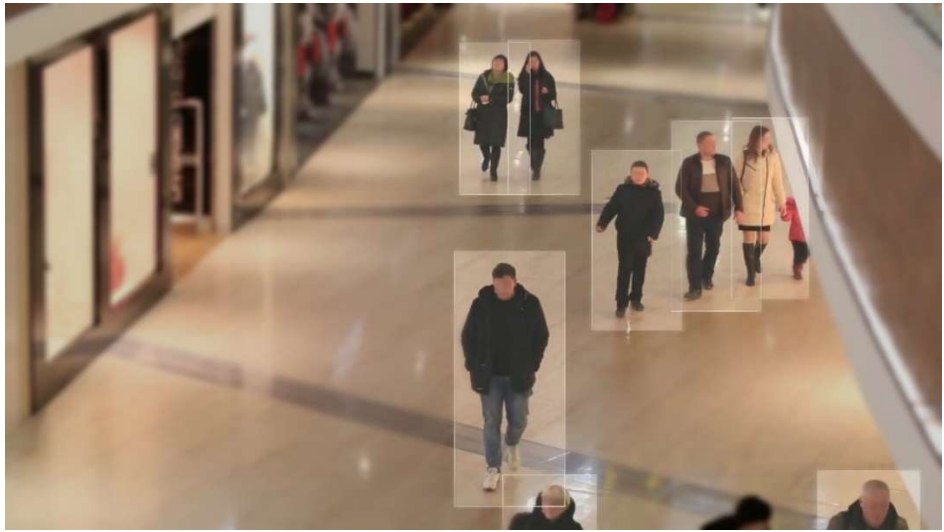

**Figure 1.5: A general picture depicting crowd anomaly detection**

It involves the use of advanced technologies such as computer vision, machine learning, and data analysis to analyze crowd dynamics and detect potential threats or disruptive incidents. Various types of anomalies (*as shown in* **Figure 1.5**) [6] can be detected, including the presence of weapons, fights, stampedes, panic situations, violence, looting, mob mentality, trampling, and crush incidents. It plays a vital role in different real-life applications by ensuring public safety and security. It can be utilized in busy places, transit hubs, or locations with vital infrastructure to identify safety risks, spot possible threats, or keep an eye out for indications of criminal activity. It also allows us to take quick response and intervention by alerting security officers or authorities, lowering risks, and preserving order as a result.

## 1.2 Motivation:

From 1959 to 1999, the world population has been doubled from 3 billion to 6 billion and it is increasing at the rate of 0.91% per annum. As predicted by the UN in November 2022, it has reached up to 8 billion. It is expected to increase up to 9 billion by 2037, and it will reach about 10 billion by the end of 2058. Due to this swift increase in world population, personal lifestyles and basic facilities are also changing. To achieve a better quality of life, people are increasingly moving towards urban areas, resulting in overcrowded cities. This urbanization trend is accompanied by a rise in social activities and gatherings such as political events, concerts, religious festivals, and sports festivals. However, the limited availability of resources and space in these cities has raised serious concerns regarding security and surveillance. Analyzing crowd behaviors has become a challenging task in order to address these concerns and take necessary steps to ensure public safety. This situation has motivated us to conduct research in the field of crowd scene analysis, with a particular focus on crowd counting and crowd anomaly detection. By accurately estimating crowd sizes and detecting abnormal behaviors within crowded scenes, it becomes possible to effectively manage and mitigate potential risks associated with overcrowded environments.

### 1.2.1 Crowd Counting:

Researchers from different fields conducted research in crowd scene analysis. Transport planners performed crowd analysis to study the number of people coming in and out of the public station within a specific time. This is helpful for them to study the crowd dynamics for planning and scheduling specific routes. Business managers of different supermarkets studied crowd analysis to analyze the behavior and volume of customers coming in and out of the supermarket. By estimating the number of customers, they can manage the store space for specific products and forecast the sale and purchase of different items. Event-organizing companies are trying their best to plan and manage successful events by estimating the volume of crowds based on events held in past years.

However, autonomous crowd counting remains a challenging task due to its unique characteristics and complex nature. ***Occlusions, complex backgrounds, non-uniform density, and scale in-variation are still major difficulties in accurate and efficient crowd counting. Another specific challenge in crowd counting is the use of annotated or labeled data for the training of the algorithm and it is a quite laborious task to collect and annotate a large amount of data.*** Most of the existing methods [7], [8], [9], [10], [11], [12] are based on supervised training. Following the recent works proposed in [13], the M-CNN architecture can efficiently perform crowd counting by capturing the information at multi-scales due to its multi-column structure.

Motivated by this, we modified the architecture of M-CNN by increasing number of columns from three to five. As a result, we propose a method that utilizes the modified M-CNN architecture in combination with *self-supervised training via distribution matching*. The additional columns enable the network to learn and extract features at different scales, leading to improved crowd counting performance. To further enhance the performance and generalization of our modified M-CNN architecture, we leverage self-supervised training. Self-supervised learning is a technique that utilizes unlabeled data to train a model to perform a specific task.

**1.2.2  Crowd Anomaly Detection:**

To this end, research and development are going on in the field of deep learning, and computer vision. Security managers from different security agencies study crowd behaviors to analyze and detect the anomalous behavior of people gathered in crowded places such as concerts, religious gatherings, political gatherings, conferences, etc. By studying anomalous behavior in crowds, they can analyze and detect suspicious activity, criminals, and emergencies. By doing so, they can provide information to relevant personnel to make decisions and take prompt action.

Transport engineers and planners study the crowd behaviors to analyze and identify the abnormal activities at public transport places such as airports, railway stations, public parking, and bus stations They can manage the crowd flow by analyzing the movement of passengers at specific terminals, ticket issuance booths, and platforms. By doing so, they can also analyze the different behaviors such as overcrowding, panic, and abnormal movement patterns. This helps in efficient crowd management, and the detection of abnormal activities, and unattended baggage.

It is necessary to propose an automatic solution for anomaly detection in crowds. Detecting anomalies in a crowd is always a strenuous task due to its complexity. Various environmental factors such as fog, camera viewpoints, scene complexity, quick movements, and weather conditions such as rain, snow, and fog. A specific challenge in this task is to detect abnormal behavior in situations where they appear instantaneously for a short duration.

Inspired by recent works in crowd anomaly detection, particularly the work proposed by [14], [15], we presented a model based on the combination of CNN and LSTM by leveraging the VGG-19 architecture and image-net weights for crowd anomaly detection. Before the fully connected layer, we added wide residual blocks to detect anomalies in the crowd. The proposed model can detect abnormal behavior with improved performance especially for the short time occurring anomalies.

## 1.3 Crowd Counting:

With the advent of the field of artificial intelligence, machine learning, and computer vision, crowd counting is gaining popularity and becoming an important problem. It has several applications in daily life such as political gatherings, disaster management, event management, transport route planning, security, surveillance, and crowd management [11].

It is always a challenging task to count the number of people in images of congested scenes. Crowd counting can be performed by different methods such as manual counting and traditional methods including the counting with the help of face detection. Manual counting is always a laborious and difficult task with high chances of mistakes such as over-or-under counting. Normally, highly dense crowd images contain more than a thousand people.

Crowd counting always faces different challenges (*as shown in* **Figure 1.6**). such as ***occlusions*** [16], ***low resolution*** [17], ***complex backgrounds***, and ***overlapping features*** [18] making the traditional methods ineffective [11]. Recent advances in deep learning have helped to overcome these problems by the use of advanced deep learning techniques. Different deep learning techniques can be used for crowd counting tasks such as Convolutional Neural

Networks (CNN) [19], Recurrent Neural Networks (RNN), Fully Convolutional Networks (FCN) [20], Generative Adversarial Networks (GAN) [21], [22], and Attention Mechanism [23].

A number of algorithms have been proposed for crowd-counting. In image-based crowd counting, CNN-based approaches have become popular due to their efficiency and accuracy in computer vision related tasks. Earlier, CNN-based approaches performed crowd counting based on object detection. The advanced approaches improved the accuracy of counting methods based on density maps. CNN-based approaches can also be categorized on the basis of their architectures. Basic CNN models consist of convolutional layers, pooling layers, and fully connected layers. Commonly used architectures [24] of CNN involve LeNet, AlexNet, VGGNet, DenseNet, ResNet, GoogleNet, and MobileNet.

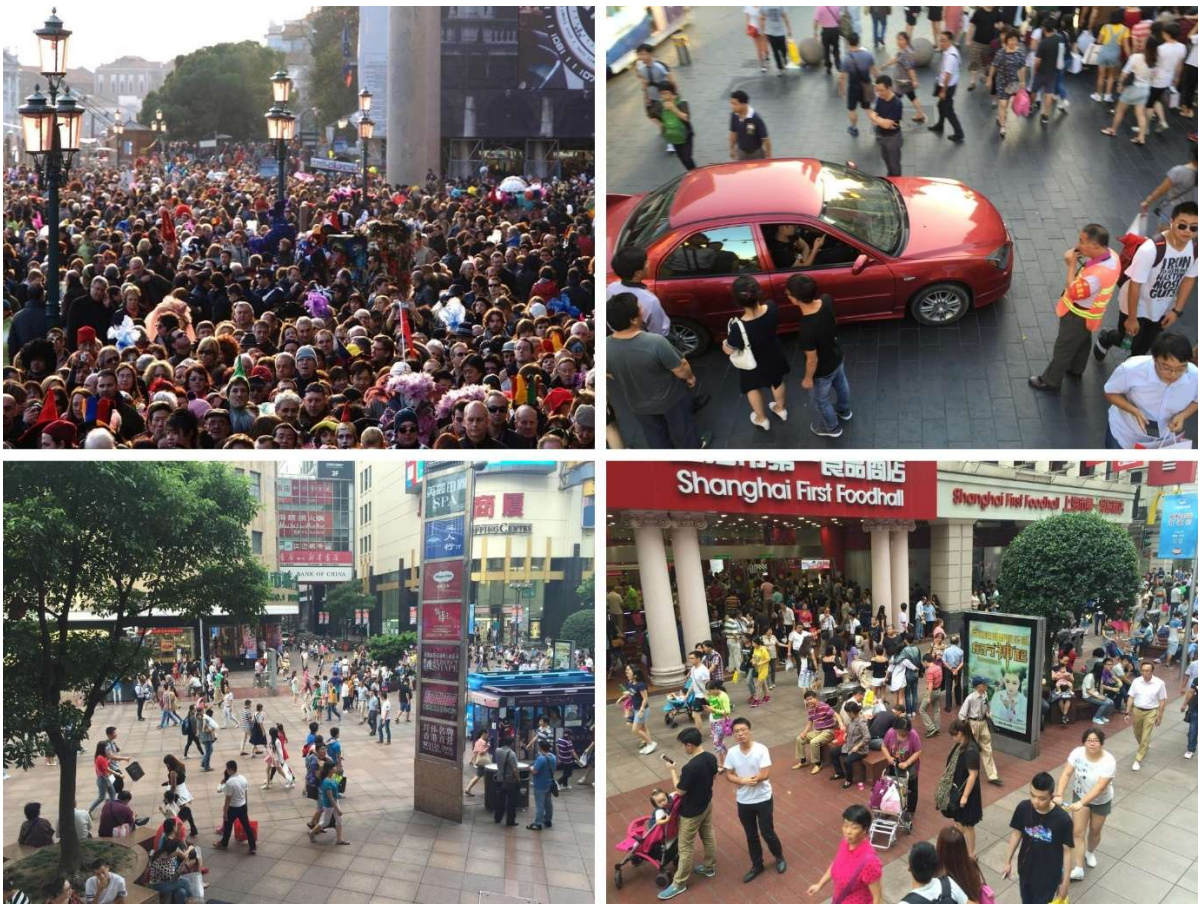

**Figure 1.6: Different challenges in crowd counting; (a) occluded scene, (b) low resolution image, (c) non-uniform distribution, (d) complex background images**

Crowd counting can be done by using two different methods; detection-based counting method and density-based counting methods (*as shown in* **Figure 1.7**) [25]. Both methods take input in the same form (e.g., images) and give output in different forms. Detection-based counting methods gives an estimated number of people in an image, while density-based methods give output in the form of density maps. Density maps consist of the number of people per square meter. The density map is then subjected to integration to estimate the exact number of people in an image. Among both of these two methods, density-based crowd counting methods outperform due to their robustness to occlusions, high accuracy for highly dense crowded scenes, and pixel-wise detailed information about how individuals are distributed throughout an image.

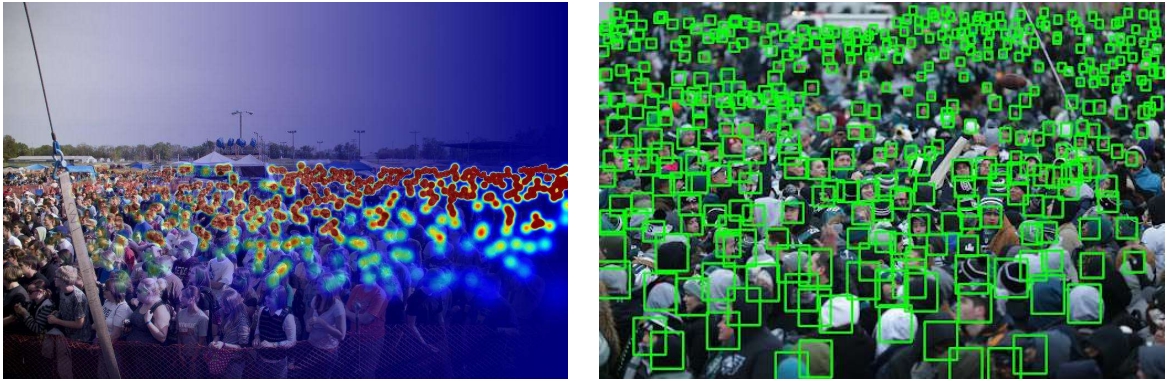

**Figure 1.7 : Different Methods for Crowd counting; (a) Density map-based crowd estimation, (b) Detection based Crowd estimation**

Crowd counting using density-based approaches involves estimating crowd count by analyzing the density of individuals in a scene. These methods focus on creating density maps or heat- maps that represent the distribution of people within an image. Techniques like kernel density estimation or Gaussian mixture models are used to estimate the density of people in different regions or pixels of an image. The total crowd count is obtained by summing up the estimated densities. Convolutional neural networks (CNNs) are also used to learn the mapping between input images and their corresponding density maps. CNNs are trained on a large dataset of annotated images, paired with ground truth density maps.

The accuracy of crowd counting heavily relies on the quality of the density maps, which requires careful calibration and tuning of density estimation techniques or training CNN models with high-quality annotated datasets. The input photos and the related ground truth density maps are used to train the model. These density maps represent the density of people in different areas of the image. It is crucial that the sizes of the input images and density maps are consistent for accurate training. To generate density maps, we feed the trained input images into our model, which then predicts the density map based on the image input. The density map highlights the populated areas within the image, providing a visual representation of the crowd density.

To evaluate the performance of our model, we compare the predicted D-maps with the ground truth images. This allows us to estimate the difference or error between the predicted and actual crowd density. Various loss formulas can be employed for this purpose, including the Mean Absolute Error (MAE), Mean Square Error (MSE), Grid Average Mean Error (GAME), Patch Mean Absolute Error (PMAE), Patch Mean Squared Error (PMSE), Mean Pixel Level Absolute Error (MPAE), Peak Signal to Noise Ratio (PSNR), and Structural Similarity Index (SSIM) [26]. The calculated error is then utilized in the process of back-propagation, where it is propagated back through the model to update the weights and biases. This iterative process of updating the model's parameters allows it to learn and improve its predictions over time. Different optimizer such as ADAM, Nadam, and RMSprop are used to optimize the training process [27]. These optimizers adjust the learning rate and update the model's parameters to minimize the error and enhance the performance of the model during training.

Researchers from different fields around the world are entering the field of crowd behavior analysis, contributing towards the research in crowd counting and resulting in the plethora of crowd counting methods. That ranges from simple neural network architectures CNN [8], Crowd-Net [11], [28], RESNET-Crowd [29] to multi-column/multi-scale neural network architectures McML [30], MS-CNN [31], M-CNN [13] to advanced architectures such as P2PNet [32], STANet [33], DSNet [34], CRANet [35], STDNet [36], MSPNet [37], CSRNet [38], CLRNet [39], SSDA [40], TAFNet [41], MAN [42], and D-ConvNET [43], [44].

These all methods were based on CNN and outperformed with the support of supervised training. Supervised training always requires annotated (labeled) data. It is practically infeasible to create and manage a large database of annotated images of people in crowded scenes.

Most of the algorithms are working on the support of annotated data with the currently avail- able data sets that are small in size with a limited range of scenes. It highlights the need for better training algorithms based on large data sets with a broad range of scenarios. It is leading towards the development of methods with the support of publicly available unannotated data sets. To address these challenges, researchers have explored alternative approaches that require less reliance on annotated data. One such approach is unsupervised learning, where the model learns from unlabeled data without the need for explicit annotations. Unsupervised methods aim to discover patterns and structures within the data to infer the crowd count or density [45], [46], [47]. Traditional methods of unsupervised learning involve autoencoders. Auto-encoders are neural networks consisting of two sub-parts; encoder and decoder. Both parts work together to encode the input data and learn meaningful representations through the bottleneck layer without labeled data. Auto-encoders have main applications in the areas of unsupervised learning, anomaly detection, and dimensionality reduction. Generative Adversarial Networks (GAN) [21] Deep Belief Networks [8], and Variational auto-encoders [48] are other techniques of unsupervised learning.

Another approach is weakly supervised learning, which utilizes less precise annotations or weak labels instead of full annotations. For example, instead of precisely labeling each individual in an image, weak labels might indicate the presence or absence of people in a particular region. This reduces the annotation effort and makes it more feasible to create a large dataset [49], [50] Furthermore, researchers have also presented self-supervised learning for crowd counting task [40], [51], [52]

Self-supervised learning is the most recent framework for learning un-annotated images; it allows models to learn in an unsupervised manner by performing a pre-text task to provide artificial supervisory signals from the input data. Self-supervised learning makes use of some auxiliary tasks or objectives that can be automatically formed from the input data rather than relying on annotated data. For example; in solving jumbled scenes, a model is trained in a self-supervised manner to predict the missing part within a scene by learning the spatial relationship between different objects within an image [53]. Other tasks of self-supervision include in-painting [54], generating gray scale to color images and vice versa [55], [13] and prediction of angle of rotation [56], [57].

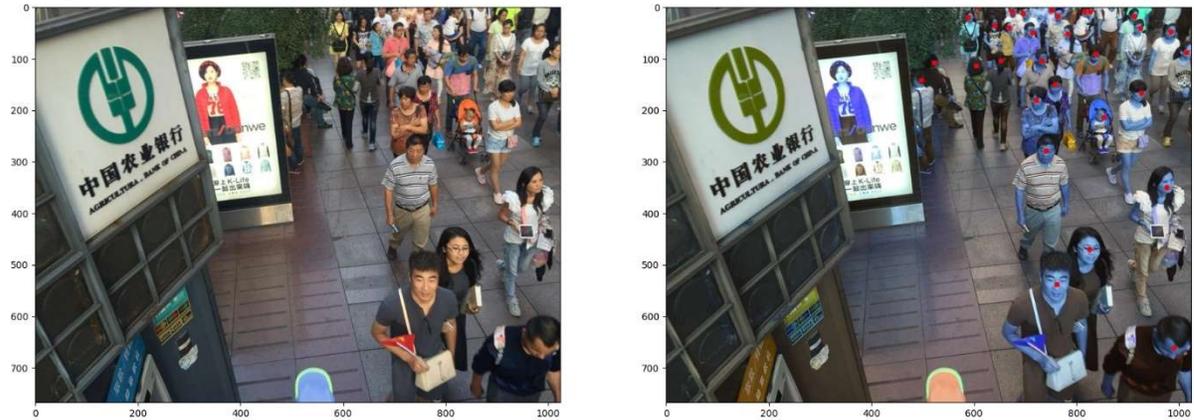

**Figure 1.8: Annotated vs. un-annotated crowd image**

**Figure 1.8** shows and image taken from ShanghaiTech Crowd Dataset and performed annotation to show difference between annotated vs un-annotated crowd image. The counting context task is a popular pretext problem used in self-supervised learning for crowd counting. In this challenge, an image is divided into smaller parts or patches, and the number of persons in each patch has to be predicted. Without knowing the overall population count of the image, the model is trained to estimate the local population count within each patch. By completing this task, the model has the ability to detect subtle changes in crowd density locally and can use this understanding to extrapolate population counts from previously unseen images. The model is trained to estimate the relative positions of individuals within an image in the relative position estimation task, which is another pretext problem. In order to accomplish this task, the model must be able to determine patterns of crowd density as well as understand the spatial relationships among individuals.

A number of algorithms have been proposed for crowd-counting. In image-based crowd counting, CNN-based approaches have become popular due to their efficiency and accuracy in computer vision related tasks. Earlier, CNN-based approaches performed crowd counting based on object detection. The advanced approaches improved the accuracy of counting methods based on density maps.

CNN [19] are among the popular choices for crowd-counting tasks because of their suitability with image processing techniques. CNN Consists of multiple layers including an input layer, feature extraction, pooling layers, fully connected layer, and output layers. Convolutional layers of CNN are trained in such a way that it can detect local features, such as edges and corners, and then combine these features to form higher-level representations of the input image. That is why; CNN is highly effective for the detection and extraction of features from images that contain complex patterns such as crowds. CNN can be used to estimate the number of people in a crowded scene after the training process. For this purpose, an input image will be fed to the trained CNN. It produces a density map [58] that represents the spatial distribution of peoples in the image. By adding the values in the density map, the number of people can be estimated in a crowded scene/image.

We proposed an extended M-CNN by increasing its number of branches from 3-branches to 5-branches. This enables the model to capture information both at multi-scales (both and local and global context). Local context means the analysis of individual persons or small groups and global context means the entire crowd. By considering multiple scales, the model can capture the density of groups across different sizes and adapt to varying object scales within the crowd. This makes it possible to understand population dynamics and spatial patterns on a more comprehensive scale.

Each column of M-CNN consists of filters of different sizes. With small size filters such as 3x3, and 5x5, the receptive field is limited to a small region around each pixel. This indicates that the filter mainly collects data from the pixels near local neighborhood. Similarly, 7x7 filter size makes the receptive field limited to medium sizes regions around each pixel. While the filter size 9x9 and 11x11 makes the model to capture information at broader context by incorporating large receptive fields. These large size filters are useful for managing crowd photos with significant density changes, capturing global context, and taking into account interactions between individuals at various scales. Proposed model is trained by leveraging self-supervised learning in such a way that the supervision signal is generated form the input data itself, rather than relying on manually labeled dataset.

## 1.4 Crowd Anomaly Detection:

A process to automatically detect physical aggression, verbal abuse, or aggressive behavior by means of different technologies such as machine learning, artificial intelligence, and deep learning may be referred to as violence detection. Similarly, crowd anomaly detection refers to analyzing real-time videos to detect the violence within the crowd or large gatherings. By employing this technology, we may enhance public security and safety by automatically recognizing and monitoring aggressive behavior, fights, and other types of violence in real-time. Different sources of data such as videos, audio, and physical patterns can be used for anomaly detection. Videos can be used to identify different patterns, physical altercations, aggressive postures, or the presence of weapons. However, audio can be used to detect sounds indicative of violence, such as shouting or screams.

Crowd violence detection technology has important consequences for public safety, especially in crowded events, protests, and high-crime regions. These technologies can assist rapid response and intervention by detecting possible violent occurrences early and informing authorities or security personnel, therefore reducing harm and maintaining order.

Before the emergence of deep learning techniques, traditional approaches were commonly employed for crowd anomaly detection. These methods included optical flow, background subtraction, crowd density estimation, statistical approaches, hybrid approaches, and trajectory-based approaches. However, these traditional methods presented certain limitations and challenges. One of the main issues with traditional approaches was limited scalability. These methods often struggled to handle large-scale crowd scenes with a high number of individuals. Additionally, traditional methods required manual tuning of various parameters, which made them less adapt- able and time-consuming to implement. Furthermore, traditional approaches had limitations in terms of feature extraction. They relied on handcrafted features, which might not capture the complex and diverse characteristics of abnormal crowd behavior accurately. This limited the effectiveness of anomaly detection systems based on traditional methods [59], [60].

To address these challenges, the advancement of deep learning techniques has significantly im- proved crowd anomaly detection. Deep learning models, such as Convolutional Neural Net- works (CNNs), Recurrent Neural Networks (RNNs), Long Short-Term Memory (LSTM), Autoencoders, and Variational Autoencoders (VAEs), have played a crucial role in overcoming the limitations of traditional approaches. Deep learning models have shown remarkable capabilities in automatically learning and extracting features from raw data. CNNs, for example, excel at capturing spatial characteristics in images or videos, making them well-suited for crowd anomaly detection tasks. RNNs and LSTM models have proven effective in modeling temporal dependencies and capturing sequential patterns, which are crucial for detecting anomalous crowd behaviors over time.

Autoencoders and VAEs have also contributed to the advancement of crowd anomaly detection by enabling unsupervised learning and feature extraction. These models can learn representations of normal crowd behavior and detect deviations from these learned representations, facilitating anomaly detection without the need for explicit labels. The adoption of deep learning techniques has helped overcome the limitations of traditional approaches by providing scalability, adaptability, and improved feature extraction capabilities. Deep learning models have revolutionized crowd anomaly detection, enabling more accurate and robust detection of abnormal crowd behaviors in various real-world applications.

## 1.5 Significance of Crowd Scene Analysis:

Crowd scene analysis plays an important role in our daily life applications.

1. ***Venue Management:*** Crowd counting techniques can be used in venues such as stadiums, concert halls, or conference centers to estimate the number of people present. This information is valuable for capacity planning, ensuring compliance with safety regulations, and optimizing resource allocation.

2. ***Public Transportation:*** Crowd counting in transportation hubs like train stations, airports, or bus terminals helps in assessing passenger loads, optimizing scheduling, and improving crowd flow management. It enables transportation authorities to anticipate peak times and allocate resources accordingly, ensuring a smooth and efficient travel experience.

3. ***Retail Analytics:*** Crowd counting plays a significant role in retail settings by providing insights into footfall data. Retailers can analyze customer traffic patterns, identify peak shopping hours, and make informed decisions regarding staffing, store layout optimization, and marketing strategies.

4. ***Political Gatherings:*** Crowd scene analysis plays an important role in the planning and management of political gatherings. By analyzing the crowd videos and footage:

- Party management can analyze the fan following of political leaders or any political figure. This information can be useful in providing information about the popularity of any political figure and can be used by the media for new purposes.
- They can manage the public space by forecasting the people gathering for the next political meeting. They can identify the crowd flow, congested areas, and size of the crowd. This can help management to take necessary steps to avoid overcrowding and to control the entrance and exit of people at different points.
- Crowd analysis can help them to analyze the crowd behavior, identify the risks and abnormal movements, and forward the information to security agencies to take any necessary action promptly.

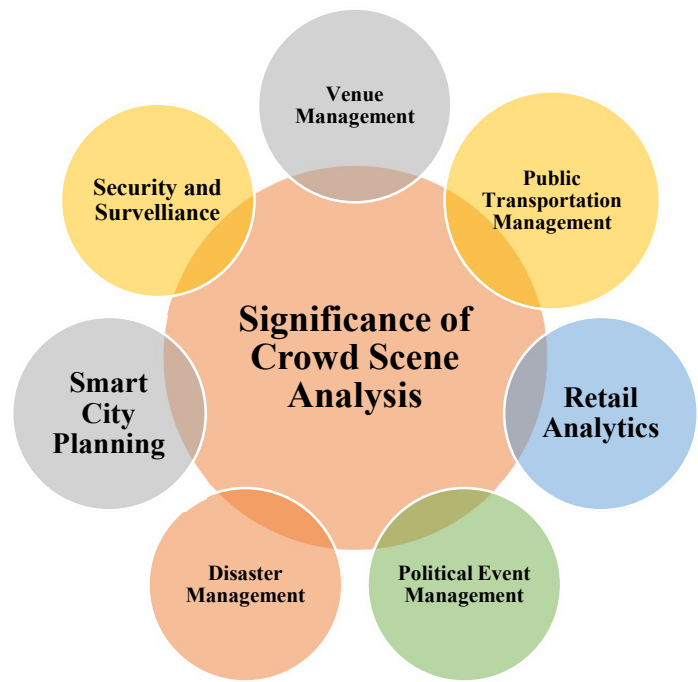

**Figure 1.9: Significance of Crowd Scene Analysis**

5. ***Emergency and Disaster Management:*** Crowd anomaly detection can aid in emergency response situations. By identifying abnormal crowd behaviors during emergencies or disasters, authorities can quickly identify areas of concern, allocate resources efficiently, and coordinate evacuation plans, ultimately saving lives and minimizing the impact of the situation.

6. ***Smart City Planning:*** Crowd scene analysis, including crowd counting and anomaly detection, is valuable for urban planners in designing smart cities. By understanding crowd dynamics, city planners can optimize the design of public spaces, transportation systems, and infrastructure to improve efficiency, safety, and overall quality of life for residents.

7. ***Security and Surveillance:*** Crowd anomaly detection is crucial for ensuring public safety in crowded areas. By monitoring crowd behavior, abnormal or potentially dangerous activities can be detected, enabling security personnel to intervene promptly. This application is particularly relevant in areas such as airports, train stations, or public gatherings where security threats or unlawful behavior may arise.

## 1.6 Challenges:

### 1.6.1 Crowd Counting:

Crowd counting is always a challenging task due to the dynamic nature of crowd scenes. As highlighted in *section 1.3*, *occlusions, density variation, non-uniform backgrounds, different weathers, lighting conditions, and camera viewpoints* are among the major challenges for crowd counting tasks.

1. Density variation and occlusions are almost linked with each other. Density variation always makes it difficult to accurately count people. Density can range from sparse to extremely dense. The high density of people always leads to occlusions. When two or more peoples collides with each other, it results in the occluded scene making the scene more complicated.

2. Lighting conditions, such as shadows, reflections, or low light, can affect visibility, making it harder to detect and count people accurately.

3. Dynamic crowd behavior, such as movement, changes, and diverse behaviors, also affects the accuracy of crowd counting.

4. The perspective and viewpoint of the camera can impact the appearance and size of individuals in the crowd. This variation in perspective makes it challenging to consistently estimate the number of people in different areas of the scene.

For an effective and efficient crowd density estimation, it is necessary to address these challenges by the implementation of robust algorithms that can handle these challenges.

**1.6.2  Crowd Anomaly Detection:**

The dynamic and complicated nature of crowd behavior presents different challenges for crowd anomaly recognition. Lighting conditions, environmental conditions, the presence of unexpected objects, short-occurring abnormal instances, and scalability are major challenges to the detection of anomalies.

1. Dynamic environmental factors, such as changes in lighting, weather, or unexpected objects, can also influence crowd behavior, making it difficult to distinguish between true anomalies and benign environmental changes.

2. Another challenge for the reliable detection of anomalies is the scalability of the crowd. As crowd anomaly detection systems need to handle large-scale crowds, which can be computationally intensive and require efficient algorithms and infrastructure.

Crowd anomaly detection faces several challenges due to the complex and dynamic nature of crowd behavior. These include the lack of labeled anomaly data, high variability in crowd behavior, and the need for contextual understanding.

## 1.7 Aims and Objectives:

The main goal of this research is centered around the development of robust algorithms for the visual analysis of crowds. To achieve this, a combination of state-of-the-art technologies such as deep learning and computer vision will be employed, enabling the creation of intelligent solutions for crowd surveillance and monitoring.

While significant progress has been made in feature engineering for various computer vision applications, this study aims to explore and evaluate different features and their combinations specifically for crowd counting, and crowd behavior analysis. By examining and assessing these diverse features, the research aims to enhance the accuracy and effectiveness of crowd analysis algorithms.

1. The proposed system should be able to give an accurate count of individuals present in a crowded scene.
2. The proposed system should be able to detect the abnormal behavior of people present in the crowd.

Following of the objectives have been proposed to achieve this aim:

1. To propose an automatic framework based on the integration of computer vision and deep learning approaches for real-time crowd counting.
2. To propose an automatic solution for anomaly detection in a crowded scene.

## 1.8 Research Question:

The aforementioned goals and motivations led to the following research questions

1. How to count the number of people present in a crowded scene?

2. How to detect abnormal behavior of people present in a crowded gathering?

## 1.9 Contributions:

### 1.9.1 Crowd Counting:

Particularly our contributions to crowd counting can be summarized below:

1. To eliminate the use of annotated data, we proposed a novel method for crowd counting based on completely self–supervised learning without using a single annotated image.

2. We perform a series of augmentation processes (such as rotation, cropping, and flipping) on input images to increase the size of the data set. The size of the data set is expanded by making a little change in images. It reduces the risk of overfitting and increases the robustness to handle diverse conditions such as changing light conditions, different weather conditions, and varying camera angles.

3. To the best of our knowledge, we are the first to propose M-CNN with five convolutional branches in parallel. Each branch has a corresponding filter size that it uses to capture the useful features at various scales. It enables models to deal with different challenges such as occlusions, low resolutions, non-uniform density of images, and complex backgrounds. The concatenation of multiple branches increases the robustness of the model.

### 1.9.2 Crowd Anomaly Detection:

For crowd anomaly detection, our contributions for crowd counting can be summarized as below:

1. A model based on the combination of Convolutional Neural Network (CNN) and Long-Short Term Memory (LSTM) is proposed for feature extraction at the multi-frame level by leveraging the VGG-19 architecture.

2. A series of data augmentation processes are performed to increase the variety and volume of the dataset. It also makes the model robust towards generalized scenarios.

3. To improve the efficiency of the model for short-occurring anomalies in videos, we proposed a Wide Residual block (WRB). The proposed method detects anomalies at individual input frame level, instead of detection at the video level.

## 1.10 Stakeholders:

- This research work aims to enhance crowd management practices and safety for pedestrians, public gatherings, political gatherings, religious festivals, and gatherings for entertainment.

- Research questions and innovative approaches are explored to improve crowd control strategies and flow management.

- Findings can be used by crowd management agencies to optimize operations and mitigate safety risks.

- Law enforcement agencies can gain a deeper understanding of crowd behavior and develop strategies to respond to incidents.

- Research provides insights into crowd psychology, communication, and situational factors impacting crowd behavior.

- The thesis contributes to crowd management by providing practical guidance and evidence- based insights.

Overall, the thesis aims to contribute to the field of crowd management and provide practical guidance for crowd management and law enforcement agencies. By offering evidence-based insights and innovative methodologies, the research can assist in the efficient and effective management of crowds of pedestrians, ultimately enhancing public safety and security.

## 1.11 Outline of the Thesis:

On the basis of the intended research questions, the whole work is divided into seven chapters. The thesis is structured as below:

- Chapter one begins by introducing the crowd, types of crowd scene analysis, and explaining its importance in different applications of the real-time field. It provides a detailed introductory overview of crowd counting and anomaly detection in crowds. It outlines the motivation behind research carried out in the field of crowd scene analysis. It addresses the major challenges that must be overcome in order to meet the research goals. It concludes by summarizing the research contributions.

- Chapter two highlights the comprehensive review of available research on the research topic. It investigates prior studies, theories, models, and methodologies that have been already proposed in the subjected discipline. This chapter also establishes the groundwork by highlighting the research gaps in current knowledge, and by emphasizing key findings and comments from prior publications.

- Chapter three highlights the proposed framework for self-supervised crowd counting using M-CNN. This chapter is divided into three parts. The first part highlights the theoretical background of crowd counting and outlines a detailed overview of deep learning, neural networks, data augmentation, convolutional neural networks, self-supervised training, multi-column CNN, and density maps. The second part of this chapter describes the end-to-end proposed framework. It explains the experimental design, data collection, and pre-processing techniques. The part outlines the specific algorithms, or techniques that will be used to address the research questions.

- Chapter four presents and discusses the implementation as well as experimental results for crowd counting. The first part of this chapter highlights the implementation details, evaluation metrics, and data sets. However, the second part includes the experimental results, performance evaluations, comparisons with existing methods, and any statistical analyses conducted. The chapter interprets the results, draws conclusions, and discusses the implications of the research questions. It also highlights any limitations or challenges encountered during the experiments.

- Chapter five highlights the proposed framework for crowd anomaly detection using conv- LSTM. This chapter is also divided into three parts. The first part highlights the theoretical background of crowd anomaly detection and outlines the VGG-19 architecture of CNN, LSTM, Wide residual blocks, and different pre-processing techniques. The second part of this chapter describes the end-to-end proposed framework. It explains the experimental design, data collection, and pre-processing techniques. The part outlines the specific algorithms, or techniques that will be used to address the research questions. Implementation details, evaluation metrics, and data sets are presented in the third part of this chapter.

- Chapter six presents and discusses the implementation as well as experimental results for crowd anomaly detection. The first part of this chapter highlights the implementation details, evaluation metrics, and data sets. It includes the performance evaluations, comparisons with existing methods, and any statistical analyses conducted. The chapter interprets the results, draws conclusions, and discusses the implications of the research questions. It also highlights any limitations or challenges encountered during the experiments.

- Chapter seven presents the conclusions and future work by summarizing the main findings and contributions of the research.

# CHAPTER TWO: LITERATURE REVIEW

A brief review of the current approaches along with their shortcomings for crowd scene analysis is presented in this chapter. The background review is divided into two parts, with each section focused on crowd counting and abnormal behavior detection in the crowd. This detailed discussion will highlight the limitations and research gaps to find research direction for future work.

## 2.1 Crowd Counting:

Analyzing the size of a crowd or counting the number of attendees gathered in a crowd can help us plan and manage an event effectively. Public safety and security can be ensured with a minimum number of security personnel which results in cost reduction of security arrangements. A large number of existing methods are reviewed in this literature review. This literature review is defined in two sub-sections as below:

1. Approaches based on density maps.
2. Approaches employing deep learning methods.

The classic approaches to crowd counting—which often employ detection-based and regression- based methods—are covered in the first half of this article. These techniques rely on object detection algorithms to identify and count individuals within a crowd. Various traditional approaches and their key features are explored, providing a foundation for understanding the evolution of crowd-counting techniques.

The second part of this section delves into methods based on deep learning for crowd counting. This part reviews the key CNN-based approaches, high- lighting their strengths, limitations, and advancements. The discussion encompasses different CNN architectures, loss functions, and training strategies utilized in crowd counting. Finally, this section concludes by discussing the distinctive features and contributions of our proposed model. It highlights how our approach builds upon the existing literature and addresses the limitations of previous methods.

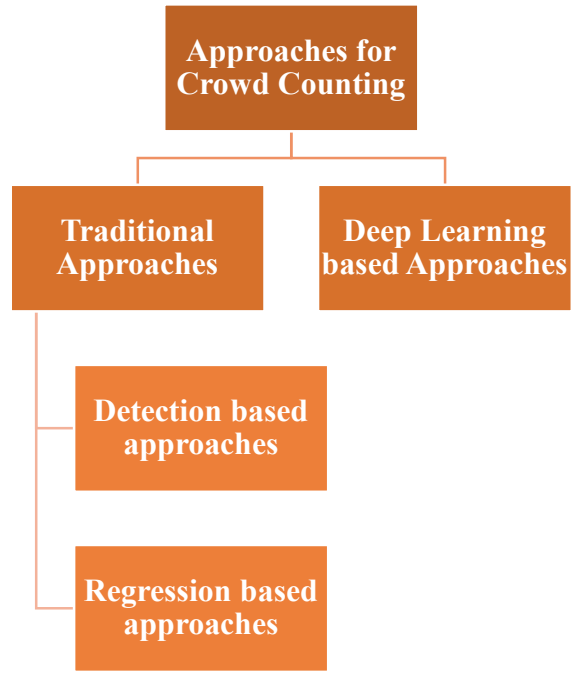

**Figure 2.1: Different Approaches for Crowd Counting**

### 2.1.1 Traditional Approaches:

#### A. *Detection Based Approaches:*

Earlier approaches to crowd counting relied primarily on detection-based methods. Detection models are used to create annotations on the persons/objects present in any image, these an- notated images are then divided into small patches, and the final count is represented by the number of annotations in an image. These methods involved detecting specific hand-crafted features such as body parts, heads, or shoulders using various image processing techniques. For instance, Lin et al. [61] utilized a detection-based approach that involved detecting body parts in crowded scenes. Tuzel et al. [62] employed a similar approach, focusing on the detection of pedestrians. Topkaya et al. [63] used a sliding window technique to detect body appearances for crowd counting. Felzenszwalb et al. [64] and Li et al. [65] also employed detection-based approaches for estimating the number of people in crowds. After being discovered, these features that were manually generated were used by different machine learning models, including Support Vector Machines (SVM), Neural Networks (NN), Random Forest, Linear Regression (LR), and Ridge Regression (RR), to estimate the crowd size [26]. The number of people in a crowd was estimated with these models by exploiting the features that were observed and adopting various regression methodologies.

Many cutting-edge models have been put forth lately to boost the precision and effectiveness of crowd-counting operations. One such model is LSC-CNN, proposed by Sam et al. [66]. LSC-CNN improves training at higher resolutions by using a new competition that takes loss function. It introduces the concept of pseudo-bounding boxes to refine the crowd count estimation. This approach has demonstrated improved performance in accurately localizing and counting people in crowded scenes. Another notable model is PSDDN, proposed by Liu et al. [67]. By adopting smaller predicted boxes, PSDDN adjusts the pseudo-bounding boxes employing a Faster-RCNN-based detector. To initialize the pseudo-bounding boxes, it likewise takes advantage of the nearest neighbor distance. The PSDDN can also detect and quantify human heads, thereby improving the precision of crowd counting.

Crowd-SDNet, proposed by Wang et al. [68], is another noteworthy model. Through the use of point-level annotations, Crowd-SDNet employs a self-training approach for counting crowds. For precise crowd counting, the system employs a Faster-RCNN-based detector to estimate the crowd size. All of these methods, including LSC-CNN, PSDDN, and Crowd-SDNet, have demonstrated superior performance in both detection and crowd-counting tasks.

They outperform traditional approaches by leveraging advanced techniques and incorporating innovatory detectors. The accuracy and efficiency of estimating crowd sizes and locations have been enhanced as a result of the use of these models, which have made major contributions to the discipline of crowd counting.

The performance of the previously mentioned models tends to decline when applied to crowded scenes. These models excel in estimating crowd counts for sparse crowds but struggle to accurately count individuals in congested scenes. Congested scenes present difficulties like occlusions, low resolution, foreshortening, and varying perspectives, making it challenging to detect and localize individuals accurately. To overcome these challenges, regression-based methods have gained prominence in crowd-counting tasks. These methods involve regressing the total count of people directly from an image, without relying on explicit detection of individuals or body parts. Regression-based approaches leverage the image to estimate the crowd count, making them more robust to occlusions and other problems.

**B. Regression Based Approaches:**

Regression-based approaches generally employ different regression models and generally work on extracting different features from input images and utilizing them to estimate the number of attendees in a crowd [42]. Generally, these methods use different annotations such as bounding boxes, polygon annotations, and point marks for counting purposes. These methodologies typically involve two steps: feature extraction and crowd count estimation. In the feature extraction step, relevant features, such as gradients, foregrounds, and textures, are extracted from the input images. These global features capture important information about the crowd. Subsequently, various regression models, including ridge regression (RR), Markov Random Field (MRF), Bayesian regression, and Random Forest, are utilized to estimate the crowd count. These regression models leverage the extracted global features as inputs and learn to map them to the corresponding crowd counts. By working on global features and employing regression models, these approaches aim to accurately estimate the crowd counts at either the image-wise or patch-wise level.

An approach based on linear mapping between features and the corresponding items in density maps within local regions was proposed by [9]. This approach was motivated by the observation that disregarding the salience of objects in local regions can lead to inaccurate crowd count estimates. The authors introduced a novel evaluation metric called Maximum Excess SubArray (MESA) to assess the performance of crowd-counting algorithms. MESA provides a comprehensive measure of the accuracy and robustness of crowd-counting methods, taking into account variations in crowd density and spatial distribution.

Idrees et al. proposed a different approach for crowd counting by combining the scale-invariant feature transform (SIFT) and Fourier transform (FT) [7]. This approach aims to leverage multiple sources of data within a single crowd image. SIFT is used to handle challenges such as low-confidence head detection and the repetition of texture features in crowded scenes. By extracting and analyzing SIFT features, the approach improves the reliability of crowd count estimation. Furthermore, frequency-domain analysis using the Fourier transform is used to precisely estimate the crowd size. This technique makes use of the spatial frequency components contained in the crowd image.

In the realm of crowd counting, both of these methods make original contributions. Lemitsky et al.'s approach addresses the importance of considering local saliency for accurate crowd count estimation and introduces a novel evaluation metric. On the other hand, Idrees et. al.'s approach combines SIFT and Fourier transform to exploit multiple data sources and enhance the reliability of crowd count estimation. These approaches showcase the diverse range of techniques and methodologies employed in the pursuit of accurate and robust crowd counting.

Learning the most effective linear mapping for crowd counting can be a difficult task, especially with increasing amounts and complexity of data. In order to address this challenge, Pham et al. addressed this challenge by proposing a novel algorithm called Crowd-FOREST [69]. This approach takes a different approach by relying on a non-linear mapping function to learn the features and their relevant items in density maps. By employing this non-linear mapping function, Crowd-FOREST aims to overcome the limitations of linear mapping and improve the accuracy of crowd counting.

In recent years, several regression-based models such as P2P-Net, ACSCP, DIHM, and MRA- CNN have been proposed for crowd-counting tasks. These models have shown superior performance in terms of counting accuracy. However, one limitation of these models is their inability to effectively localize crowd patterns and generate a comprehensive crowd distribution in the image. This localization challenge hinders their ability to provide detailed insights into the spatial distribution of crowds within an image.

While regression-based methods have been effective in addressing challenges such as occluded scenes, limited pixels per image, varying perspectives, and complex backgrounds that were encountered in detection-based approaches, they have shown limitations in handling dense crowd images. The poor performance of regression-based methods in dense crowd scenarios can be attributed to the reliance on low-level handcrafted features. These handcrafted features may not capture the complex patterns and variations present in dense crowd images, leading to reduced accuracy in crowd counting. As a result, alternative approaches that can effectively capture and represent high-level features and contextual information are being explored to overcome these limitations and improve the performance of crowd-counting methods in dense crowd environments.

## 2.1.2 Deep Learning based Approaches:

With the advent of research in the field of deep learning, especially in the domains of computer vision, different researchers worked on deep learning-based approaches for crowd counting and improved the performance of traditional methods. In deep learning-based methods, CNN-based methods are highly effective and have emerged as state-of-the-art approaches for crowd- counting-related tasks [26] because they can capture both local and global features. These methods outperformed the regression-based methods for two reasons;

1. Accurate counting estimation of the crowd, and
2. Generating the distribution of the crowd by capturing the spatial information which can both estimate the crowd and generate the distribution.

A deep CNN based model is proposed for crowd counting and crowd density by Zhang et. al. Counting and density maps are generated on the basis of switchable training schemes [8]. An extended version of Deep CNN was proposed by Marsden et. al. to improve the overall performance and enhance the model generalization by reducing the redundant data in training samples of augmented data set [10]. Marsden et. al. also extended their work by proposing RESNET- Crowd [29] to perform three different tasks; crowd counting, abnormal behavior detection, and crowd classification based on density. A research gap observed that the model was evaluated on the basis of a small data set of 100 images. This results in high values of MAE and MSE, and it can be reduced by evaluating the model on large-size data sets. However, the vast number of parameters and large storage space that plague the current

RESNET-Crowd continue to limit their practical applicability by requiring high storage and computer resources.

In order to resolve this issue Ding et. al., proposed RESNET-based deep recursive CNN [70]. The network has the ability to capture statistical regularities in the context of the population is improved by the recursive structure, which makes the network deeper while maintaining the same number of parameters. The proposed model was also trained on the large-size data set, generated from the video monitoring data of the Beijing bus terminals. Switch CNN based on the combination of three CNN modules and a switch-based classifier was presented by Sam et al. Switch-CNN [71] uses a classifier (switch) to choose the best CNN regressor for an input crowd scene patch out of three regressors with various CNN architectures.

Different CNN methods proposed for crowd counting have been found more accurate as com- pared to conventional techniques that rely on handcrafted features. Another author proposed multi-resolution attention-based CNN for crowd counting and density-based crowd classification. An attention mechanism is proposed to focus on the head region using a score map and suppress non-head regions [72].

Boominathan et. al. proposed Crowd–Net [11] to estimate the crowd density with a combination of deep and shallow fully convolutional networks. This combination helps to extract both high-level and low-level features. High-level features include body and face detectors, low- level features include blob detectors. The data set of less than 100 images is used for training purposes. Testing is also performed on the basis of only a single data set (e.g., UCSD-CC-50). MAE values are very high due to the small size of the data set, and it can be improved by training and testing on large sizes of data sets.

Multi-Column CNN (M-CNN) [13] was proposed by Zhang et. al. to estimate the size of the crowd using a single image. M-CNN consists of three convolutional columns connected in parallel. Columns are having filters and kernels of different sizes, and are used for compensation of perspective distortion. The final output from three different columns is concatenated to get out- put in the form of a density map. Finally, the output is estimated from density maps. The model was evaluated on transfer learning and a performance gap was observed between fine-tuning the whole network and fine-tuning the last two layers of the network. This performance gap was due to the limited training data of UCF-CC-50 and can be resolved by using another data set of large size. Further, this model was trained using the supervised data. It can be improved by using self- supervised learning. MAE and MSE can also be reduced by assigning attention mechanisms. CNN performance and effectiveness were improved by introducing gradient boosting and selective sampling. This has increased the crowd-counting accuracy and reduced the processing time.

Occlusion is the major challenge that we are facing in crowd counting. To resolve this challenge, an object detection-based solution with the combination of CNN and Markov Random Field was proposed by Han et. al. [73]. Dense crowd images were divided into overlapping patches; features of these patches were extracted by CNN. Adjacent patches had having high correlation with crowd count due to overlapping portions. High correlation and MRF are used to smooth the crowd count. A switching architecture Switch-CNN [71] based on variation in density within a single image was proposed by Sam et. al. It automatically switches appropriate regressors for particular population patches. These methods, however, are very scene-specific and might not be suitable when analyzing multiple scenes at once.

CSR-net with a combination of two different components at the front end and back end was proposed by Y. Li et. al. Features were extracted at the front end with the help of 2D CNN and pooling operations were replaced at the back end by using dilated CNN to obtain larger reception fields. Significant drops in the values of MAE and MSE were observed as compared to the previous state-of-the-art approaches [74]. Four different CSR-net models by varying dilation rates from 1 to 4 are evaluated. The dilation rate can be further increased to capture more context and global information by enlarging the receptive field of the network.

Crowd counting frequently employs density estimation, whereas traditional approaches use pixel-by-pixel regression without explicitly accounting for pixel dependency. Independent pixel- wise predictions can therefore be erratic and noisy. In order to resolve this issue, RA-net was proposed by Zhang et. al. [75] self-attention mechanism was used to capture short-range as well as long-range inter dependencies between pixels, where local self-attention (LSA) and global self-attention (GSA) are used to refer to these implementations, respectively. LSA and GSA can be used in a relation module to produce more useful aggregated feature representations.

Sindagi et al. [76] suggested a different CMTL strategy based on the integration of two CNN units. The second CNN is recommended to extract contextual information and physical relationships within the crowd, while the first CNN module serves to estimate the crowd's volume. A robust solution for estimating the number of people in a crowded scene is presented by lever- aging the multi-scale multi-column CNN [31]. Proposed models showed improved accuracy by offering a reliable approach for precise crowd counting in in variety of settings. This model performs more effectively than all single-scale techniques and greatly advances computer vision and deep learning.

Hu. et al [77] proposed a CNN-based architecture for crowd counting of mid-to-high level crowd images by using Conv-NET to extract features. The model was trained using point-wise annotated images, to count the number of people in a single image. Another author proposed Count-NET [78] based on a combination of CNN and auxiliary learning mechanisms. The proposed approach outperforms the existing approaches because it can adapt to different crowd scenarios by leveraging end-to-end training. Auxiliary learning makes this approach a significant contribution to crowd-counting tasks.

An approach based on the combination of multiple CNN mechanisms was presented by Kumagai et al. [79] to deal with various crowd scenarios by leveraging an ensemble mechanism for intelligent selection of CNN. The selection of CNN is based on specific crowd appearances such as density, scale, and occlusion pattern. All of these above-mentioned algorithms [77], [78], [79] outperformed traditional methods by predicting the number of people, heads or shoulders present in images. These methods outperformed all regression and detection-based traditional methods, and relied on CNN-based detection mechanisms rather than density maps. To achieve more remarkable results, different studies were also presented based on density maps using CNN architectures.

## 2.2 Crowd Anomaly Detection:

A large group of people who gather for a particular purpose, such as going to a sporting event, a music concert, or other similar events, is referred to as a crowd. Conversely, an anomaly can be defined as an abnormal entity/behavior that deviates or disturbs the normal behavior within a crowd; frequently standing out as an outlier in the broader distribution. This literature review goes into recent studies on crowd anomaly detection, offering light on relevant research on this subject.

### 2.2.1 Traditional Approaches:

Violent actions can be detected on the basis of different features such as motion analysis, behavior analysis, blood/flame detection, skin/blood pattern analysis, or by analyzing the audio-visual features [80]. Pang et. al. [81] proposed a method to detect violence in videos by fusing audio- visual features. The proposed model consists of neural networks with three additional modules; an attention block, a fusion block, and a transfer learning block. The attention module emphasizes relevant features, the fusion module integrates visual and audio features using bi-linear pooling, and the mutual learning module allows the model to learn from different architectures. Similarly, [82], [83], [84] proposed models for violence detection based on the fusion of audio-visual features. Another author proposed a model based on skin and blood detection for violence behavior detection [85]. Indeed, it is true that violence detection studies that specifically focus on fusing visual and audio information are relatively scarce in the existing literature. While there have been advancements in violence detection techniques using either visual or audio features separately [81]. It is a common challenge in violence detection that audio information is often not available in most surveillance footage. By focusing solely on visual cues, there is a plethora of methods that can still provide meaningful insights and contribute to the effective detection of violence in videos.

Various feature descriptors have been proposed in the field of video analysis to acquire insights into the features of video frames. These descriptors typically focus on low-level characteristics, such as shape features that capture specific image information. One of the most common low- level features is key point detection. It attempts to detect points in an image that are unique from their surrounds [86].

Scale Invariant Feature Transform (SIFT) based methodology was proposed by Ojha et.al. [87] by focusing on three important features such as dynamic appearances, anomalies in spatial eccentricity, and change in crowd flow based on inherent eccentricity. The proposed model was designed to detect anomalies both in structured, and semi-structured crowd environments. Anomalies are detected by calculating the vector positions of objects within the picture frame. These positions are then compared with calculated pixel points across the sequence of frames. The distance measurement of an object served as an indicator of its speed, potentially identifying it as an anomaly within the crowd. SIFT [88] features are invariant to rotation or changes in scale. They extract local features from video frames, providing robust representations that are unaffected by variations in size or orientation. SURF (Speeded-Up Robust Features) [89] is a feature extraction technique that is based on the principles of SIFT (Scale-Invariant Feature Transform) and allows efficient feature extraction from video frames. It is specifically developed to meet the computational efficiency needs of video processing tasks. SURF characteristics are robust to rotation, scaling, and affine transformations, making them outstanding for video analysis. The integral image approximation technique is used by the algorithm to gain computational efficiency. When compared to traditional methods, this technique allows for the rapid computing of picture features, resulting in a considerable speedup.

Another author proposed the BRISK feature descriptor [90], formed by combining the FAST key point detector and the BRIEF descriptor. The robustness of both components is transmitted through to this descriptor, making it robust to rotation, scale changes, affine transformations, and noise. It provides a binary representation of features that is suitable for tasks like object detection, image matching, and tracking. Oriented FAST and Rotated BRIEF (ORB) [91] is a feature descriptor that combines the FAST key point detector and the BRIEF descriptor. It is renowned for its ability to be rotation and scale invariant while still being resilient to affine transformations and noise. Overall, ORB is a reliable and efficient feature descriptor that can effectively extract unique information from images, making it a useful tool in the field of computer vision.

Yang et. al [92] proposed an approach based on Bag-of-Words (BoW) that depicts video frames as a single vector, which reduces sensitivity to noise and unpredictability in the data. This is accomplished by creating a vocabulary of visual terms from the frames and picking the most representative features for classification. The initial stage of the BoW method is to extract local characteristics from video frames, such as SIFT or SURF. These features record important information about the visual content of the frames. Following that, a clustering method, such as k-means, is used to group related features together and generate a vocabulary of visual words.

Hassner et. al. [93] proposed a framework based on a combination of Support vector machines (SVM) and violent flow descriptor (ViF) for anomaly detection crowd scenes. The Violent Flows (ViF) descriptor is created by analyzing flow-vector magnitudes over time, which are then used to classify frames as violent or non-violent using a linear support vector machine (SVM). The proposed model is evaluated by introducing a novel data set, which contains real-world surveillance videos. This method, ViF descriptor, and dataset all help to advance surveillance video analysis and improve the identification of violence in crowded scenarios.

Inspired by ViF descriptors proposed by [93], Gao. et. al proposed Oriented Violent Flow descriptors to detect violent flows in crowded scenes by leveraging the orientation information into the feature vectors of flow magnitudes. The integration of orientation information in the flow magnitude feature vectors helps to a more robust and informative representation, ultimately boosting the effectiveness of violence detection in crowded settings.

### 2.2.2 Deep Learning based Approaches:

With the recent advances in computer vision, machine learning, and deep learning, the task of detecting violent actions has received a lot of interest from the research community. It uses video analysis to detect and highlight abnormal actions. Numerous surveys have been done to provide detailed overviews of the existing literature and approach to violent behavior detection. A framework based on a social multiple-instance learning (MIL) was proposed by Lin. et. al. [94] that incorporates a dual-branch network to capture the dynamic interaction among groups, individuals, and the environment, resulting in an attentive spatial-temporal feature representation. MIL was used to overcome the limitations of limited training data for anomalous samples by employing a social force map to simulate behavior interactions and give previous knowledge to improve anomaly detection. Self-attention modules were also included to improve the discriminate capacity of the C3D network's retrieved spatial-temporal properties.

Pawar et. al. [95] proposed a deep learning approach by utilizing unsupervised learning techniques within a one-class classification paradigm. The model effectively learns and extracts meaningful features from video data, enabling accurate anomaly detection. The effectiveness of the proposed model has been demonstrated on benchmarked anomaly detection data sets, showing significant results in terms of equal error rate, area under the curve, and detection time. Both [94], [95] haven't extensively explored the detection of rare and novel anomalies in crowd behavior. Anomalies that occur infrequently or exhibit unique patterns can be challenging for any anomaly detection model.

Mohan et. al. [96] proposed a framework based on the combination of CNN and Principal Component Analysis (PCA) for anomaly detection in video-based surveillance. Principal Com- ponent Analysis (PCA) and Support Vector Machine (SVM) were employed for tagging and classifying abnormal activities and anomalies. The effectiveness of the proposed model was evaluated on three data sets: Avenue, UCSD, and UMN, with the Avenue data set yielding better results.

Mehmood et. al. [97] make a valuable contribution to the field of anomaly detection by presenting a methodology based on 2D-CNN. The proposed model was pre-trained for selected spatial and temporal streams, with a focus on gathering different types of anomalies, evaluated on three different data sets, and achieved remarkable accuracy of more than 98% to 99.5%. Another author Feng et. al. [98] proposed a model based on the combination of PCA-Net and GMM to serve a dual purpose by detecting anomalies, and performing model building to identify patterns in video events. The authors conducted assessments using frame-level, pixel-level, and object-level measures on two data sets. The results demonstrated that the proposed model successfully extracted features from 3D gradients, providing valuable insights into the motion and appearance characteristics of anomalies. All of these above methods used image-based data for training and testing and have not evaluated these methods on video-based data sets.

Pustokhina et. al. [99] proposed an anomaly detection framework to detect anomalies on pedestrian walkways for the safety of pedestrians. The model was based on Masked Region-based CNN, supported by DenseNET 169 model. This method allowed for the correct identification and classification of abnormalities, which contributed to improved pedestrian pathway safety. Several pre-processing operations were also performed on data set including, videos to frame conversion, and image de-noising for noise removal. to improve the quality of frames extracted from videos. The proposed method demonstrated excellent results and accurately detected the anomalies. However, the accuracy of this model can be affected due to the de-noising of im- ages. Because, this de-noising process may produce blurred images and this blurriness can potentially result in the loss of accurate features, which may affect the overall performance of the model.

Hinami et. al. [100] proposed CNN based framework for detecting and counting abnormal events. Convolutional neural networks (CNN's) have demonstrated promising outcomes in visual concept learning. However, properly utilizing CNN's for anomalous event identification remains a difficulty, owing to the environment-dependent nature of anomaly detection. This problem is solved in this study through the combination of a general CNN model with environment-dependent anomaly detectors.

Another author Liu et. al. [101] proposed an approach by leveraging the spatial and motion constraints to predict the accurate prediction of future frames of normal events. The suggested method is thoroughly tested on both a toy data set and publicly available data sets. The experimental results show that the technique is beneficial in terms of robustness to uncertainty in normal events and sensitivity to abnormal events.

## 2.3  Summary:

This chapter reviews crowd analysis literature, focusing on crowd counting and crowd anomaly detection. It highlights both traditional and deep learning-based approaches for these tasks. The majority of research in crowd counting focuses on counting and estimating crowd density using supervised learning mechanisms. Supervised learning algorithms require labeled data, which is collected from large datasets of crowd images for accurate counts. This manual annotation process is time-consuming and resource-intensive, and can be subjective and error-prone. Consistent and reliable annotations are crucial for effective crowd counting models. Traditional counting methods face challenges such as occlusions, complex backgrounds, and non-uniform density, making it difficult to accurately count individuals in crowd images. Therefore, consistent and reliable annotations are essential for effective crowd counting models.

To address these gaps, we have proposed self-supervised training with multi-column convolutional neural network. By utilizing a combination of self-supervised learning and M-CNN, our approach offers an innovative and effective solution for crowd counting. This research fills the gap in the literature by demonstrating the effectiveness of multi-column CNN and a novel self-supervised learning technique to improve crowd counting accuracy by leveraging the un- annotated data.

However, most of the research in the subject of crowd anomaly identification focuses clip- level prediction over frame-level prediction, indicating that certain frames might not be taken into account while analyzing violent action. Recent research has demonstrated the successful capture of both spatial and temporal data in videos with the combination of a Long Short-Term Memory (LSTM) and a Convolutional Neural Network (CNN). Enhancing CNN-LSTM model for violence detection is our goal, based on this understanding.

To address these gaps, we proposed a spatio-temporal model based on combination of VGG-19 and LSTM. We modified the architecture of VGG19 by replacing the fully connected layers layer with Dense residual block for spatial features extraction. Temporal features are extracted by the use of LSTM block. This approach aims to improve the performance of the model by predicting violent activity for each input frame, which is particularly beneficial for videos that contain short-duration violent actions.

# CHAPTER THREE: CROWD COUNTING METHODOLOGY

The focus of this chapter is to highlight the proposed framework for crowd counting using various deep learning techniques, particularly Multi-Column Convolutional Neural Networks. M-CNN is proposed in this research to capture multi-scale features by focusing on head and shoulder locations. M-CNN consists of multiple branches of convolutional neural networks with various filter sizes. It enables the network to model density maps associated with heads of different scales. Using filters with larger receptive fields—which perform better at capturing the complexity of density maps that correspond to larger heads—is made possible through this strategy.

Crowd Counting can be done in two different configurations by using convolutional neural networks; ***1. Detection based approach, 2. Density-based approach. Detection-based approaches*** work by taking an input image and giving a crowd count by counting the number of heads in the image. On the other hand, ***density-based approaches*** regress the density map, give the total count of people per square meter, and subsequently calculate the headcount through integration. The density-based approach is proposed in this research work for the following reasons:

1. First of all, density maps enable us to measure the number of people in the image more precisely as we have a more accurate depiction of the crowd distribution. As compared to detection-based approaches, this approach generates an accurate and meaningful estimation.

2. Second, a density map gives you additional possibilities for analyzing the dynamics of crowds.   It makes it possible to analyze and interpret crowd behavior in more detail, including pinpointing high- and low-density areas and recognizing spatial patterns within the crowd.

3. As density maps provide spatial information about crowd images, they offer potential future applications such as crowd flow analysis, and crowd anomaly detection.

## 3.1 Theoretical Background

### 3. Data Augmentation

Deep learning methods are data-hungry paradigms and always require a large amount of data for better training and prediction purposes. Data always acts as fuel for deep learning methods, so the efficiency of any deep learning model and the size of the data set are equally crucial. Deep learning models are always designed in such a way that they can learn useful patterns and information from data used for training purposes. The larger the training data, the better a model can perform and respond to unseen/new samples. A large amount of training data can enhance the performance of the model and result in an accurate prediction for real-world scenarios [102]. Contrasting, deep learning models don't work best on a limited amount of data and cannot respond to unseen/new samples and it may result in over-fitting (***as shown in* Figure 3.*1***). In this scenario, a model can work very well on trained data, but cannot provide an accurate prediction for new/unseen samples [103].

It is noteworthy that both the variety and quality of data are just as significant as the amount of data. The model can learn a broader spectrum of trends and become better at handling a variety of situations with the help of a varied data set that covers a range of conditions. *Fig 3.1(a)* clearly depicts that the training error starts to increase, and indicates that the model has overfit to the training data, resulting in poor performance on the testing set compared to the training set. On the other hand, *Fig 3.1(b)* is showing an ideal behavior of training and testing of data, as both training and testing errors are simultaneously decreasing. This shows that the model is effectively learning the underlying patterns and features present in the data without overfitting. The model demonstrates good generalization ability, performing well on both the training and testing sets. It always remains a challenging and time-consuming task to collect and label a large amount of data for better training of a deep learning method [104].

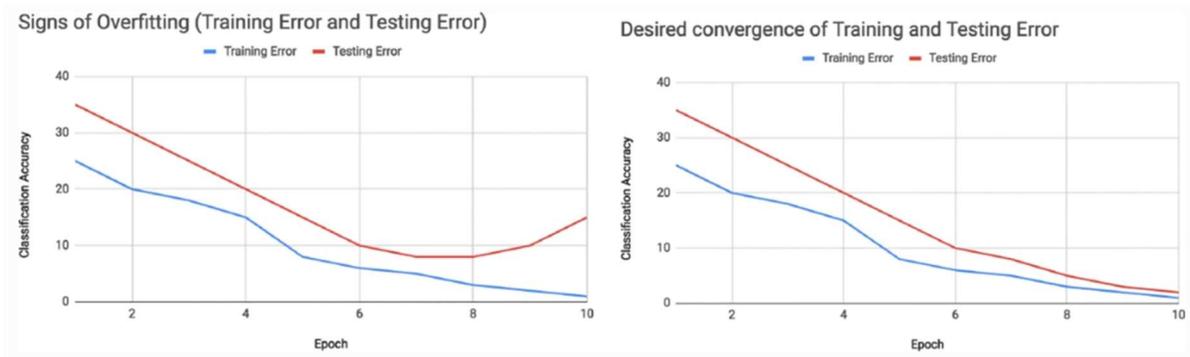

**Figure 3.1: (a) Training and testing error due to over-fitting, (b) Training and testing error with ideal data set size**

To alleviate this problem, data augmentation has emerged as one of the most popular techniques to increase the size of quality data by artificially expanding the training data for better training of deep learning models [102]. With the growing demand for large-scale data sets, data augmentation offers a practical solution for increasing the effective size of the training data without the need for additional data collection. In the context of deep learning, data augmentation can be referred to as a technique to generate/expand training data in an artificial manner by applying different operations and modifications to the original available data set. We may apply different modifications such as cropping, rotation, flipping, translations, scaling, and many others (*as shown in* **Figure 3.***2*) [105].

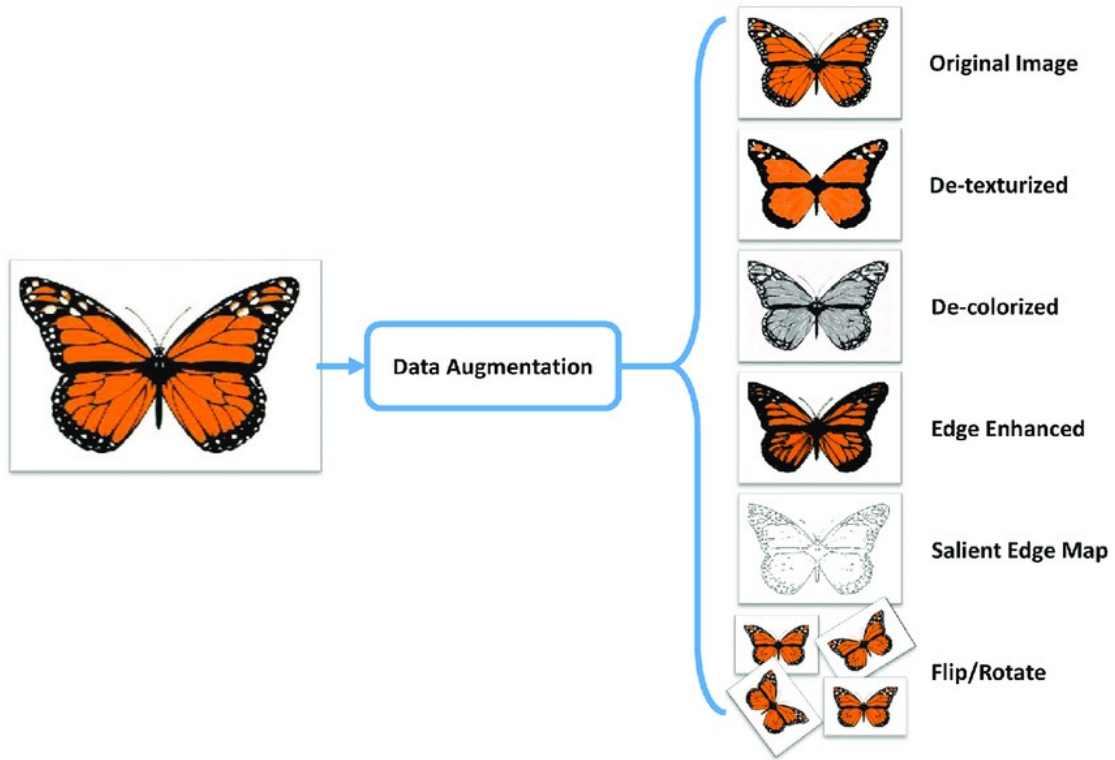

**Figure 3.2: Different Approaches for Data Augmentation**

We may also change the background color, and contracts, or may add the noise or distortion in an original image to increase the size of the data set. Data augmentation cannot only be applied to image data, it may also apply to any sort of data such as videos, audio, and textual data. Another way to increase the size of the data set is to generate synthetic data with the help of machine learning models such as generative adversarial networks (GAN) [21] and Variational autoencoders (VAE) [48]. These models may generate new samples that resemble the attributes of the original data set by analyzing the basic distribution of the original data [103]. Depending on the intended purpose and kind of problems, data augmentation can be categorized (*as shown in* **Figure 3.3**) into two types of approaches 1), Traditional approaches, and 2). Machine learning based approaches [103], [106].

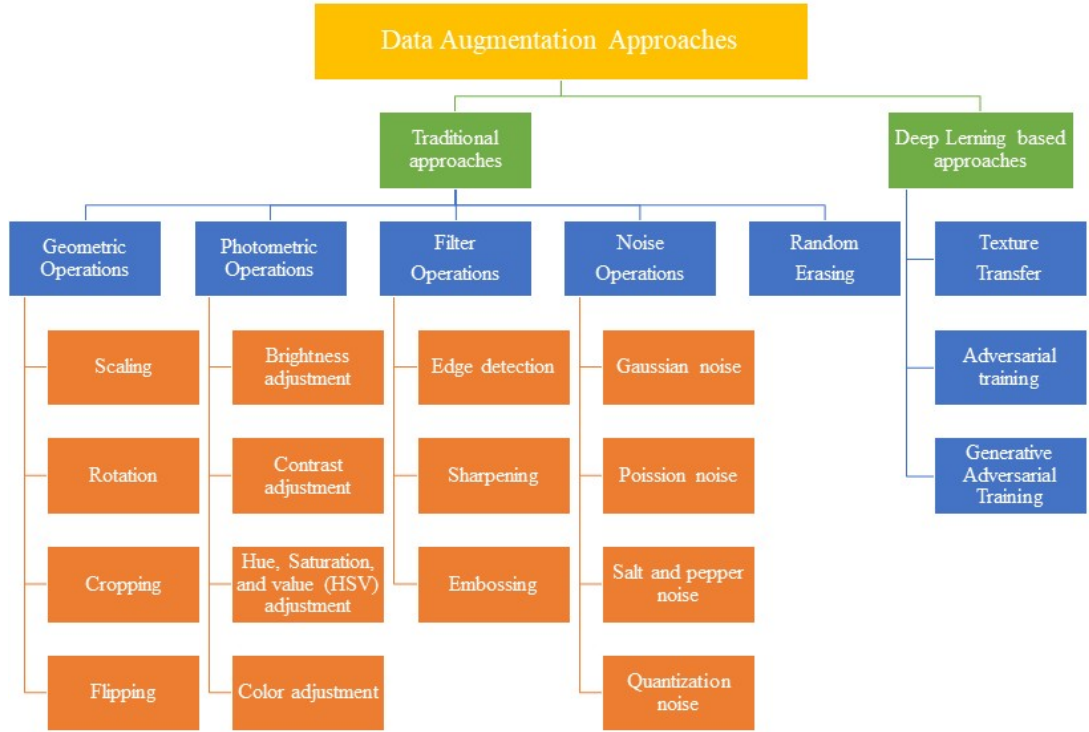

Figure 3.3: Traditional vs. Deep Learning based Approaches for Data Augmentation

### 3.1.1.1 Traditional Approaches:

### A. Geometric Operations:

Geometric transformations are a powerful data augmentation approach that is frequently used in machine learning and image processing. Different spatial transformations are carried out on original images, including as rotations, translations, scaling, and flipping. These transformations help to enhance the training data set and boost the model's capacity to generalize to other positional variations.

- **Rotation:** The image is rotated by a specified angle in this transformation.
- **Translation:** The image is translated by shifting it horizontally and vertically.
- **Scaling:** This operation causes an image's dimensions to be either larger or smaller.

- **Flipping:** This type of transformation flips the image either in the horizontal or vertical direction.

By randomly applying these geometric operations, the model learns to be invariant to mirror reflections, numerous orientations, various positions of objects in an image, and different sizes. These operations also make the model more robust to unseen/general scenarios. Paste Crowd pictures of above four transformations.

### B. *Photometric Operations:*

The main goal of these transformations is to add variations and improve the generalization capacity of the model by modifying the pixel values and attributes of images. A variety of photometric operations are available to modify the contrast, color, and lighting of images. These operations generally include brightness adjustment, contrast adjustment, HSV adjustment, and color adjustment.

- **Brightness Adjustment:** Scaling pixel values to alter overall image brightness by adding or subtracting the constant value to each pixel.
- **Contrast Adjustment:** Changing the pixel intensities between bright and dark regions to highlight or diminish image details.
- **Color Adjustment:** Changing the values of individual color channels to introduce variety and learn robust features.
- **HSV Adjustment:** This is the process of separating color information into hue, saturation, and value components.

Photo-metric transformations enhance the training dataset's diversity by introducing lighting, color, and contrast variations. This helps deep learning models recognize and generalize to real-world scenarios, making them more robust and adaptable to image appearance variations, resulting in improved performance and generalization on unseen data

## C. Filter Operations:

Data augmentation can also be done by means of different filters. Filters (also known as kernels) can be defined as a small size of matrices, generally convolved with an image. This convolution operation adjusts the values of pixels supplied in the filter. •

- **Edge Detection:** This operation involves convolving the image with an edge detection kernel, such as the Sobel or Prewitt operators. It highlights the edges and boundaries in the image, aiding the model in learning edge-related features.

- **Embossing:** By applying an embossing kernel, this operation creates a 3D effect on the image, making it appear raised or engraved.

- **Sharpening:** By convolving the image with a sharpening kernel, this operation enhances the edge contrast and brings out finer details in the image

## D. Noise Operations:

Data augmentation can also be done by introducing noise in an image to introduce variations and enhance the model's ability to generalize to noisy and diverse data. Some commonly used types of noise for data augmentation include Gaussian noise, salt and pepper noise, and speckle noise.

- **Gaussian Noise:** By adding Gaussian noise to images, we introduce random fluctuations in pixel values, simulating real-world noise. This helps the model learn to be robust to noise and variations in image intensity.

- **Salt and Pepper Noise:** Salt and pepper noise is a type of random noise that randomly replaces some pixels in the image with either the maximum intensity (salt) or the minimum intensity (pepper). By adding salt and pepper noise, we simulate random pixel dropout or corruption, forcing the model to learn from partially corrupted data.

- **Poisson Noise:** Poisson noise is a type of noise that occurs due to the statistical nature of photon detection in imaging sensors. It is commonly observed in low-light conditions. By adding Poisson noise, we simulate the variations in photon counts and improve the model's ability to handle low-light scenarios.

### 3.1.1.2    Deep Learning Based Approaches:

### A. Texture Transfer:

Texture transfer is another popular technique for data augmentation to enhance the diversity of data and performance of the model by improving its ability to generalize to different textures. Texture transfer can be defined as a process to transfer the texture characteristics from one image to another image. Different algorithms and methods can be used for texture transfer such as patch-based transfer, neural-style transfer, and generative adversarial networks. These techniques work by combining the content of the target image with the source image and generate an augmented image. Texture transfer generally involves three steps; texture extraction, texture synthesis, and texture blending.

- **Texture Extraction:** In this step, unique textural features are captured by using different filter banks, texture descriptors, or deep learning-based feature extraction. 2

- **Texture Synthesis:** This step involves the generation of new texture patterns that resemble the texture of the original image while preserving the target image's content.

- **Blending:** This step blends the synthesized texture with the original target image and generates augmented data.

By applying texture transfer as a form of data augmentation, the training data set is expanded with images that exhibit different textures. This helps the model learn robust features that are invariant to texture variations and improves its performance on unseen data with similar textural characteristics.

### B. Adversarial Learning:

Adversarial training *(also known as machine illusion)* is also a technique used for the generation of augmented images, to make models more robust towards generalized/unseen images. It is used to train a machine learning (or deep learning model) both on original and adversarial created images. The whole training process consists of two parts; an original model and an adversary model. The main objective of the adversary model is to create adversarial images by applying small filters, or modifications to an input image.

Adversarial images are designed to deceive the model and cause it to produce incorrect predictions. The model is trained using the adversarial scenarios, which are first generated and then mixed with the original, clean data. The model is trained to identify and categorize these misleading inputs by exposing it to both the original and adversarial samples [103], [106]. This process helps the model become more robust and resilient against adversarial attacks and improves its ability to correctly classify both clean and adversarial inputs

### C. Generative Adversarial Learning:

Generative modeling *(also known as generative adversarial networks)* is also an amazing technique used for data augmentation. It can be defined as a technique to generate augmented images by imitating the original images from a data set in such a way that both original and imitated images should have the same characteristics. Generative adversarial networks (GANS) generally consist of two major components; a generator and a discriminator.

Both components work in parallel to each other in such a way that the generator generates an imitated image and discriminators evaluate the authenticity of these augmented images. In the context of data augmentation, the generator can be used to create additional training examples.

This can be particularly useful when the available training data is limited or imbalanced. The generated data can help the model learn more diverse patterns and generalize better to unseen data. For example, in image classification tasks, GANs can be used to generate new images that augment the original data set. These images can include variations that might not be present in the original data, such as different lighting conditions, angles, or transformations. However, it's important to note that GAN-based data augmentation must be used carefully. If the generator produces data that are too different from the real data distribution, it could lead the model to learn incorrect patterns. Similarly, if the generated data is too similar to the existing data, it might not provide any additional benefit. Therefore, it's crucial to monitor the quality of the generated data and the impact of the augmented data on the model's performance.

### 3.1.2 Introduction to Deep Learning

Artificial intelligence is a branch of computer science that refers to the development of machines to perform tasks where human intelligence is required. It plays a vital role in different rea-life applications such as natural language processing, machine learning, computer vision, and robotics. It includes different techniques and tools to analyze the vast amount of data, learn from patterns, and make future predictions with continually improving accuracy and efficiency. Machine Learning is the subset of artificial intelligence, which generally deals with algorithms and models that enable machines to analyze, learn, and predict by manually selecting the relevant features from input data. The performance of models always depends on the selection of relevant features. Models are trained on labeled detests.

Deep Learning is the subset of machine learning and generally involves the more advanced and complex algorithms with multiple number of hidden layers. Deep learning algorithms have the capacity to identify complex patterns and relationships in data through automatic learning and generating hierarchical representations of the data. As a result of this, deep learning is particularly useful in fields like natural language processing and computer vision. Deep learning involves neural networks for learning and making predictions. Deep learning is a data-hungry paradigm and always requires a large amount of data as compared to machine learning. This capability allows deep learning models to automatically identify relevant patterns and make accurate predictions [107].

### 3.1.3 Neural Networks

Neural networks (NN) (*as shown in* **Figure 3.4**) [108], specifically artificial neural networks (ANNs) or simulated neural networks (SNNs), are a specialized branch of machine learning that forms the foundation of deep learning algorithms. These networks are generally inspired by the human brain because of their anatomical and functional similarities and are formed by interconnecting a large number of small neurons. These small neurons are responsible for transmitting signals. These networks exhibit adaptive properties, making them allow machines to learn and improve performance from their errors. Through the use of neural networks, machines are able to analyze data, identify patterns, and modify their algorithms in response to input. This iterative process makes machines enhance their performance over time, making them more proficient at solving complex problems and making accurate predictions.

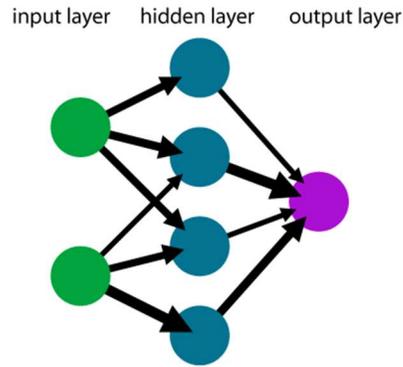

**Figure 3.4: Architecture of a Simple Neural Network**

Artificial neural networks (*as shown in* **Figure 3.5**) consist of serially interconnected layers of nodes [109]. These layers generally include; an input layer, an output layer, and a number of hidden layers. Each layer consists of one or more than one number of nodes (also known as artificial neurons). These artificial neurons receive input data, and use different mathematical functions (also known as activation functions) to process or analyze that data and generate output. All of these nodes have their associated weights to indicate the strength of the connection. ANN always works on the learning process. These weights are modified during the process of learning to optimize the network's productivity as well as its ability for accurate classifications or predictions [110], [111].

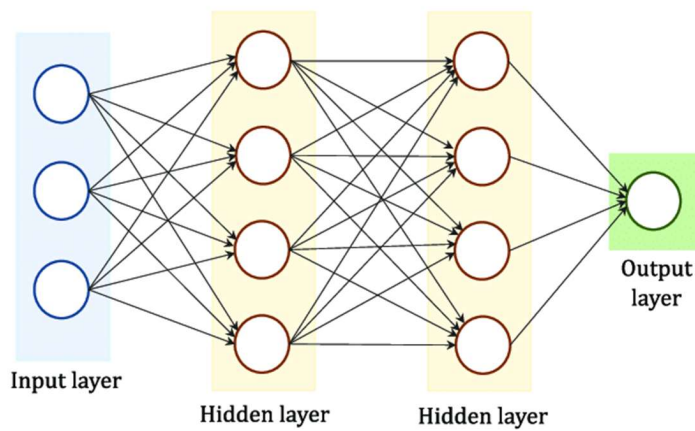

**Figure 3.5: General Architecture of Artificial Neural Network**

ANN is similar to the neural network present in our body (*as shown in* **Figure 3.6**) [111]. In ANNs, dendrites serve as the information terminals of these neurons, receiving input signals from other neurons via synapses. The input signals received by a neuron are processed through its axon, and the resulting output is transmitted to other neurons. The connections between neurons in an ANN are represented by lines, and the weight of each line is increased based on the input signals traveling through it. Signals transmitted by dendrites in the human body are added to form the cell body. The axon begins signal transmission if the total of these signals is greater than an established threshold.

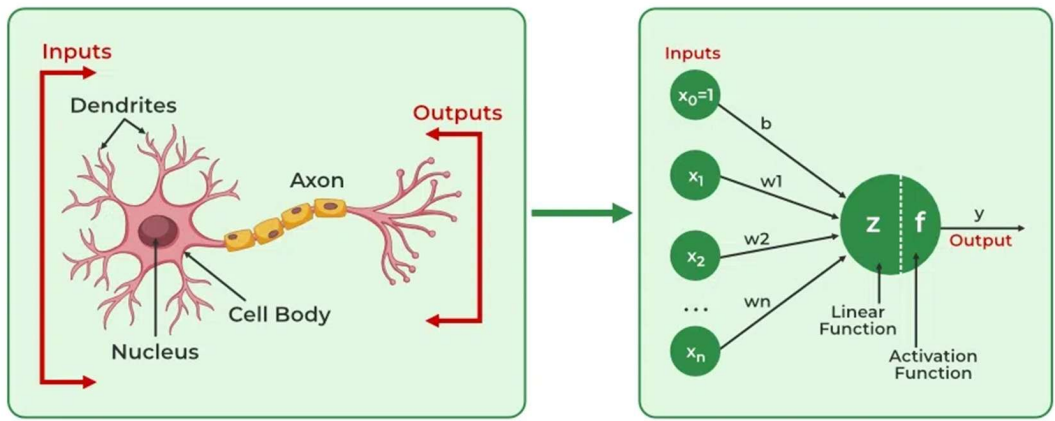

**Figure 3.6: Artificial Neural Network and its resemblance with human brain**

Similarly, in a mathematical or numerical model, a similar approach is employed. The threshold value is determined by applying an activation function, usually represented by the symbol "*f*". The sigmoid function [112] is one of the most extensively used activation functions. it works on the principle that it converts the sum of input values between 0 to 1. This conversion enables the model to make decisions or predictions. If the output value of A.F. is closer to 1, it indicates a stronger activation and vice versa [111].

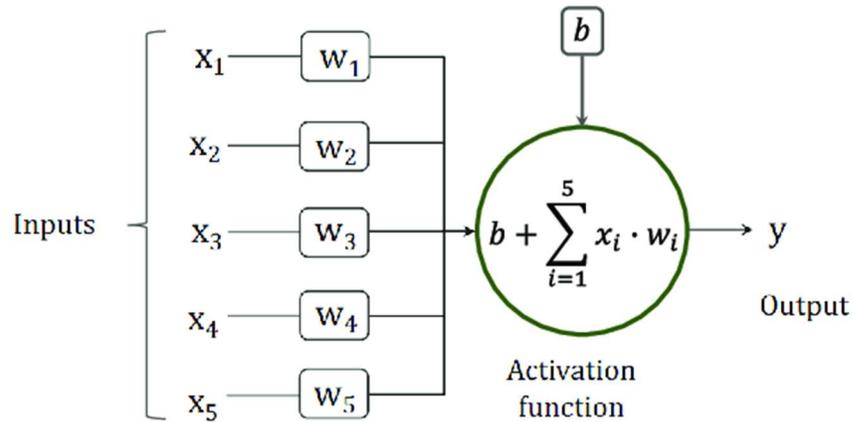

**Figure 3.7: Mathematical Structure of Artificial Neural Network with 5-inputs and 1-output**

Depending on the flow of data, neural networks can be categorized into different types such as feed-forward NN, back-propagated neural networks, and Convolutional Neural Networks (CNN).

***Feed forward NN (also called Multi-Layer Perceptron MLP)*** refers to networks that process in a unidirectional way, only in the forward direction. It generally consists of an input layer, a hidden layer, and an output layer [113]. Each neuron in a feed-forward network (FFN) applies inputs from previous layers and uses an activation function to produce an output. Hidden layers extract and transform features from input data, creating a fully connected architecture. Feedforward neural networks are versatile and effective in tasks like pattern recognition, classification, regression, and function approximation.

***Back propagated neural network*** [114] is a type of neural network that utilizes the backpropagation algorithm for training. Backpropagation (*as shown in* **Figure 3.8**) is a widely used learning algorithm that allows the network to adjust its weights and biases based on the discrepancy between the predicted output and the desired output. Back-propagated neural networks use feed-forward information flow from input to output layers. The network produces predictions via forward passes during training. The error between the outputs and the predictions is calculated by the loss function. In order to mitigate errors, the backpropagation technique adjusts weights and biases while generating the gradient of the loss function. Updates are made repeatedly until the network converges or reaches a suitable accuracy. In order to understand complex patterns and relationships in data, back-propagated neural networks iteratively adjust parameters based on back-propagation errors.

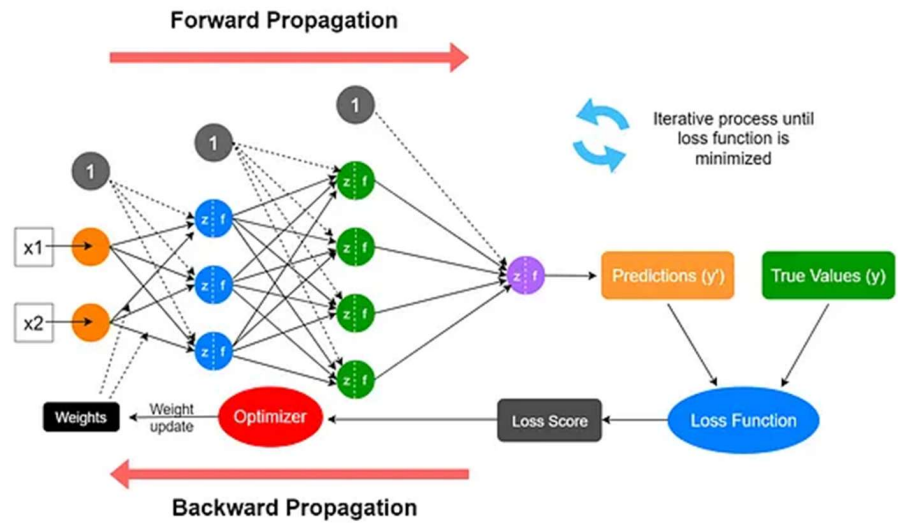

**Figure 3.8: Backpropagated Neural Networks – How it works?**

### 3.1.4 Convolutional Neural Networks

The main aim of artificial intelligence is to reduce the gap between human capabilities and machine capabilities [115]. It becomes possible with the help of deep neural networks because of their ability to learn complex patterns and relationships. It is also the fact that deep hidden layers of Deep NN have performed better as compared to traditional machine learning techniques [116]. Convolutional neural network *(also known as ConvNet)* is generally a type of deep neural network having wide applications in the field of computer vision to analyze and learn visual data (images and videos), and make predictions. General architecture of CNN is (***as shown in*** **Figure 3.9**).

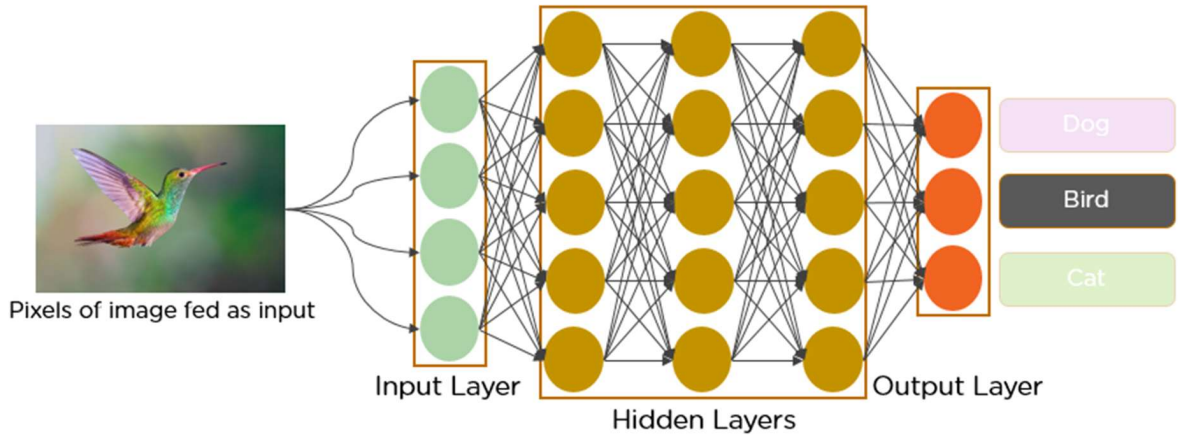

**Figure 3.9: General Architecture of Convolutional Neural Network**

- It takes images in the form of input, processes them through hidden layers, assigns weights/biases to important features, and processes them for prediction or classification tasks. A convolutional neural network generally comprises three main parts; input, feature extraction, and output.

- The Input part generally takes images and passes it to the feature extraction network. The number of nodes in the input layer of a neural network is typically equal to the number of pixels in an image. This is because each pixel in an image can be considered as an individual input to the network.

- Feature extraction network generally consists of three types of layers; convolution, activation, and pooling layers. A feature extraction network may have one or more than one number of these layers. The greater the number of these layers, the deeper will be the neural network and the better will be the performance of the network.

- Convolution layers perform the convolution operation (or multiplication). A filter is applied to the image to extract the meaningful / required information from the image and generates a feature map (or activation map). Feature map generally consists of specific features at different spatial locations.

- These feature maps are then processed through pooling layers to reduce the size of feature maps. By reducing the size of the feature map, complexity, and computational power will also be reduced. Pooling can be of different types; max pooling, min pooling, and average pooling. During convolution operation, a filter of size *(nxn)* slides over the input image. At each position, the maximum value from within the window is selected and retained. This is called max pooling. On the other hand, if we take an average of all values within the sliding window then it will be called average pooling.

### 3.1.5 Multi-Column Convolutional Neural Networks

Inspired by the remarkable success of Multi-Column Deep Neural Networks (MC-DNN) [117] in different computer vision tasks, Zhang et. al [13] proposed Multi-Column Convolutional Neural Networks (MCNN). Multi-Column CNN are type of deep neural network, specially designed to deal with multi-channel input data and address the challenges posed by objects with varying sizes and scales. This capability allows M-CNN to make accurate and robust predictions by capturing and analyzing the complex relationships and patterns with in the input data. Its architecture (*as shown in* **Figure 3.10**) makes it particularly well-suited for a range of tasks, including image classification, object detection, and semantic segmentation.

M-CNN consists of multiple convolutional columns, connected in parallel. Input data is divided into multiple columns and each column is responsible for performing separate convolution operations. Like CNN, each column of M-CNN consists of different operations/layers such as convolution, pooling operations, non-linear activation functions, and fully connected layers as well. These layers help to extract meaningful features from the input data and reduce the spatial dimensions of the feature maps. The outputs from each column are then concatenated and passed through fully connected layers, which further process the extracted features and make predictions based on them. The final output of the MC-CNN is typically a probability distribution over the different classes or labels in the task.

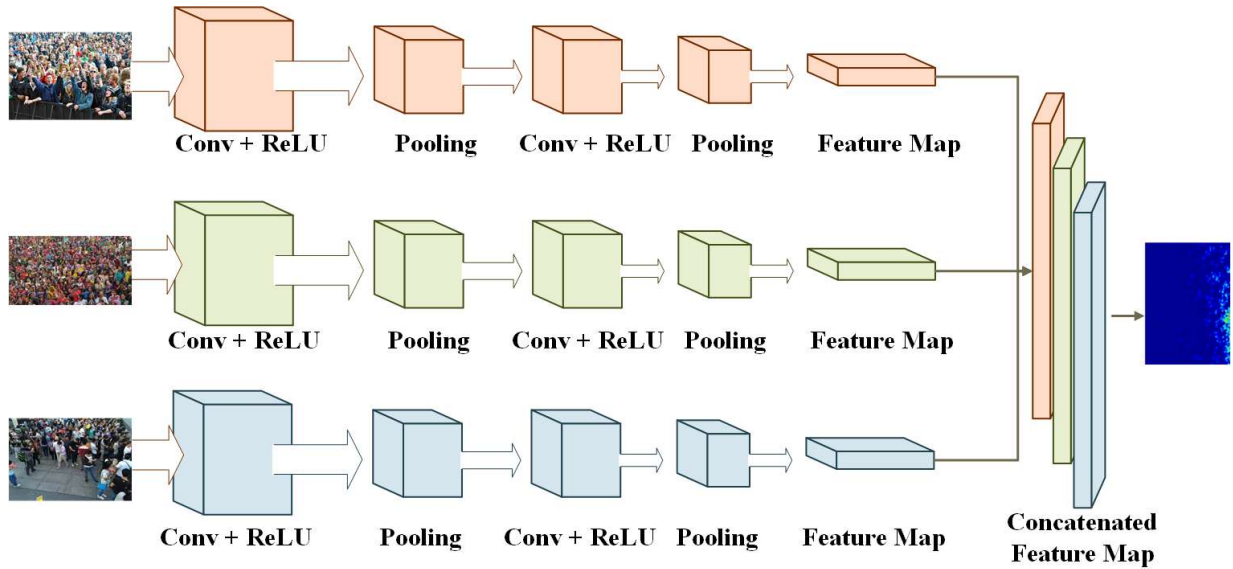

**Figure 3.10: General Architecture of Multi Column Convolutional Neural Network (MCNN)**

In the case of crowd scene analysis, M-CNN excel at accurately estimating the number of individuals in a crowded scene. Multi-column structure of M-CNN makes it suitable for analyzing a crowded scene at different scales and captures different densities of a crowd. Images of crowd scenes are taken as an input of M-CNN and then subjected to convolutional layers. Convolutional layers perform convolution operations on input images with the help of filters of different sizes to extract meaningful features. These features are then subjected to pooling layers to down sample the size of the feature map by reducing the spatial dimensions and capturing the most important information. Non-Linear activation functions are then applied to introduce non-linearity in the network. It makes the model learn more complex and expressive representations. Feature maps from the output of each column will be then concatenated to generate the integrated representation of crowd density in the form of a concatenated feature map. This concatenated feature map will be then subjected to a fully connected layer to estimate the final count of the crowd.

One of the key advantages of using M-CNN is that the multi-column structure can capture spatial information at different levels (both local and global scales). Each column can focus on learning different representations to extract fine-grained details at a local scale, as well as capture broader contextual information at a global scale. It allows the network to specialize in different spatial scales, enabling it to learn more accurate and comprehensive representations of the data. This enhances the network's ability to extract meaningful features and make precise predictions.

Another advantage of M-CNN is that it can process multi-modal input data by taking each modality corresponding to an individual column within the network. It makes M-CNN suitable for tasks with multiple types of input data such as image-text matching or audio-visual fusion. For example, in image-text matching, image data can be handled by one column and textual data can be processed by another column. It allows the model to facilitate accurate matching and understanding between two different modalities by learning the joint representations from both modalities that align the visual and textual elements. Another example is audio-visual fusion tasks, where audio and video data will be taken by two separate columns. It makes the model efficiently integrate acoustic and visual cues, improving interpretation and analysis of the combined information by jointly learning the representations from both modalities.

### 3.1.6 Density Maps:

Density maps can be defined as pixel-by-pixel representations of an image, in which each pixel represents local crowd density at that specific location. Generally, these maps are used to indicate the density of people in an image for counting tasks. The density values in the map are typically obtained by assigning higher values to areas with more individuals and lower values to areas with fewer individuals. The density map (*as shown in* **Figure 3. 11**) is generated using ground truth annotations, which label or annotate each individual in the image.

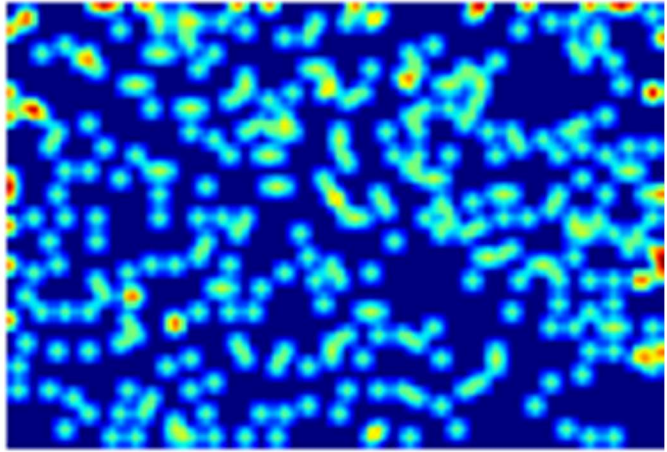

**Figure 3. 11: Density Map**

## 3.2  End-to-end Proposed Framework:

The proposed framework Self-supervised multi-column convolutional neural network for crowd counting aims to count the number of people in a crowded scene. The proposed model consists of two stages; Stage one is dedicated to the initial training of self-supervision. This stage involves the training process using rotated crop images by learning to predict the rotation angle of input images. The network learns useful features related to edge and orientation detection from this self-supervised activity. We used the self-supervised training module to eliminate the need for annotated images This approach allows the model to learn from unlabeled data. By formulating pretext tasks or artificial supervisory signals, the model can generate training signals from the data without relying on costly and time-consuming image annotation.

After completing the initial training phase, the FEN developed insight into significant features associated with density estimates. Features extracted from FEN are frozen now and the rotation classification has also been removed. These frozen features are now used in the density estimation through distribution matching. SINKHORN matching is proposed for distribution matching SINKHORN matching refers to the use of the SINKHORN distance to measure the similarity between the predicted crowd density distribution and a prior distribution that approximates the statistical characteristics of natural crowds.

The first stage consists of a feature extraction network for the extraction of useful features from input images. This network is further attached with two separate convolutional blocks derived from VGG19 architecture. VGG19 block includes max pooling layers to further down-sample the feature maps. Finally, a fully connected layer is incorporated to classify the input images and predict the angle of rotation for the rotated crops.

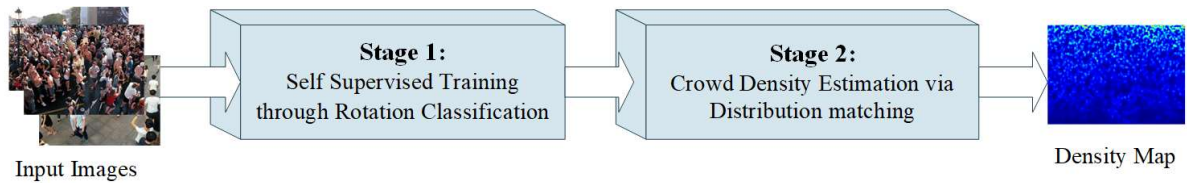

**Figure 3.12: Proposed architecture for crowd counting using self-supervised training. Stage 1 utilizes the rotation task for self-supervised training, and stage 2 uses distribution matching for density estimation**

The feature extraction network of the first stage is then extended to the second stage to serve as a density regressor. This extension involves the addition of two more VGG-based convolutional blocks. These added convolutional blocks transform the FEN from self-supervised training to density regressor and enable the network to refine the extracted features and efficiently map them for density regression. Overall structure improves the accuracy of crowd counting. After this process of feature extraction, the next step is to match the statistics of the density predictions with those of the prior distribution using optimal transport. This matching process involves optimizing the transportation plan between the predicted density and the prior distribution to achieve the best possible alignment. Each component of the proposed framework will be discussed in detail in this section. Overall architectures of proposed framework are (*as shown in* **Figure 3.13**).

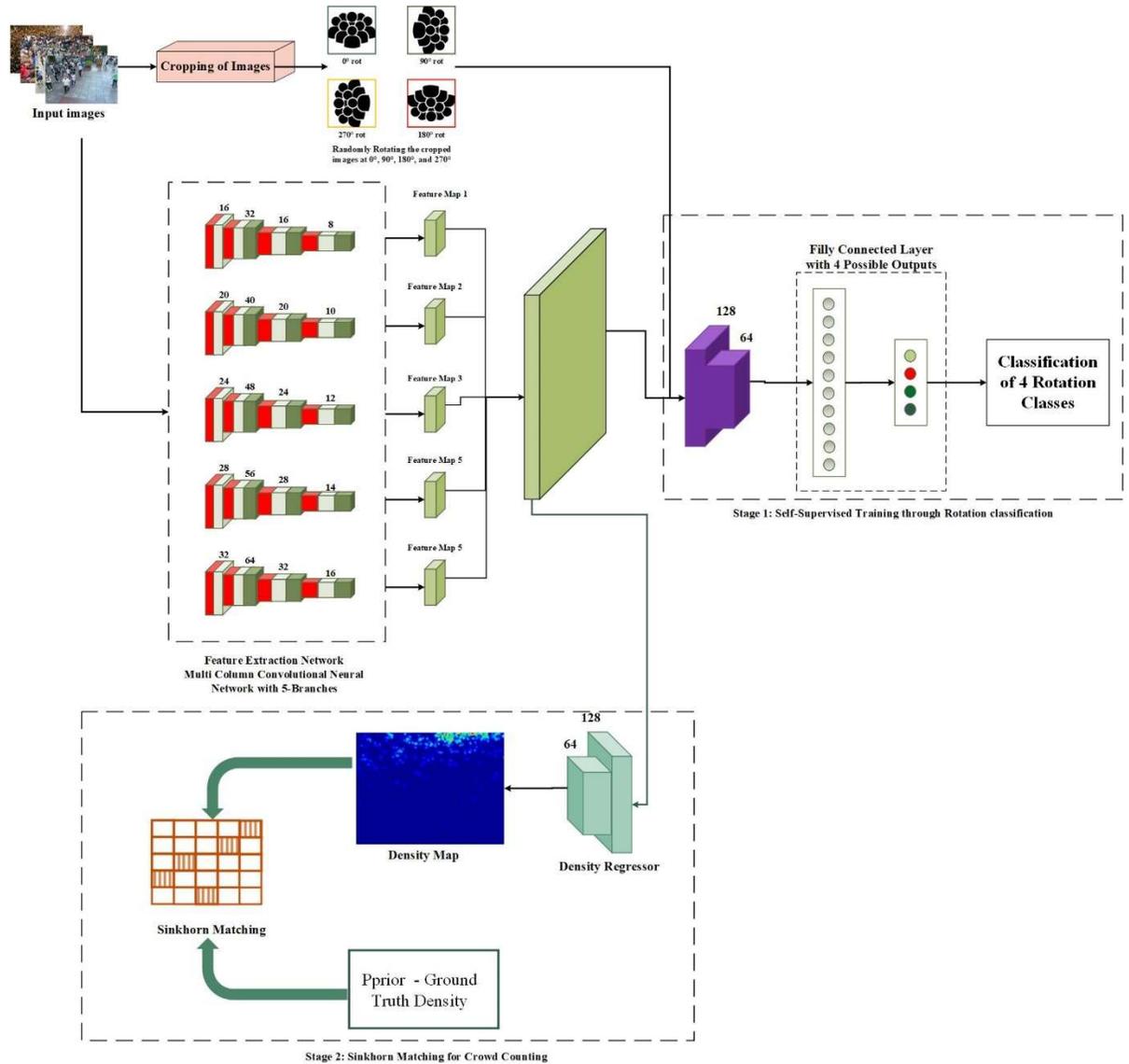

**Figure 3.13: Proposed Model for Crowd Counting**

For the feature extraction network, we first introduce the Multi-column CNN with five convolution branches in parallel. Each branch uses filters of different sizes to capture information at different scales and improves the accuracy of prediction. This design aims to enhance the robustness of the system towards various challenges such as complex backgrounds, occluded scenes, low-resolution images, variable sizes of people, and their heads/shoulders. Each column of MCNN (*as shown in* **Figure 3.14**) consists of a sequence of convolution and pooling operations in such a way that *(Conv + Pool + Conv + Pool + Conv + Pool).*

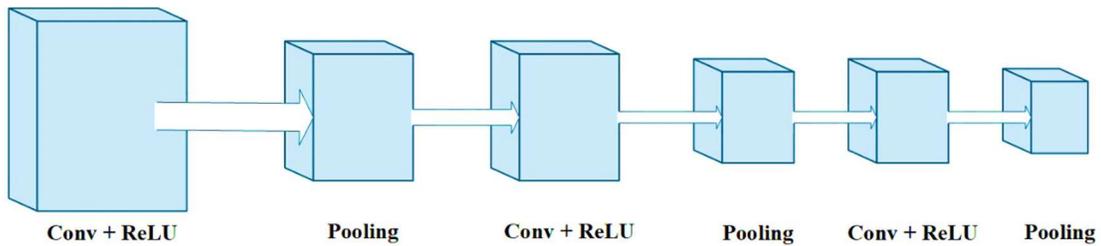

**Figure 3.14: Single branch of M-CNN: Sequence of Convolution and Pooling operations**

### 3.2.1 Data Augmentation:

For crowd counting related tasks, data augmentation is a useful technique as it enables the model to handle different challenges such as variations in crowd appearances, scale, and orientation. Data augmentation enhances the model's capacity to generalize and precisely estimate crowd density in real-world settings by exposing it to a larger range of crowd configurations. In our proposed model, data augmentation plays a crucial role by applying random rotations as a form of data.

Two different types of data augmentation techniques are used in our proposed model; cropping and rotation. We cropped input images at size *112x112*. These crops are then served as input to the rotation process. The input photos are rotated by four different angles *((0, 90,*

***180, or 270))*** degrees. This random rotation adds variety to the training set and facilitates the model's ability to correctly classify the input image in any orientation. The network learns properties that are rotation-invariant by introducing random rotations, which strengthens its resistance to changes in population orientation. Due to the random rotation augmentation, the proposed model is better able to generalize to real-life situations where crowd orientations may change. Regardless of number of people inside the crowd, it enables the network work to learn how to accurately predict crowd density.

### 3.2.2    Feature Extraction Network:

It is an important part of our proposed model that is used to extract useful information and important features (such as spatial arrangement, sizes, and other features) from crowd images to predict an accurate count of individuals.

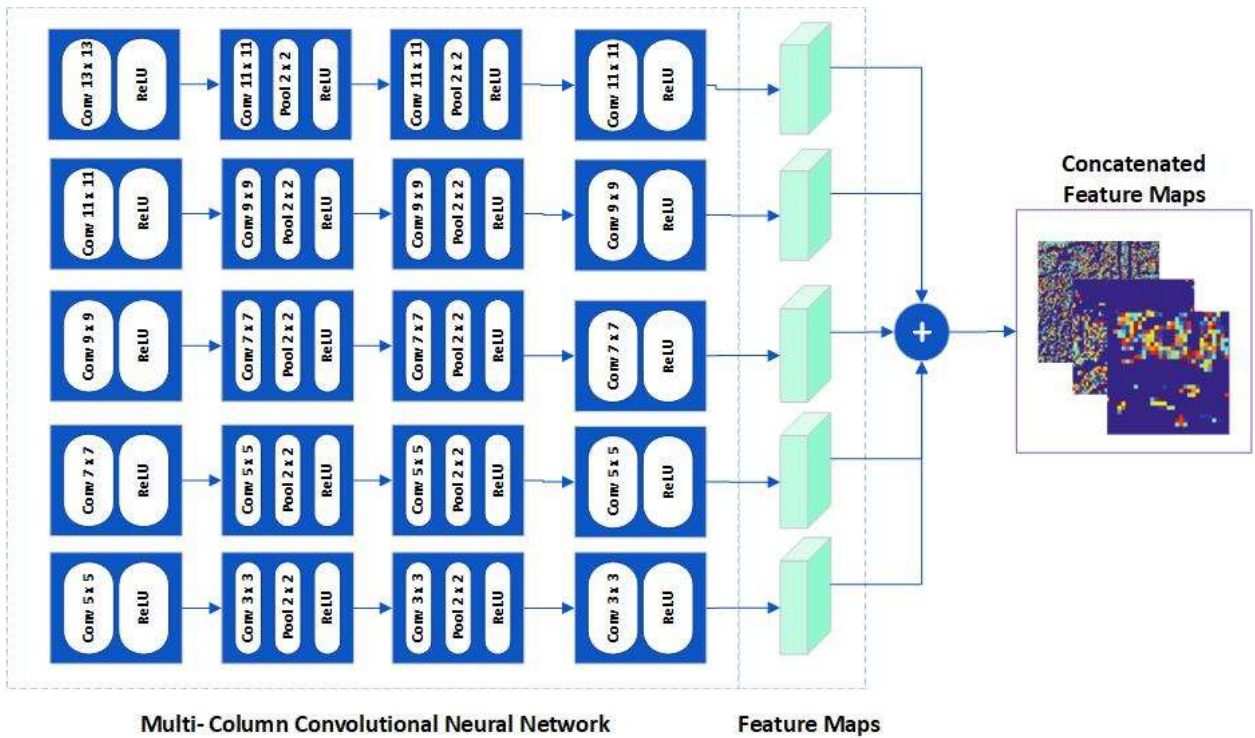

**Figure 3.15: Proposed Architecture of M-CNN**

Generally, a FEN is composed of several convolution branches, pooling layers, non-linear activation functions, and fully connected layers. Inspired by M-CNN of Zhang et. al. [13], our approach involves an extension of M-CNN by upgrading the number of branches. We proposed M-CNN with five convolutional branches in parallel. General architecture of proposed M-CNN (*as shown in* **Figure 3.15**). Each column consists series of convolutional operations with max pooling layers in between them and non-linear activation function. Filters of different sizes are used for convolution operations to extract information at different scales (both at local and global levels). Max pooling layers are proposed to down sample the feature map to capture and process important information.

We proposed ReLU as an activation function. The main reason to propose ReLU is its ability to address the vanishing gradient problem, which occurs when gradients become small during backpropagation, preventing slow convergence and difficulty in training deep neural networks. Its computational efficiency, which is inexpensive compared to complex functions like sigmoid or tanh, makes it suitable for large-scale crowd-counting tasks [118], [119] In our proposed model, FEN plays a vital role in the accurate prediction of crowd density.

Features extracted from FEN serve as the foundation for both stages of self-supervision and density regression. These features are used by stage one to generate prior distribution by using dense crowd images, while stage two utilizes these features to match the predicted density distribution with prior distribution and generate density maps. The density regressor of stage two is a critical component that makes use of the extracted features. By leveraging the discriminative features captured by the FEN, the density regressor can effectively estimate the density of the crowd in the given input image. These extracted features are also utilized by the distribution matching of stage two. The distribution matching module aims to match the predicted density distribution with a prior distribution. This process helps to align the estimated crowd density with a predefined distribution, ensuring that the predicted density maps adhere to certain statistical properties or characteristics.

### 3.2.3 Stage 1: Completely Self-Supervised Learning:

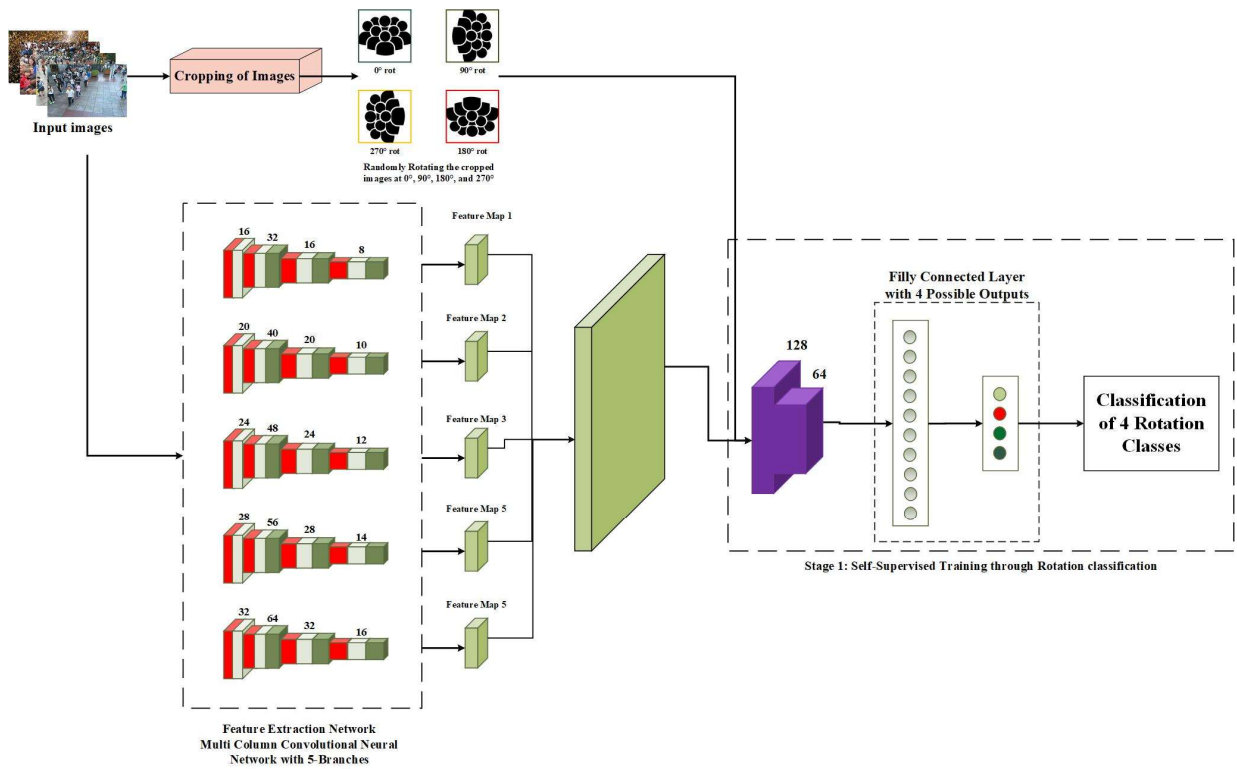

**Figure 3.16: Proposed Architecture of Stage 1 for Crowd Counting. Self-Supervised training through Rotational Classification**

The model (*as shown in* **Figure 3.16**) is trained by leveraging self-supervision to learn effective and generic features that are beneficial for the ultimate task of density estimation. The whole self-supervised training process will occur in two different steps. The first step focuses on capturing patterns that frequently appear in the input images. Since we only provide dense crowd images during this stage, our intention is for the model to primarily learn features that are relevant to crowds. These features can range from specific edges that distinguish head-shoulder patterns to higher-level crowd semantics. By utilizing self-supervision in this manner, the features acquired by the first stage serve as an accurate initialization for the second stage, which involves distribution matching. It is important to note that the model is not explicitly instructed to learn representations specifically for density estimation.

However, through self-supervision, the model naturally converges on learning crowd patterns as they are the most prominent aspect of the input data distribution. Overall, we use self-supervision to train the model in phases, with the first stage focusing on learning crowd-relevant attributes. The obtained features serve as a reliable initialization for the second step, in which distribution matching is used to reliably estimate crowd density.

This stage comprises feature extraction networks (*as described in 3.2.2*) followed by two convolutional blocks of VGG and a fully connected layer. This stage serves to enhance the model's capacity to cope with variations in crowd image orientation and accurately estimate crowd density in various situations. To achieve this, *112×112* crops are extracted from the crowd images as part of a data augmentation strategy. These crops serve as the input for the rotation classification branch. In order to introduce diversity into the training data, each crop is randomly rotated by one of the four predefined angles: *0, 90, 180, or 270 degrees*.

Two additional convolutional layers and fully connected layers collectively enable the classification task, helping the model learn to accurately classify the rotation angles of the input images. Indeed, self-supervision provides a variety of approaches for producing pseudo labels that can be used to effectively train models. Predicting picture rotations is one highly efficient strategy that gives a simple yet robust method for creating meaningful representations. The network learns to recognize characteristic edges and even higher-level patterns important for determining object orientation by randomly rotating an image and training the model to predict the degree of rotation.

We trained our model by leveraging self-supervision via rotation to generate pseudo labels. By using a rotation-based pretext task, self-supervision is attained. The first step of training involves teaching the model to predict the rotation angle of input images. The network learns to capture rich and discriminate information useful for crowd counting by randomly rotating the input photos and training the model to accurately detect the rotation angle. This technique of self-supervision enables the model to learn from un-annotated data. It reduces the time and cost of manual annotation and hence improves the model accuracy to predict the accurate count of unseen scenes.

### 3.2.4 Stage 2: Crowd Density Estimation via Distribution matching:

Stage two (*as shown in* **Figure 3.17**) also comprises of feature extraction network with two more VGG-based convolutional branches. To reduce over-fitting and improve generalization, a special training technique is used in this stage. The weights of the layers of the feature extraction network (FEN) are not updated at this stage since they are frozen. However, only a portion of the newly added convolutional layer parameters are left open for training. The network keeps the valuable features gained during the self-supervised training step by freezing the majority of the FEN layers. These characteristics are critical for estimating crowd density accurately. A density regression branch is used in this stage to map the features retrieved in the first stage to the required density map output. As input, the density regression branch uses the features learned by the Feature Extraction Network (FEN). These layers enable the network to learn the mapping between the retrieved features and the density map representation. The main objective of the density regression branch is to generate a density map by matching the prior distribution with the predicted density distribution.

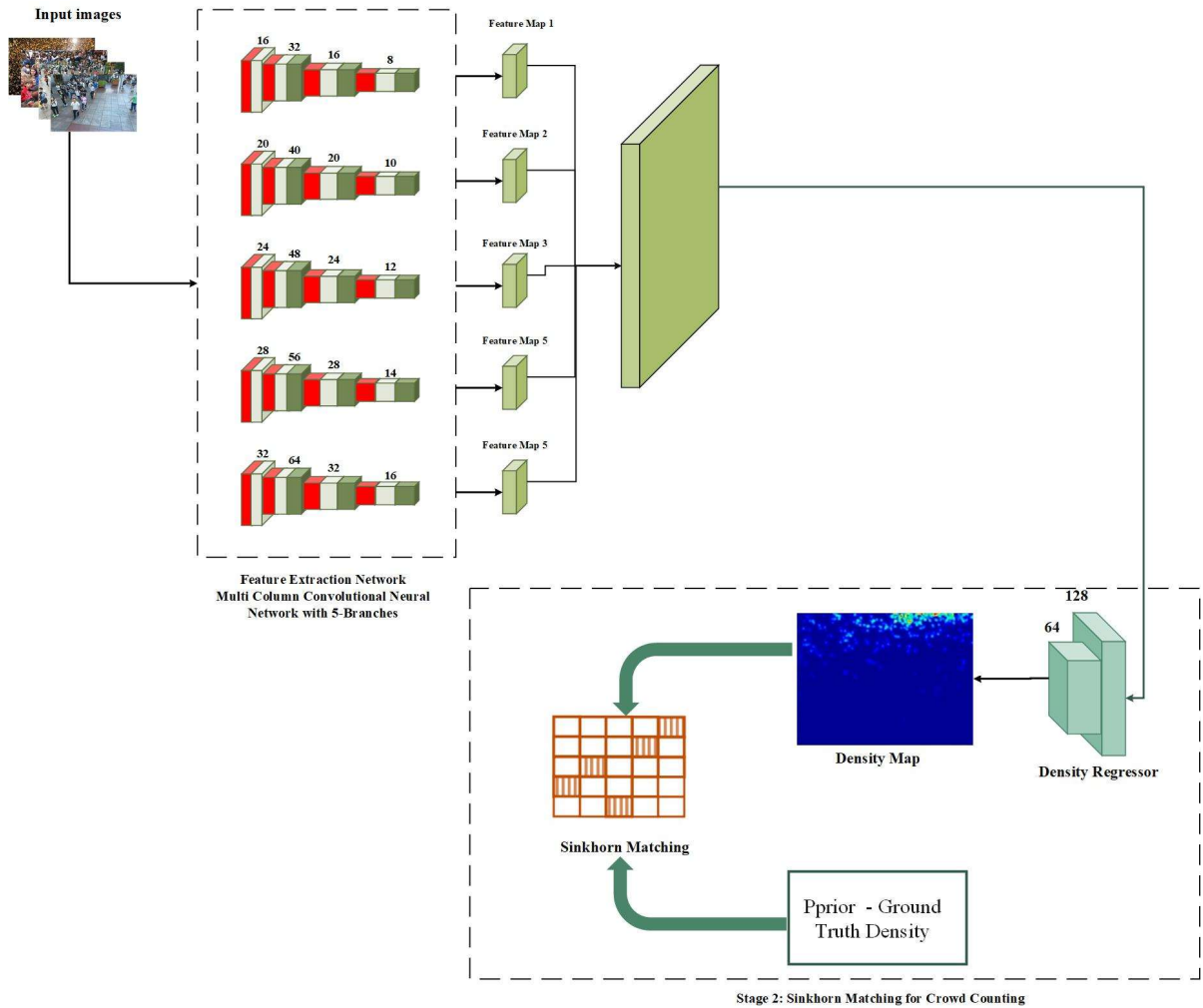

**Figure 3.17: Proposed Architecture of Stage 2 for Crowd Counting. SINKHORN Matching**

# CHAPTER FOUR: CROWD COUNTING EXPERIMENTAL RESULTS

## 4.1 Implementation Details:

### 4.1.1 Network Architecture:

Network architecture pre-dominantly revolves around the extended version of a multi-column convolutional neural network. It comprises five convolutional branches in parallel having different feature sizes with the aim to capture useful information at different levels at the same time. Each branch comprises four convolution operations in series. After each convolution operation, max pooling with filter size 2x2 is performed to reduce the spatial dimension of the feature map. It results in the reduction of the feature map by a factor of 1/4. Network architecture also comprises branches for rotation classification and distribution matching. The rotation classification branch enables the proposed model to learn and classify the input images with different angles. While the distribution matching branch utilizes SINKHORN matching to match the actual ground truth density map and predicted density map. This branch aids the model to accurately estimate the people in a crowded scene.

### 4.1.2 Training Data:

Annotated and unannotated data is used for training purposes. The annotated dataset consists of crowd photos along with manually annotated density maps which provide accurate information about the number as well as arrangement of people in the photos. This annotated dataset is utilized for the pre-training of the model. The unannotated dataset is utilized for self-supervised training.

### 4.1.3 Data Augmentation:

Two different types of augmentation techniques are employed on data sets; including cropping and rotation. We took *112x112* crops from input images and employed them to four different rotational angles; *0, 90 180, and 270 degrees.*

## 4.2 Training Details:

We trained our proposed model in two stages; pre-training and fine-tuning by leveraging the self-supervision technique.

### 4.2.1 Pre-training:

In the first step, the model will be trained on a labeled dataset with the help of an annotated dataset by adopting a supervised training mechanism. The model is fed with images and their relevant ground-truth maps. These ground-truth maps are manually annotated and will be used to predict the count of individuals present in a crowded scene. The main objective of pre-training in a supervised mechanism is to enable the model to capture the basic capabilities.

### 4.2.2 Self-supervised training:

The network goes through a phase of self-supervised fine-tuning after pre-training. In this stage, the network is trained using unlabeled data rather than manual annotations. The goal is to use self-supervision approaches to improve the network's counting capabilities. Several techniques can be used in self-supervised fine-tuning. Estimating density maps for the self-supervision task is one such method. The network is trained to produce density maps that depict the concentration or density of individuals inside the image.

The density maps are compared to the ground truth density maps created with Gaussian kernels. This overall process enables the models to learn in such a way as to decrease the variance between its predicted density maps and the ground truth density maps. Self-supervised training is further divided into two tasks: rotation classification, and distribution matching.

### A. Rotation Classification:

During training, we took 112x12 crops of images and subjected them to rotational augmentation by rotating them at four different angles; 0, 90, 180, and 270 degrees. It creates a diverse set of training examples with various rotation angles. In this stage, the model is trained to learn and classify the rotation angle of the augmented image. It makes the model learn useful features, that are necessary for reliable and efficient crowd density estimation. Being able to handle various rotation angles allows the model to generalize toward different scenes, permitting it to perform well on crowd images regardless of orientation.

### B. Distribution Matching:

Distribution matching, the second part, aims to automatically correlate the ground truth density maps to the predicted density maps. The objective of the network's training is to minimize the difference between the ground truth density maps and the predicted density maps. By doing this, the network is encouraged to record the more intricate aspects of the crowd distribution, which raises the density estimation's accuracy

## 4.3 Datasets:

### 4.3.1 ShanghaiTech:

This data set is publicly available, and consists of 1198 annotated images divided into two parts; Part A and Part B. These 1198 annotated images consist of 330,165 annotated persons. Each person is annotated with the help of points close to the head location. There are 30,165 annotated.

- Part A consists of highly crowded images collected from different internet resources such as YouTube videos, and images from social media applications. Part A consists of 482 images and is divided into two sub-parts; test and train sub-parts. The test part consists of 300 and the train part consists of 182 images.
- Part B consists of sparse crowded images taken from different busy streets, and areas of SHANGHAI. Part B is also divided into two sub-parts; the train part consists of 400, and the test part consists of 316 images.

### 4.3.2 UCF-QNRF:

This is also one of the publicly available data sets consisting of images taken from FLICKER, Web search, and Hajj Footage. The numbers of people vary from 50 to 12000 across 1535 number of high-definition images. This data set contains real and diverse scenes and background types with large and high-quality annotations.

## 4.4 Evaluation Metrics:

Mean Absolute Error (MAE), and Mean Square Error (MSE) are used for calculating the loss between actual and predicted density maps. The accuracy of the crowd-counting estimation is represented by MAE, and the robustness of accuracy is represented by MSE.

$$MAE = \frac{1}{n} \sum_{i=1}^{n} |y_c - y_{GT}| \quad \quad (Eq.\ 1)$$

$n = number\ of\ images$
$y_c = number\ of\ peoples\ counted$
$y_{GT} = number\ of\ peoples\ predicted\ from\ ground\ truth\ images$

$$MSE = \frac{1}{n} \sum_{i=1}^{n} (y_c - y_{GT})^2 \quad \quad (Eq.\ 2)$$

$n = number\ of\ images$
$y_c = number\ of\ peoples\ counted$
$y_{GT} = number\ of\ peoples\ predicted\ from\ ground\ truth\ images$

## 4.5 Experimental Results:

### 4.5.1 ShanghaiTech (Part A)

#### *4.5.1.1 Case 1: Results on images from ShanghaiTech Part A with Scale in-variation*

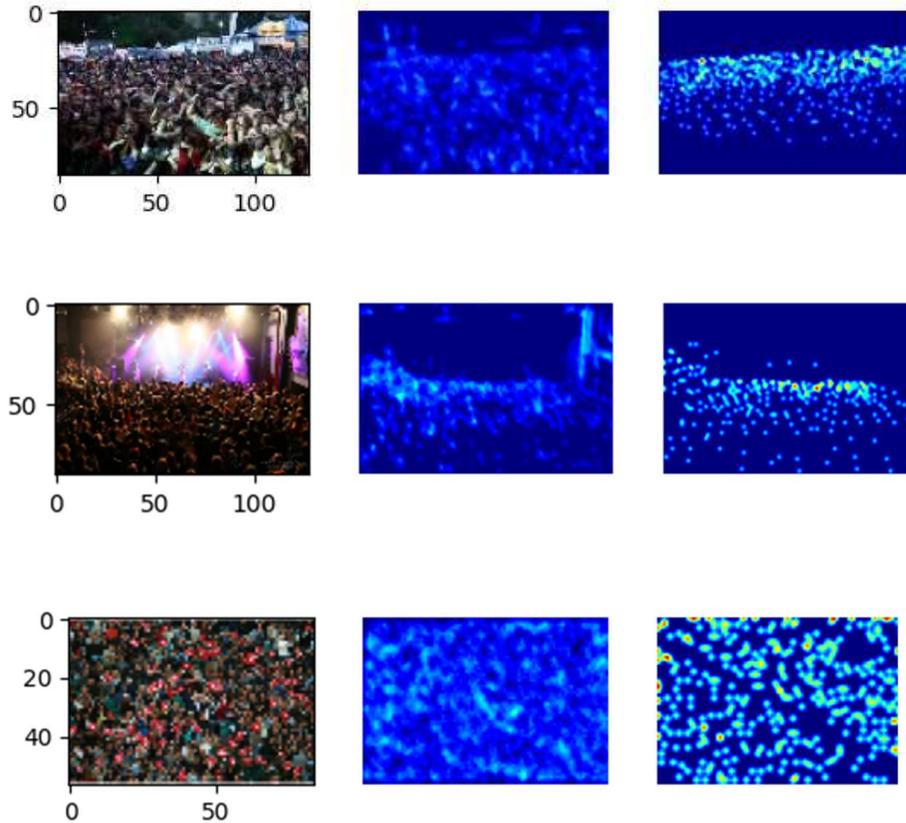

**Figure 4.1** Results on Images from ShanghaiTech PartA with scale in-variation

**Table 4.1:** Ground Truth vs. Predicted Count for Images with Scale In-variation

| Image #: | Ground Truth Count | Predicted Count |
|---|---|---|
| Image 1 | 427 | 423.16 |
| Image 2 | 240 | 286.89 |
| Image 3 | 320 | 288.69 |

*4.5.1.2     Case 2: Results on images from ShanghaiTech Part A with Occluded Scenes*

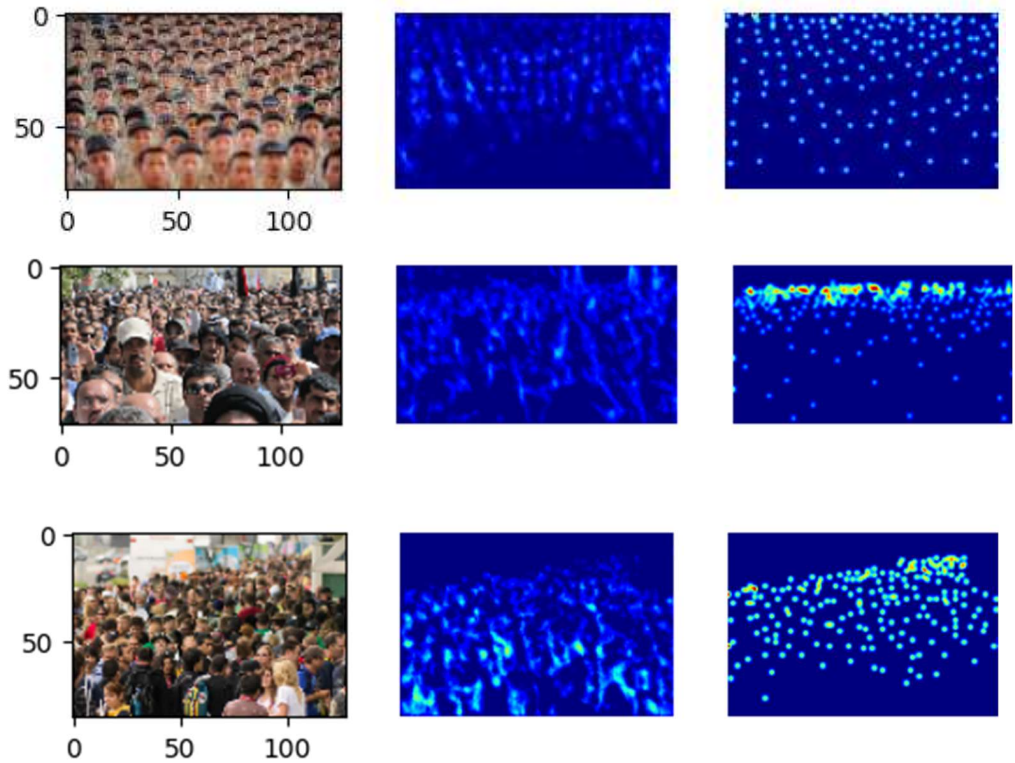

**Figure 4. 2** Results on Images from ShanghaiTech PartA with occluded scenes

Table 4.2 Ground Truth vs. Predicted Count for Images with Occluded Scene

| Image #: | Ground Truth Count | Predicted Count |
|---|---|---|
| **Image 1** | 147 | 139.35 |
| **Image 2** | 279 | 283.74 |
| **Image 3** | 199 | 223.75 |

*4.5.1.3     Case 3: Results on images from ShanghaiTech Part A with Complex*

*Background*

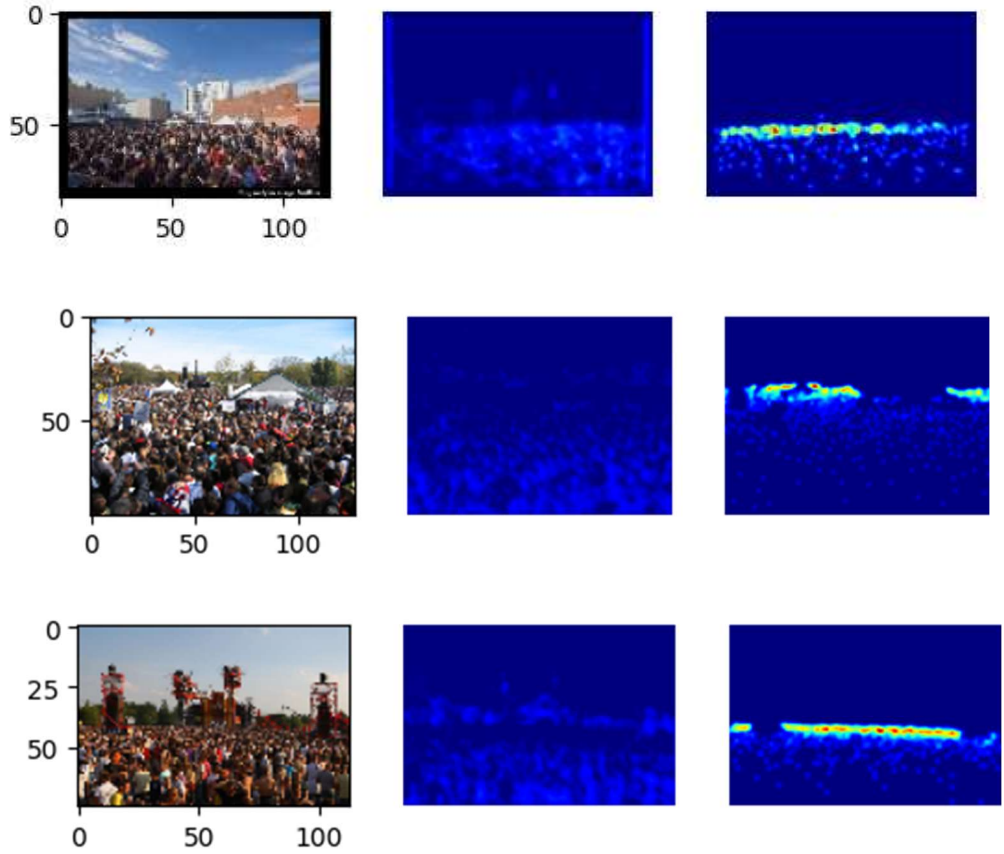

**Figure 4.3:** Results on Images from ShanghaiTech PartA with complex background

**Table 4.3:** Ground Truth vs. Predicted Count for Images with Complex Background

| Image #: | Ground Truth Count | Predicted Count |
|---|---|---|
| **Image 1** | 268 | 259.05 |
| **Image 2** | 962 | 799.58 |
| **Image 3** | 707 | 431.64 |

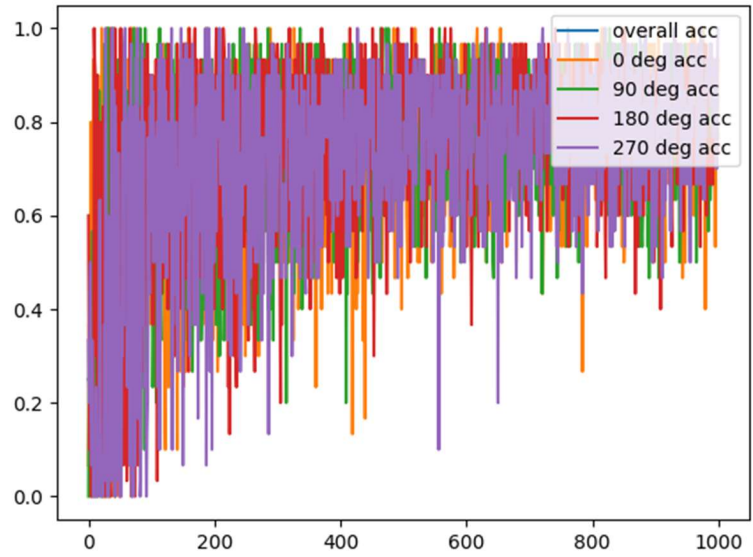

**Figure 4.4:** Overall Rotational Accuracy for classification of rotation angle on ShanghaiTech Part A

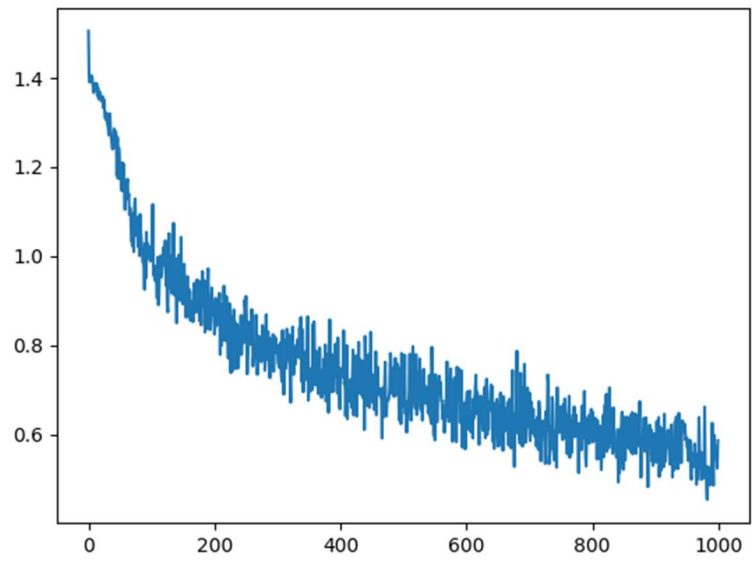

**Figure 4.5:** Overall training loss during self-supervised training

## 4.5.2 ShanghaiTech (Part B)

### 4.5.2.1 Case 1: Results on images from ShanghaiTech Part B with Scale In-variation

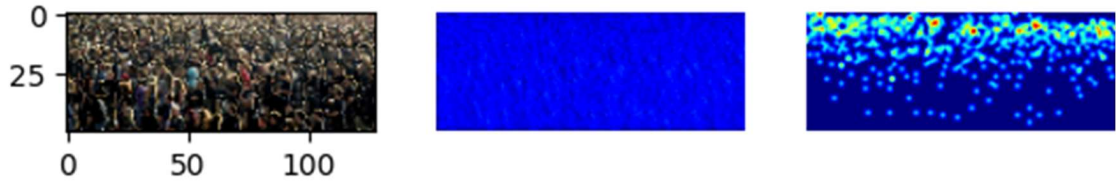

**Figure 4.6:** Results on Images from ShanghaiTech PartB with scale in-variation

**Table 4.4:** Ground Truth vs. Predicted Count for Images with Scale In-variation

| Image #: | Ground Truth Count | Predicted Count |
|---|---|---|
| **Image 1** | 407 | 376 |

### 4.5.2.2 Case 2: Results on images from ShanghaiTech Part B with Complex Background

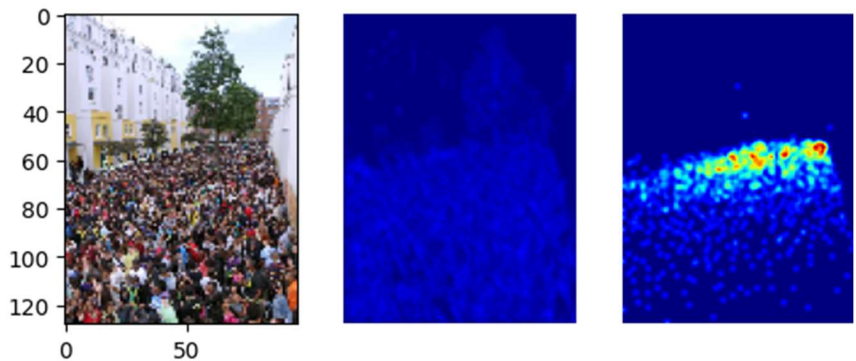

**Figure 4.7:** Results on Images from ShanghaiTech PartB with Complex Background

**Table 4.5:** Ground Truth vs. Predicted Count for Images with Scale In-variation

| Image #: | Ground Truth Count | Predicted Count |
|---|---|---|
| **Image 1** | 975 | 577 |

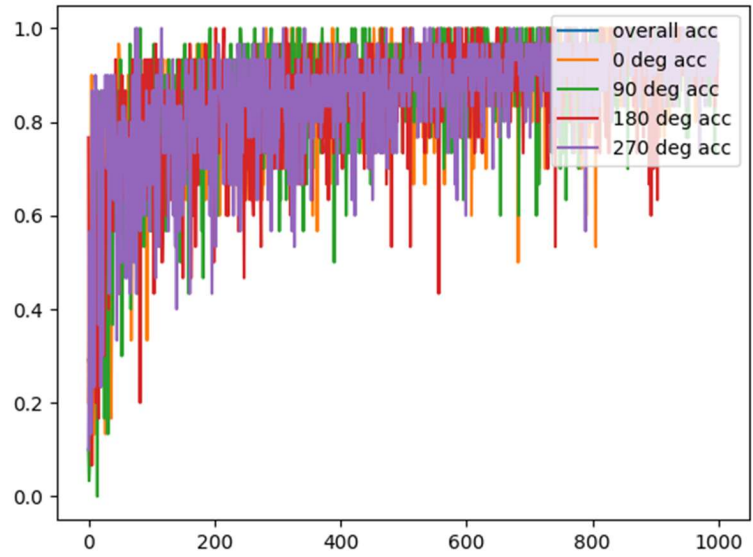

**Figure 4.8:** Overall Rotational Accuracy for classification of rotation angle on ShanghaiTech Part B

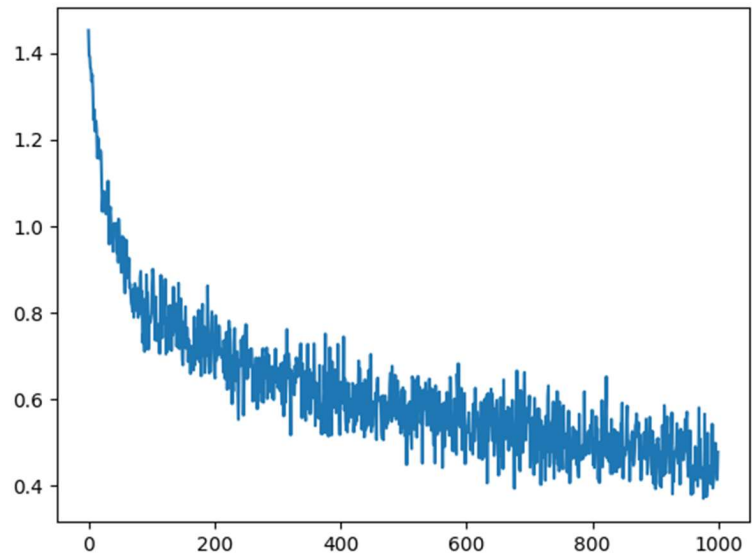

**Figure 4.9** Overall training loss during self-supervised training for ShanghaiTech PartB

# CHAPTER FIVE: CROWD ANOMALY DETECTION METHODOLOGY

## 5.1 Theoretical Background:

### 5.1.1 Frame Extraction:

A process to extract frames from video may be referred to as frame extraction. This process involves the extraction of still images at different time stamps from a video for analysis purposes. A video is always comprised of the number of frames that appear as a continuous stream. We may analyze a video by extracting the individual frame from a video. This process can be useful for different real-life applications such as video analysis, object recognition, and action recognition.

Extracted frames can be analyzed for object recognition by detecting and classifying the different objects present in the video. Similarly, in video analysis, frames can be used to analyze and understand the content, different movement patterns, or detection of events. These frames can also be useful for creating image-based data sets by extracting them at regular intervals. These image-based datasets can then be used for training, evaluation, and testing of machine learning models.

### 5.1.2 VGG-19 Architecture:

The Visual Geometry Group at The University of Oxford was the first to introduce it, and it was given the name VGG-19. VGG-19 [120] is an extended version of VGG-16 [120] with an increased number of convolutional layers, making it deeper than VGG16. While VGG16 served to establish the foundation for the study and creation of deep convolutional neural networks, VGG19 improved upon it by incorporating more layers to enhance representation learning capabilities. General network architecture of VGG-19 is (*as shown in* **Figure 5.1**) comprises of three different blocks as described below:

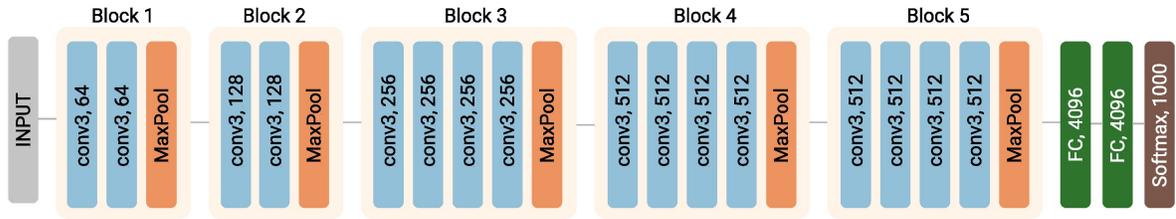

**Figure 5.1:** General Network Architecture for VGG-19

1. **Convolutional Block:** It starts with 16 convolutional layers and progresses to fully connected layers and a SoftMax classifier. Convolutional layers are further arranged in 5 blocks. All of these blocks are responsible for extracting spatial features by means of input data. Block 1 and 2 have two convolutional layers. Block 3, 4, and 5 have 4 number of convolutional layers. The convolutional layers are designed with small 3x3 filters and a stride of 1.

   i. **Block 1:** It consists of 2 convolutional layers followed by one max pooling layer. Each layer comprises 64 number of filters with a kernel size of 3x3.

   ii. **Block 2:** Just like block 1, it also consists of 2 convolutional layers followed by one max pooling layer. However, each layer of this block comprises 128 filters with a kernel size of 3x3.

   iii. **Block 3:** This block is different from block 1 and 2 in such a way that it comprises 4 convolutional layers followed by one max pooling layer. A kernel size of 3x3 with 256 number of filters is used for each layer.

   iv. **Block 4 and 5:** Block 4 and 5 are similar to each other in such a way that both of these blocks comprise of 3x3 kernel with 512 number of filters.

VGG-19 may detect more complex and abstract properties as information goes through the network as each block includes a number of convolutional layers

2. **Pooling Block:**

The main purpose of pooling layers is to down-sample the feature map. It plays an important role in downscaling the spatial dimensions of the feature map. VGG-19 utilizes max pooling layers for down sampling. Max pooling is one of the common types of pooling operation used in CNN and is generally effective in capturing the prominent features. Pooling layers are designed with a filter size of 2x2 to reduce the spatial dimensions by 1/2.

This down sampling reduces the computational complexity and memory needs of the following layers, making the network more computationally effective. The pooling layer concentrates on the most prominent features, regardless of their specific spatial location, by picking the highest value within each pooling region. As a result, it improves the ability of the network to generalize toward diverse areas of the network within the input data.

3. **Fully Connected Block:**

The purpose of the fully connected layers is to learn complex relationships and patterns in the data, enabling the network to make accurate predictions. These layers contribute to the end-to-end learning process of the VGG-19 network, allowing it to classify images into various classes based on the learned representations.

- It consists of 3 fully connected layers, and comes after 16 convolutional layers.
- These fully connected layers enable the model to capture high-level features and make predictions based on the extracted information.

- There are *4096* units present in the first two fully connected layers. With a high number of units, the network can successfully understand complicated data linkages and patterns, allowing it to generate more nuanced and accurate predictions.

- While the third layer consists of *1000 units*, which enables the model to classify among 1000 different classes or categories.

- The final fully connected layer's output is passed into a SoftMax classifier, which generates the probability distribution across distinct groups or categories. All of these units of FC layers connect to each neuron in the previous layer, resulting in a completely connected structure.

**5.1.3  Wide Residual Block**

Convolutional neural networks (CNNs) have revolutionized image classification by learning hierarchical features and integrating low, mid, and high-level features through multiple layers, with the depth of the network determining its performance [121]. Recent studies have shown that the depth of a deep learning model plays a vital role in achieving better results, and it has also been demonstrated by researchers the models with depth varying from sixteen to thirty layers have produced leading results [120], [122]. In research carried out by He et. al., [123], the author has demonstrated that the deeper network may suffer from a phenomenon name *"degradation".* This phenomenon occurs when the network's accuracy initially approaches a saturation point before rapidly dropping. Unexpectedly this phenomenon is not the result of over-fitting, as adding more layers to a sufficiently deep model actually increases training error. *(as shown in* **Figure 5.2***),* it is clear that the training error increases for a deeper network. Networks with 56 layers have high training errors as compared to networks with 20 layers. Researchers are actively investigating the degradation problem in deep learning to develop strategies to overcome it and fully utilize the potential of deep architectures, thereby advancing deep learning advancements. In order to resolve this issue of degradation problem, He et. al., introduced wide residual networks [123].

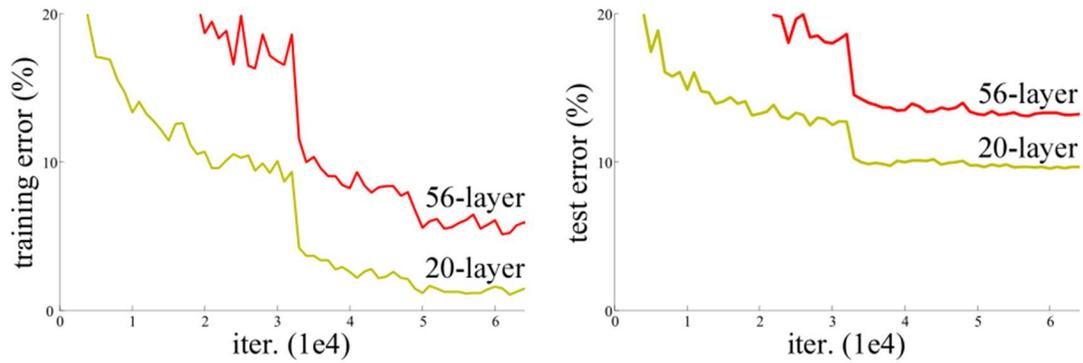

**Figure 5.2** Experimental results on Deep network using CIFAR-10. (Left) Training Error, and (Right) Testing Error

*Wide residual network* is a key component in deep learning architectures and is designed in such a way that it introduces wider blocks within the network to improve the effectiveness and performance of the model. A Standard residual block (*as shown in* **Figure 5.3**) is designed in such a way that involves convolutional layers. A residual connection adds the input to the output of convolutional. This residual connection helps in addressing the vanishing gradient problem and facilitates the learning process [124].

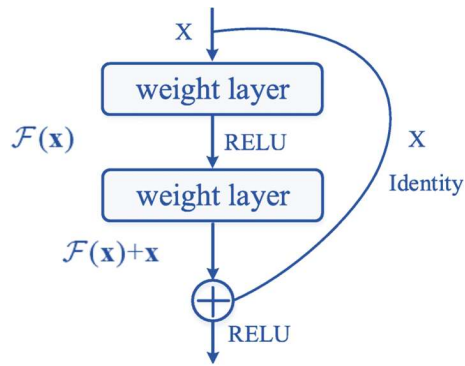

**Figure 5.3** General Architecture of Standard Residual Block

On the other hand, Wide Dense Residual Network [125] is (*as shown in* **Figure 5.***4*) an extension of the standard residual block, and can be designed in such a way that the convolutional branches or number of filters can be increased to get better performance for deeper networks. This enhanced width enables the neural network to detect more diverse and complicated elements in the input data. The network can learn richer representations and gain from higher model capacity if it has more channels. Additionally, the skip connection—which adds the original input to the convolutional layers' output—remains in the wide residual block. During training, this skip connection helps the gradient flow and allows the network to learn residual mappings. The wide residual block, a combination of wider and skip connections, is a powerful foundation for deep neural networks, enhancing learning and representation of complex features, thereby enhancing performance.

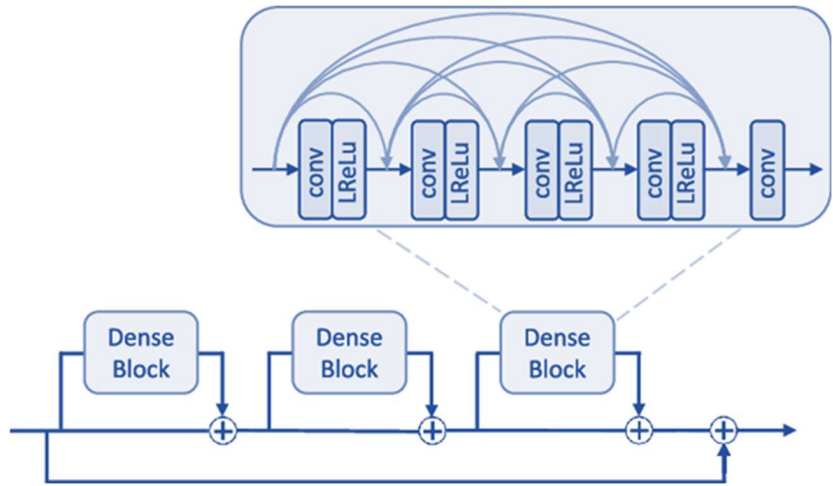

**Figure 5.4** General Architecture of Wide Dense Residual Block

### 5.1.4 Long Short-Term Memory (LSTM):

Long Short-Term Memory (LSTM) is an advanced form of Recurrent Neural Network (RNN), designed in such a way as to address the challenges faced by traditional RNN in capturing long-term dependencies. vanishing gradient is one of the main problems that was observed in traditional RNN, as the gradient diminishes exponentially when they propagate back in time. It makes the model learn and remember the data over extended periods of time, making the model unable to effectively model and process sequential data. It is achieved through the addition of memory blocks and gating mechanisms by controlling the flow of information within the network. It generally comprises of the following parts (*as shown in* **Figure 5.5**):

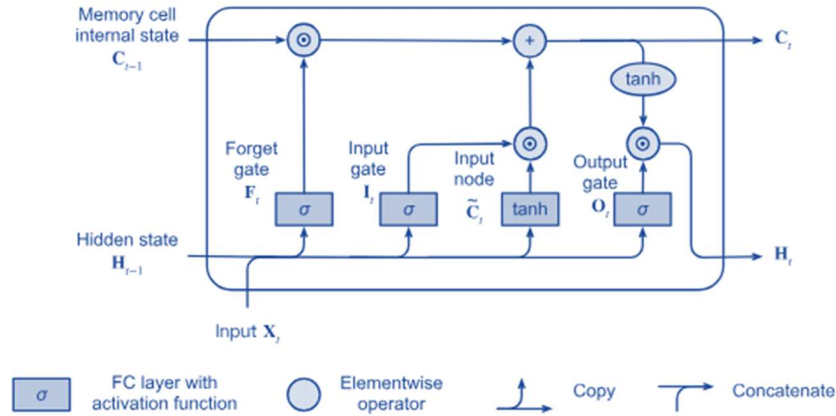

**Figure 5.5** General Architecture of LSTM

1. **Gates:** Generally, the LSTM structure consists of different gates with the aim of controlling the flow of information within the network. Sigmoid activation function is used to implement these gates, and having values between 0 and 1.

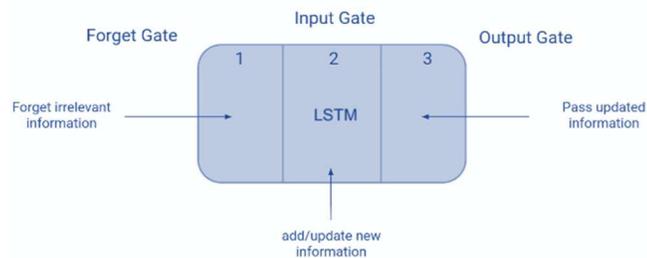

**Figure 5.6** Three Gates of LSTM Unit

   a. **Input Gate:** This gate is generally responsible for determining the amount of new data that needs to be kept in the memory cell at each time interval. Input is taken in the form of a combination of current input and previous hidden state. Its output will be denoted as "i", and responsible for deciding the amount of new data that should be stored in the memory cell. The output will be obtained by applying the Sigmoid activation function to the input.

   b. **Forget Gate:** This gate is generally responsible for determining the amount of data that should be forwarded to the next layer. Just like the input gate, the input of the forget gate will also be taken in the form of a combination of current input and previous hidden state. Its output will be denoted as "f", and responsible for deciding the amount of previous that should be forgotten from the memory cell. The output will be obtained by applying the Sigmoid activation function to the input.

   c. **Output Gate:** This gate is generally responsible for determining the amount of old/previous data that needs to be discarded from the memory cell at each time interval. Input is taken in the form of a combination of current input, previous hidden state, and current cell state. Its output will be denoted as "o". Output will be obtained by applying the Sigmoid activation function to the input. This output is then multiplied element-wise with the current memory cell state after applying the input and forget gates to produce the final hidden state or output.

2. **Memory Cells:** It is one of the fundamental units and is generally responsible to store and update information over time. It maintains the internal state and is generally denoted as "C". The internal state, main function of the internal state is to act as a conveyor belt that transports data between various time steps. It enables the network to keep and maintain relevant data while removing unnecessary elements. The input gate regulates how much new information should be stored in the cell state at each time step, updating the internal state of the memory cell.

The forget gate regulates the amount of prior knowledge that should be removed from the cell state. By allowing the network to selectively remember or discard data, these gates make sure that significant patterns and dependencies are preserved in the memory cell. Memory cells are crucial for capturing long-term dependencies in sequential data.

The LSTM network can efficiently store and use information to produce meaningful outputs or make accurate predictions by updating and preserving its internal state over time. Spatial features can be defined as the arrangement and distribution of persons or objects in space. While temporal features may refer to the relative change and patterns that occur over time.

## 5.2 End-to-End Proposed Framework:

In our research, we proposed a spatio-temporal model *(as shown in* **Figure 5.***7)* based on the combination of Convolutional neural network (CNN) and Long Short-Term Memory (LSTM) for detection of anomalies in crowd scenes. Videos will be taken as input and the proposed model is designed in such a way that it can learn both spatial and temporal features by leveraging the supervised training.

The main advantage of this combined extraction of spatial as well as temporal features is that it makes the model generalize towards real-world scenes, and provides a comprehensive understanding of videos. Spatial features are captured by leveraging CNN to learn the position/arrangements of individuals present in the crowd. while temporal features learn the movement and change of persons present in the crowd with the help of LSTM blocks by capturing the relationship between two consecutive frames. The model works on binary classification and can detect either normal or abnormal behavior.

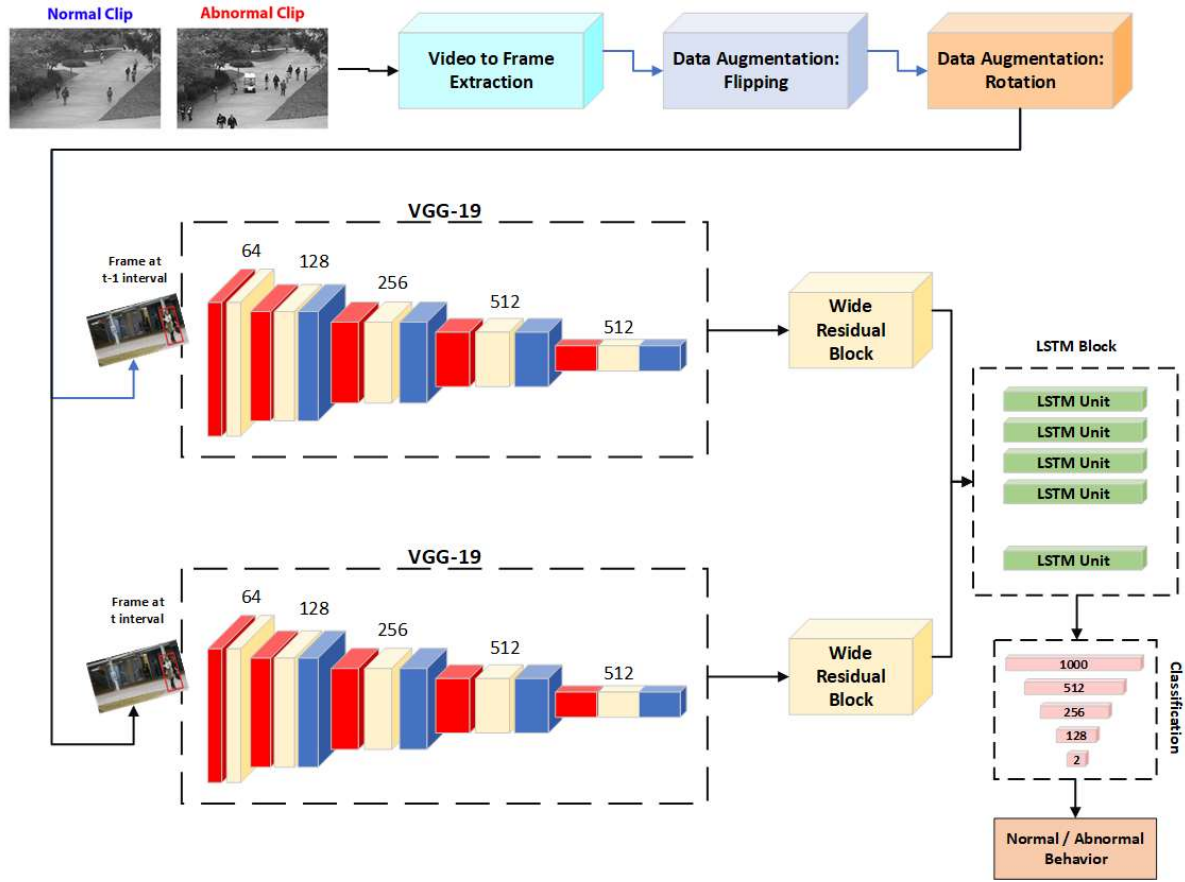

**Figure 5.7** Proposed Model for Crowd Anomaly Detection

We proposed VGG-19 architecture of CNN for the extraction of spatial features by analyzing individual frames of the videos. VGG-19 consists of 5 convolutional blocks (16 convolutional layers), and 1 fully connected block. It follows a uniform structure, with small-sized convolutional filters (3x3) and a stride of 1, ensuring more detailed feature extraction. Following each convolutional layer, a rectified linear unit (ReLU) activation function is employed to introduce non-linearity, increasing the network's ability to understand complicated relationships within the input.

On a regular basis, max-pooling layers are used to minimize the spatial dimensions of the feature maps while maintaining the most important characteristics. Max pooling divides feature maps into small regions and retains just the maximum value within each zone, lowering overall computing costs and giving translation invariance. We modified the architecture of VGG-19 *(as shown in* **Figure 5.***8)* by replacing the fully connected layers with a deep residual block. We added dense residual layers at the output of the 4th layer of the 5th convolutional block. By doing so, we may be able to capture the short occurring scenes and enhance the performance of the model in terms of efficiency and accuracy as a result. The original VGG-19 architecture focused on depth and smaller 3x3 filters for convolutional layers, but it can become computationally expensive due to the large number of parameters. To address this, the modified architecture introduces wide dense residual blocks, which combine residual connections and dense connections, which have been successful in improving deep neural network training and performance.

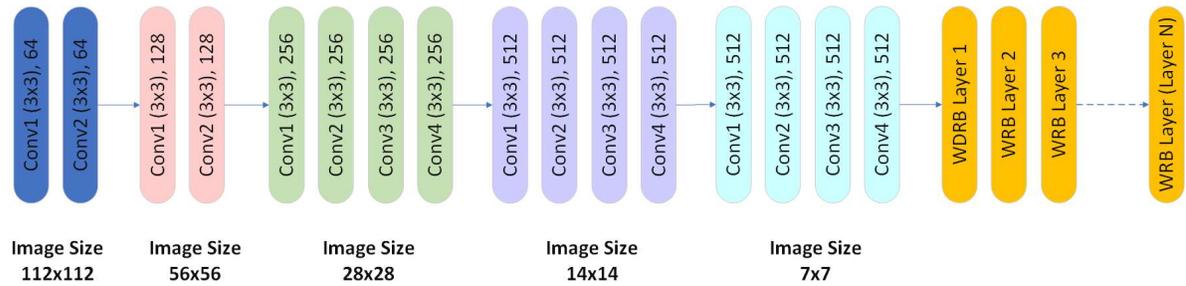

**Figure 5.8** Proposed architecture of modified VGG19

In our work, we proposed a ***Wide Dense Residual Block with VGG-19*** by modifying the standard residual block with the aim of improving the performance of the proposed model. Generally, the residual block in VGG19 is made up of two convolutional layers in series with a skip connection connecting the input of 1st convolutional block to the output of the second convolutional block. Inspired by this setup, we extend the standard residual network by incorporating the additional convolutional layers (with an increased number of filters), batch normalization layers, activation function layers, and skip connection. The output of each layer is concatenated with the previous layer's output, forming a dense connection. Additionally, residual connections are introduced, allowing the direct flow of information from earlier layers to later layers. Wide dense residual blocks have one key benefit: they make it easier for gradients to flow efficiently during training, which improves optimization and speeds up convergence. The network can access a rich set of features at various levels by mixing the outputs of all previous layers, which improves representation learning. General architecture of our proposed WDRB is (***as shown in*** **Figure 5.2**).

First of all, videos will be taken as input and will be subjected to video-to-frame extraction. These frames will be then subjected to different data augmentation techniques. these process frames will be then used for video analysis by considering the frames at time stamps ("t" and "t-1"). These frames at time stamps ***("t" and "t-1")*** will be fed to two separate CNN blocks of modified VGG-19 for extraction of spatial features. These spatial features will be then fed to the LSTM unit for extraction of temporal features. These spatial features enable the model to recognize actions and capture video descriptions. However, temporal features help the model to learn about movements by learning the patterns of motion. Combined spatio-temporal features will be then fed to a fully connected layer for binary classification of normal and abnormal events. The fully connected layers serve as the classifier at the end of the network. They take both spatial and temporal features extracted by the preceding convolutional layers and map them to the desired output classes. Flow of our proposed system is ***(as shown in Figure 5.9)***

# CHAPTER SIX: CROWD ANOMALY DETECTION EXPERIMENTAL RESULTS

## 6.1 Implementation Details:

### 6.1.1 Data Pre-Processing:

Data Pre-processing involves the following steps:

1) First of all, we split the dataset into two parts; one is named as violent, and the other is named as non-violent. This helps researchers understand the prevalence and characteristics of violent content, distinguishing violent from non-violent videos, and establishing a clear framework for future analyses.

2) The data sets utilized in the experimental results comprise videos with a duration ranging from 1 to 5 seconds. This range ensures that the data sets encompass a variety of video lengths, enabling a comprehensive analysis and evaluation of the proposed methods.

3) A dedicated function *extract_frames()* is defined for the extraction of frames from videos. By using this function, we performed the video-to-frame conversion and extracted frames from videos. An equal number of frames from each video are extracted at regular intervals, ensuring a balanced representation across the dataset.

4) A function named as *process_videos()* is defined for the processing of videos by taking three parameters as input; directory, label, and max-frames. Directory takes the path of videos as input. Label represents the label value to assign to the extracted frames. max_frames represent the maximum number of frames to be extracted from videos.

a. Two empty lists named as *data[ ]* and *labels[ ]* are defined inside the function. *data[ ]* takes extracted frames and labels take the corresponding label of the video.

b. The function checks if a file's file extension is '.avi', assuming it's a video file. It obtains the video's full path by joining the directory path and filename.

c. The extract_frames() function extracts frames from the video, appending them to the data list and labeling them. After processing all videos, the data and label lists are returned as NumPy arrays.

5) These frames are then resized to a standardized dimension of *128x128* pixels. This uniform resizing facilitates consistent analysis and processing of the video content.

6) Frames are extracted from videos at the rate of 25 frames per second.

7) Frame padding is also done by using *extract_frames()* function. If the length/duration of the video is too small then it will create a number of frames less than a defined number of frames. In this scenario, we performed zero padding. It works in such as way that ZERO will be assigned to empty spaces of NUMPY array so that the length of the array should be the same as the desired length.

8) Videos are processed in such a way that 1 is assigned to violent videos and 0 is assigned to non-violent videos. Violent and non-violent videos will be then combined in a new array denoted by X. The data is arranged in such a way that violent videos come before non-violent videos.

9) Similarly, a new array denoted by Y is created to combine the labels of videos. Labels are also arranged in the same order as followed in the arrangement of videos.

*10)* After that, the data (including videos and labels) are shuffled by using the built-in function *(np.random.permutation()).* An unpredictable sequence of indices that spans from 0 to the data length. Data length is determined by the length of the array that was used to store videos. As a result, two new arrays ***X-Shuffled*** and ***Y-Shuffled*** will be created. These arrays contain shuffled data in the form of videos and their respective labels.

*11)* Additional training samples are generated due to the limited availability of data by applying different data augmentation techniques such as flipping, rotation, changing brightness, and zoom transformation.

    a. Images will undergo horizontal flipping, where they are flipped from left to right. This flipping operation is applied with a probability of 1.0, meaning that all images in the dataset will be flipped.

    b. The augmentation technique includes a zoom transformation applied to the images, whereby they are scaled up by a factor of 1.3. This zooming operation results in the images being magnified or zoomed in. The transformation enhances the model's learning ability by providing a closer view of depicted objects, improving its generalization to different scales and perspectives.

    c. The brightness of the images is randomly adjusted. This adjustment is achieved by multiplying the pixel values by a random factor between 1 and 1.3.

    d. A rotation transformation is applied to the images. This transformation randomly rotates the images either clockwise or counterclockwise by an angle between -25 and 25 degrees.

## 6.2  Training Details:

We used VGG19 architecture of CNN for the extraction of spatial features by leveraging the pre-trained weights of the model. Input data is trained using additional blocks of LSTM and WDRB. As we are using pre-trained architecture, we didn't train it again on input data and it remains unchanged. By using this method, the pre-trained VGG-19 model's robust feature extraction characteristics are able to be utilized by the model. It also helps the model to concentrate on improving the LSTM units with respect to the assigned task. By keeping the pre-trained VGG19 model stable, the model can use the learned characteristics and patterns collected by the VGG-19 model, while the LSTM units are educated to adapt and specialize in the specific job at hand.

- *Total units of LSTM = 256:* It represents the dimensions of the hidden state and cell state vectors within each LSTM cell.
- *Dropout = 0.2:* To prevent over-fitting during the training process, we applied dropout regularization with a probability of 20% (0.2) between LSTM cells. It improves the model's ability to generalize towards unseen and new data by randomly setting 20% of the inputs to the LSTM cells to 0 at each update during training.
- *Optimization function:* Cross entropy loss is minimized by using the ADAM Optimizer as an optimization function for backpropagation of weights and biases. It allows the model to efficiently update the weight and biases of the model. It also allows the model to converge fast with improved performance during backpropagation.
- *Number of Epochs = 50:* For every training set, the network undergoes 50 epochs of training. Each epoch consists of one forward pass, and one backward pass to improve the performance of the model by updating the weights and biases.

- *Learning Rate = 0.001:* For every training set, the network undergoes 50 epochs of training. Each epoch consists of one forward pass, and one backward pass to improve the performance of the model by updating the weights and biases.

## 6.3   Datasets:

### 6.3.1   Hockey Fight Dataset

This dataset consists of two different types of videos; fighting and non-fighting videos. These videos were taken from ICE Hockey matches played in National Hockey League (NHL). The NHL is the premier professional ice hockey league in North America, featuring teams from both the United States and Canada.

1) Total number of videos: 2000 Nos.
2) Total number of fighting videos: 1000 Nos.
3) Total number of Non-fighting videos: 1000 Nos.
4) Approximate duration of each video: 02 Seconds.
5) Approximate number of frames in each video: 50 frames.
6) Resolution of each frame: 360x280

### 6.3.2   Smart City Violence Detection Dataset

## 6.4 Experimental Results on Hockey Fight Dataset

### 6.4.1 Videos to Frame Extraction

In this step, we defined a special function to extract images/frames from videos by taking the video as input, resize as per the desired size, extracting frames from the video, and stored them for further processing. All of the extracted frames are stored in a NUMPY array.

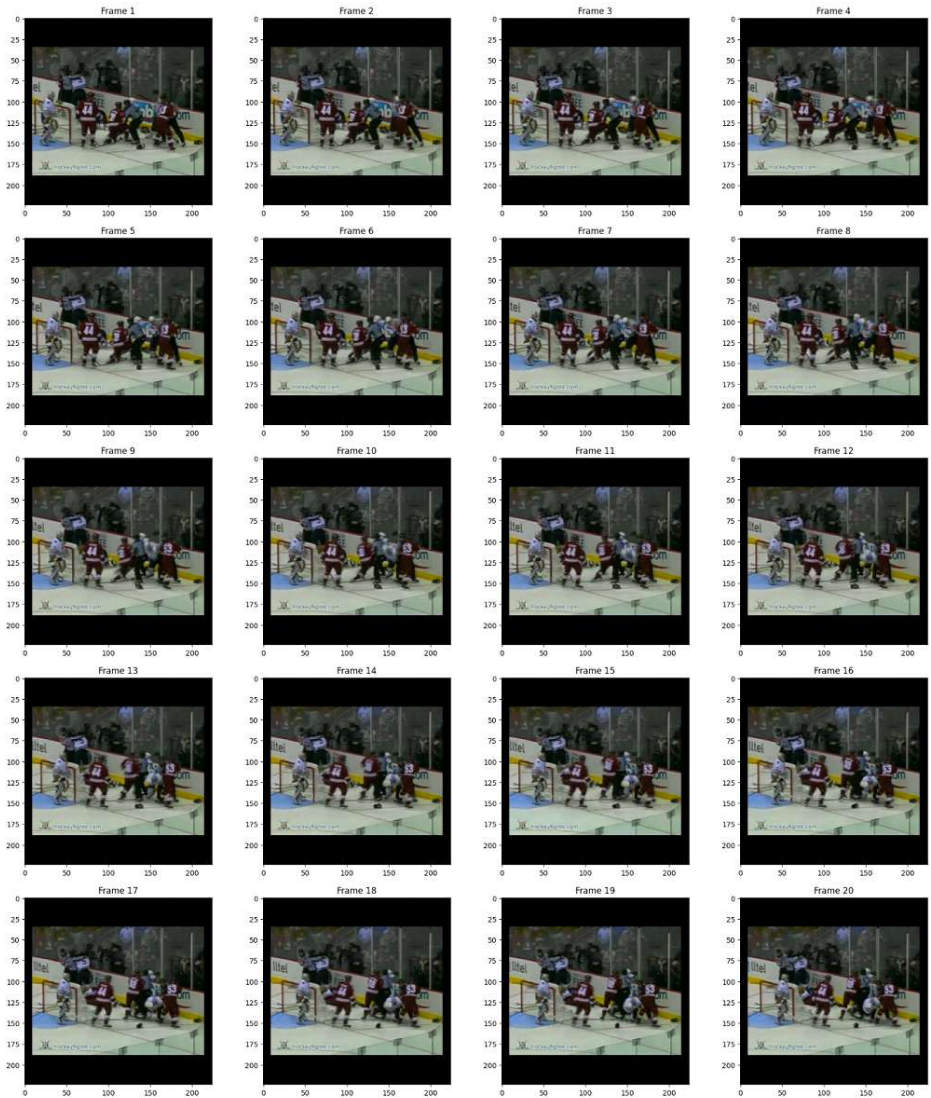

**Figure 6.1:** Violent Video: Frames 1 to 20

### 6.4.2 Confusion Matrix:

Confusion matrix for results on Hockey fight dataset is (*as shown in* 6.2). Abnormal activities are represented by Label 1 (Positive). While normal activities are represented by Label 0 (Negative).

1) Recall, also known as sensitivity, is 0.82. This means that the model correctly identifies 82% of the actual positive instances.
2) The F1-score, which is the harmonic mean of precision and recall, is 0.86. This statistic provides a balanced evaluation of the model's performance, accounting for both precision and recall.

These performance measures indicate that the model has a high precision, which means that the majority of its positive predictions are right. The recall is also relatively high, showing that the model can detect a significant percentage of true positive cases. The F1- score, which is a combination of precision and recall, provides an extensive evaluation of the model's efficacy.

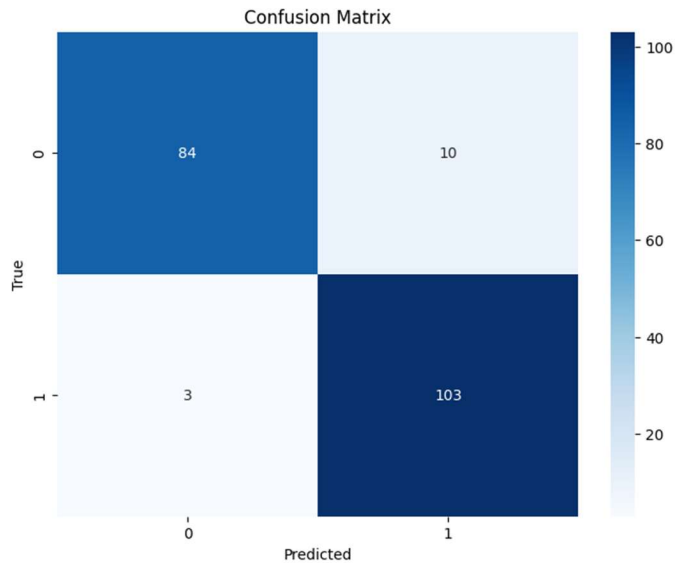

**Figure 6.2:** Confusion Matrix for Experimental Results on Hockey Fight Dataset

### 6.4.3 Evaluation Results

As presented in **Table 6.*1***, the precision is 0.91, recall is 0.82, and F1-score is 0.86.

**Table 6.1:** Evaluation Results on Hockey Fight Data set

| Dataset Name | Precision | Recall | F1-Score |
|---|---|---|---|
| **Hockey Fight Dataset** | 0.91 | 0.82 | 0.86 |

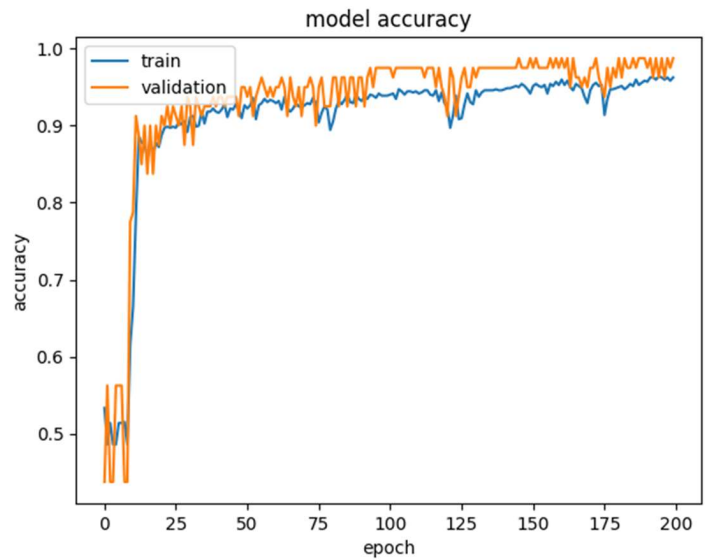

**Figure 6.3: Model Accuracy on Hockey Fight Dataset**

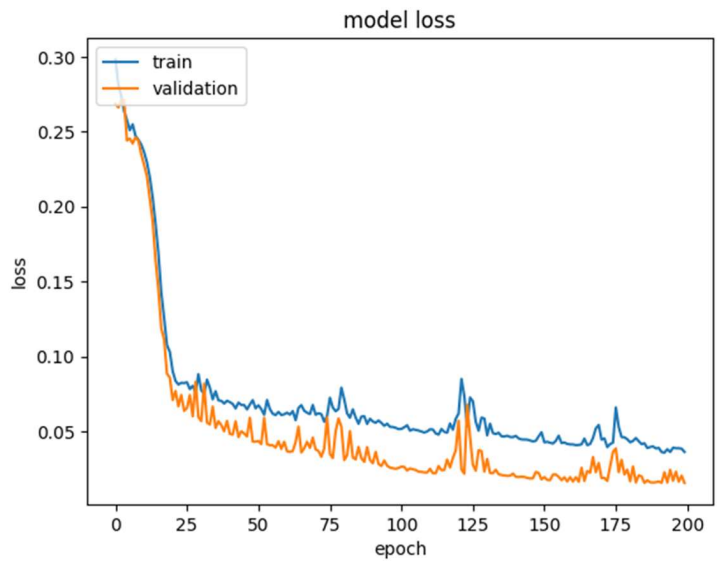

**Figure 6.4: Model Loss on Hockey Fight Dataset**

## 6.5 Experimental Results on SCVD:

### 6.5.1 Confusion Matrix:

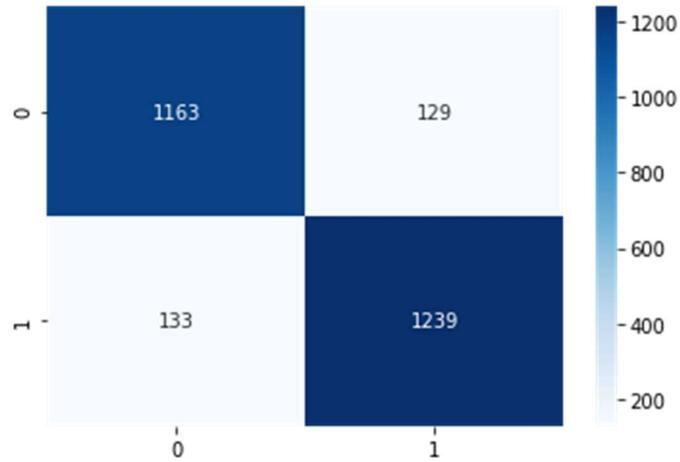

**Figure 6.5:** Confusion Matrix for Experimental Results on SCVD

### 6.5.2 Evaluation Results:

**Table 6.2:** Evaluation Results on SCVD

| Dataset Name | Precision | Recall | F1-Score |
|---|---|---|---|
| **Hockey Fight Dataset** | 0.93 | 0.92 | 0.90 |

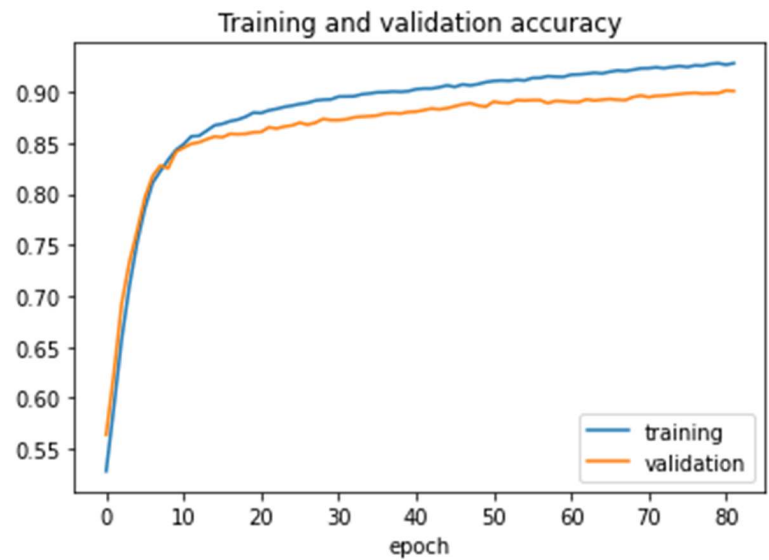

**Figure 6.6:** Model Accuracy on SCVD

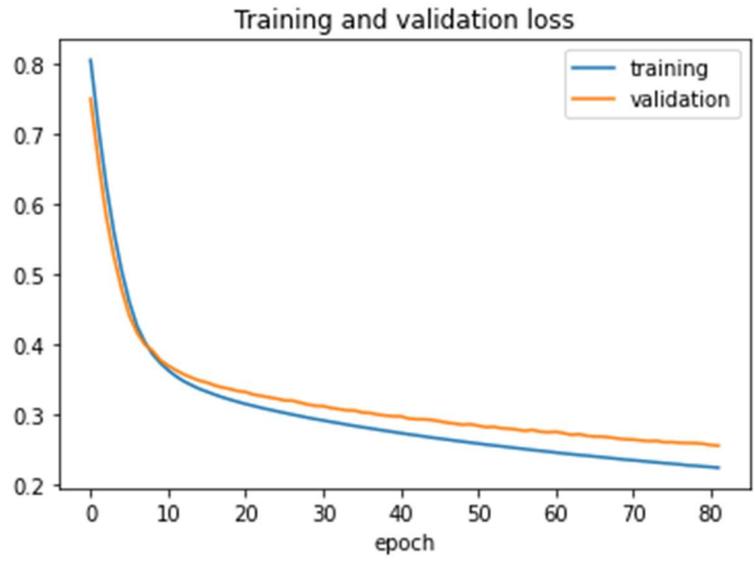

**Figure 6.7: Model Loss on SCVD**

# CHAPTER SEVEN: DISCUSSIONS

Deep learning is a powerful tool for addressing complex computer vision tasks using large-scale datasets. It has revolutionized various aspects of daily life, including crowd counting, traffic monitoring, public safety, and city planning. Crowd counting involves estimating and tracking the number of people in densely populated areas in real-time, enabling proactive measures to prevent accidents. Surveillance cameras equipped with crowd counting algorithms can continuously monitor crowded areas, providing valuable information for efficient traffic management, crowd control, and public safety. This data can also aid in city planning by identifying high-footfall areas, facilitating resource allocation and infrastructure development. Deep learning models capture complex patterns and variations in crowd dynamics, enabling accurate and reliable crowd counting. By leveraging deep learning, crowd counting becomes an automated and efficient process, reducing reliance on manual counting methods and enhancing situational awareness. In addition to crowd counting, different type of crowd analysis such as behavior analysis, flow analysis, and anomaly detection also plays an important role in our daily life. Providing safe and secure environment during events and gatherings is also necessary. In our research work, we carried out our research on two different domains of crowd scene analysis; crowd counting and crowd anomaly detection. In the first part, we proposed a novel approach based on combination of self-supervised training and M-CNN for an effective and reliable estimation of individuals present in a crowded scene. M-CNN is used as feature extraction network with different branches such as rotation classification, self-supervised training, and distribution matching. We evaluated our proposed model on publicly available dataset Shanghaitech and found that our proposed model performs state-of-the-art results. In future, we may extend our work in the field of self-driving cars by estimating the number of cars on the way to suggest better suitable selection of route.

In second part, we proposed a spatio-temporal model for anomaly detection in a crowded scene based on combination of VGG19 and LSTM. VGG19 is modified by replacing the fully connected layer with dense residual block to extract spatial features and LSTM is incorporated to learn the temporal features. We evaluated our proposed model on hockey fight dataset and found that the model performed an excellent result. We may extend this work in the field of self-driving cars to analyze the abnormal behaviors on road such as peoples walking or running on the road, peoples fighting on the road or, crowd due to accidents